\pdfoutput=1

\documentclass[11pt,xcolor={table}]{article}

\usepackage{acl2023}

\usepackage{times}
\usepackage{latexsym}
\usepackage{listings}
\usepackage[most]{tcolorbox}
\usepackage{colortbl}
\usepackage{multirow}
\usepackage{booktabs}
\usepackage[T1]{fontenc}

\usepackage{amssymb}

\usepackage[utf8]{inputenc}
\usepackage{booktabs}
\usepackage{microtype}

\usepackage{inconsolata}
\usepackage{xspace}
\usepackage{siunitx}
\usepackage{enumitem}

\usepackage{bbm}

\newcommand{\codefont}[1]
{{\fontfamily{qcr}\selectfont #1}}

\newcommand{\bench}{\texttt{SURGE}\xspace}

\definecolor{commentcolor}{rgb}{0.5,0.5,0.8}
\definecolor{symbolcolor}{rgb}{0.5,0.2,0.8} 
\definecolor{sortcolor}{rgb}{0.5,0.5,0.8}  
\definecolor{darkblue}{rgb}{0, 0, 0.5}
\hypersetup{colorlinks=true, citecolor=darkblue, linkcolor=darkblue, urlcolor=darkblue}
\definecolor{tacticcolor}{rgb}{0.16, 0.32, 0.75} 

\lstdefinestyle{leanstyle}{
    language=lean,
    basicstyle=\small\ttfamily,
    keywordstyle=\color{blue},
    stringstyle=\color{red!60!black},
    commentstyle=\color{green!60!black},
    breaklines=true,
    showstringspaces=false,
    numbers=left,      
    numberstyle=\tiny\color{gray}, 
    stepnumber=1,
    numbersep=5pt,
    mathescape=true
}

\newtcblisting{leancode}[1][]{
    listing engine=listings,
    listing style=leanstyle,
    boxrule=0.5pt,
    colback=white,
    colframe=gray!70,
    arc=0pt, 
    boxsep=0pt,
    left=2pt, right=2pt, top=2pt, bottom=2pt,
    listing only,
    title style={colback=white, colframe=gray!70, coltext=black},
    fonttitle=\bfseries,
    #1, 
}

\newtcblisting{plainleancode}[1][]{
    listing engine=listings,
    listing only,
    listing options={
        language=lean,
        basicstyle=\small\ttfamily,
        keywordstyle=\color{black}, 
        stringstyle=\color{black},  
        commentstyle=\color{black}, 
        breaklines=true,
        showstringspaces=false,
        numbers=none,              
        mathescape=true
    },
    boxrule=0.5pt,
    colback=white,
    colframe=gray!70,
    arc=0pt,
    boxsep=0pt,
    left=2pt, right=2pt, top=2pt, bottom=2pt,
    breakable,
    fonttitle=\bfseries,
    #1,
}

\title{\bench: On the Potential of Large Language Models as General-Purpose Surrogate Code Executors}

\author{
 \textbf{Bohan Lyu\textsuperscript{1\ *\ †}}\quad
 \textbf{Siqiao Huang\textsuperscript{2\ *}}\quad
 \textbf{Zichen Liang\textsuperscript{1\ *}}
\\
 \textsuperscript{1}{Department of Computer Science and Technology, Tsinghua.}
 \\
 \textsuperscript{2}{Institute for Interdisciplinary Information Sciences (IIIS), Tsinghua.}
\\
{\tt lyubh22@gmail.com, \{huang-sq23, liang-zc22\}@mails.tsinghua.edu.cn}
\\[2ex]
\textsuperscript{*}Equal contribution \hspace{2em} \textsuperscript{†}Corresponding author
}

\begin{document}
\maketitle
\begin{abstract}
    Neural surrogate models are powerful and efficient tools in data mining.  Meanwhile, large language models (LLMs) have demonstrated remarkable capabilities in code-related tasks, such as generation and understanding.  However, an equally important yet underexplored question is whether LLMs can serve as surrogate models for code execution prediction.  To systematically investigate it, we introduce \bench, a comprehensive benchmark with $1160$ problems covering $8$ key aspects: multi-language programming tasks, competition-level programming problems, repository-level code analysis, high-cost scientific computing, time-complexity-intensive algorithms, buggy code analysis, programs dependent on specific compilers or execution environments, and formal mathematical proof verification. Through extensive analysis of $21$ open-source and proprietary LLMs, we examine scaling laws, data efficiency, and predictive accuracy. Our findings reveal important insights about the feasibility of LLMs as efficient surrogates for computational processes. The benchmark and evaluation framework are available at \url{https://github.com/Imbernoulli/SURGE}.
\end{abstract}

\section{Introduction}

\begin{figure}[!t]
    \centering
    \includegraphics[width=0.95\linewidth]{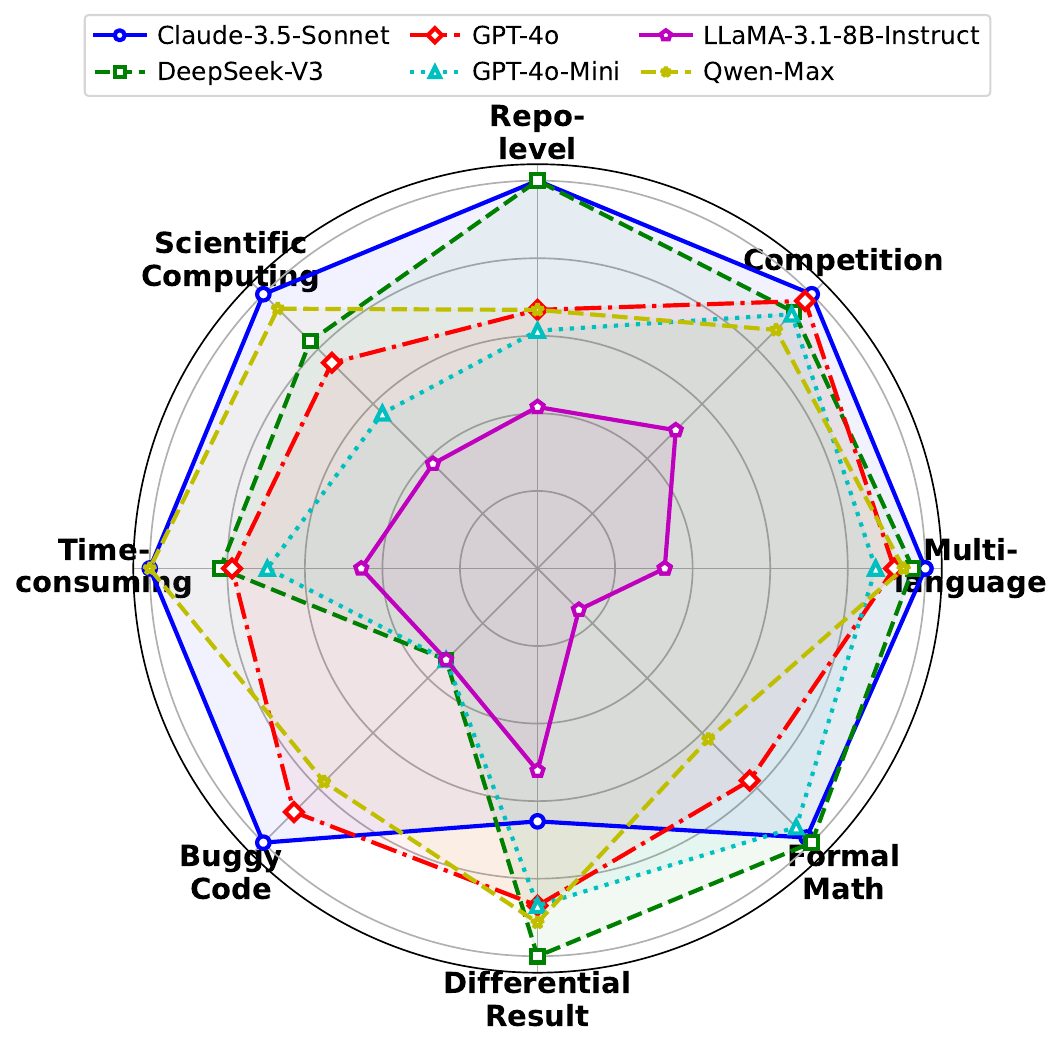}
    \caption{Performance of $6$ typical models on \bench.}
    \label{fig:case1}
\end{figure}

Neural surrogate models~\citep{zhang2024artificialintelligencesciencequantum,sun2019review} are powerful tools in data mining and machine learning, which efficiently approximate complex computational processes. Meanwhile, Large language models (LLMs)~\citep{reid2024gemini, llama3, claude3, hui2024qwen2, bi2024deepseek} have demonstrated remarkable capabilities in code-related tasks~\citep{lu2021codexglue, zheng2023codegeex, luo2023wizardcoder, codeqwen, guo2024deepseek}, including code understanding~\citep{ahmad2020transformer, chakraborty2020deep} and code generation~\citep{li2017code, parvez2018building}. However, an equally important yet underexplored question is whether LLMs can serve as general-purpose surrogate code executors, which predict the behavior of a program without actually running it. A recent study~\citep{lyu2024largelanguagemodelscode} acknowledges its importance, however, it focuses on a case study rather than a systematic analysis.

The ability to predict code execution outcomes without execution has tremendous significance. In scientific computing, running simulations often requires substantial computational resources and is time-consuming, making it impractical to test every possible configuration~\citep{lu2019efficientsurrogatemodelingmethods, Non-intrusive, doi:10.1137/130932715}. In security-sensitive environments, executing untrusted code poses inherent risks, necessitating alternative mechanisms for assessing program behavior without exposing the system to potential vulnerabilities~\citep{nebbione2023methodological, ccs, wang2024uniquesecurityprivacythreats}. Additionally, some code requires highly specific execution environments, which may not always be available, making surrogate execution a valuable alternative~\citep{10.1145/3617591, gu2025softwaretestingextendedreality}. Moreover, accurately predicting a model’s potential outputs or errors is crucial for improving traditional tasks such as code understanding, code generation, and even math reasoning~\citep{li2025codeiocondensingreasoningpatterns}. Lastly, many works use LLMs as reward models (RMs) in reinforcement learning. For code tasks, accurate execution prediction is key to a reliable RM~\citep{ouyang2022traininglanguagemodelsfollow}.

Traditional approaches to surrogate code executing~\citep{10.1145/360248.360252, 10.1145/2408776.2408795} struggle to generalize across languages and suffer from scalability issues when applied to complex real-world codebases. Containerized environments~\citep{10.5555/2600239.2600241} mitigate dependency issues but still require full code execution. Recent efforts to train neural executors~\citep{yan2020neuralexecutionengineslearning} focus on narrow tasks and lack the generality needed for real-world code. In contrast, LLMs’ capacity to internalize patterns from vast code corpora~\citep{CodeXGLUE, codealpaca} suggests a path toward general-purpose surrogate code execution. 

To understand the potential of LLMs as \textbf{GE}neral-purpose \textbf{SUR}rogate code executors, we introduce \textbf{\bench}. It includes $8$ components: (1) fundamental programming tasks in multiple languages, (2) competition programming problems requiring deep logical inference, (3) repository-level codebases that test long-range dependencies, (4) scientific simulations and optimizations where direct execution is high-cost, (5) time-consuming logical algorithms that have high time-complexity, (6) buggy code that examines LLMs' ability to predict runtime errors, (7) programs whose behavior depends on specific compiler versions or execution environments and (8) math theorem proving in formal language~\citep{de2015lean, moura2021lean} which expects compilers to testify.

Through extensive evaluation of 21 open-source and proprietary LLMs on \bench, we provide the first large-scale study of LLMs' capabilities as computational surrogates. Additionally, we investigate the impact of various factors, including prompt engineering strategies, programming language characteristics, computational complexity, and execution time requirements, on surrogate performance. Our findings reveal both the promising potential and current limitations of LLMs as general-purpose code execution surrogates. The performance of typical models on \bench is shown in Figure~\ref{fig:case1}.

\begin{figure}[!t]
    \centering
    \includegraphics[width=\linewidth]{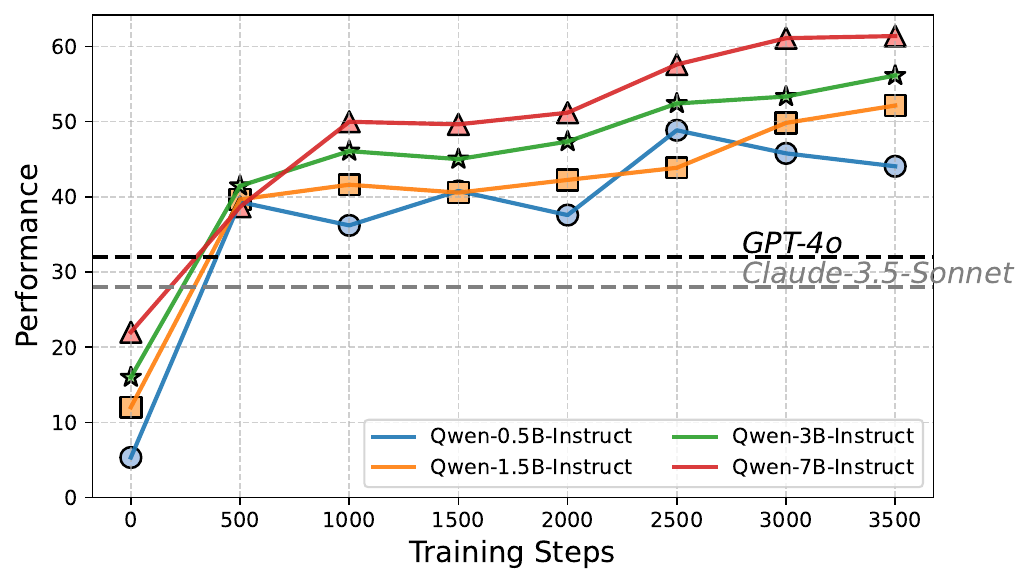}
    \caption{Performance scaling across model sizes and training steps.}
    \label{fig:case2}
\end{figure}

Beyond benchmarking, we conduct a scaling law study on whether LLMs' performance improves with model size and training data. We train models with 4 different sizes on different scales of training data from the formal language subset of \bench. Our experiments demonstrate that models' performance consistently improves with both model size and training steps, with larger models showing stronger learning capacity and higher performance ceilings throughout the training process~(Figure~\ref{fig:case2}).

In short, our work makes the following key contributions:

\begin{itemize}[noitemsep, topsep=0pt]
    \item We introduce \bench, the first holistic benchmark for evaluating LLMs as general-purpose surrogate code executors. It consists of $8$ subsets and $1160$ problems.
    \item We evaluate $21$ open-source and proprietary LLMs on \bench and conduct the first large-scale analysis on them.
    \item We present a scaling law study with models of varying sizes and scales of training data, providing empirical insights on the scaling law of LLMs on these tasks.
\end{itemize}
\section{Related Works}

\begin{figure*}[!ht]
\centering
\includegraphics[width=\linewidth]{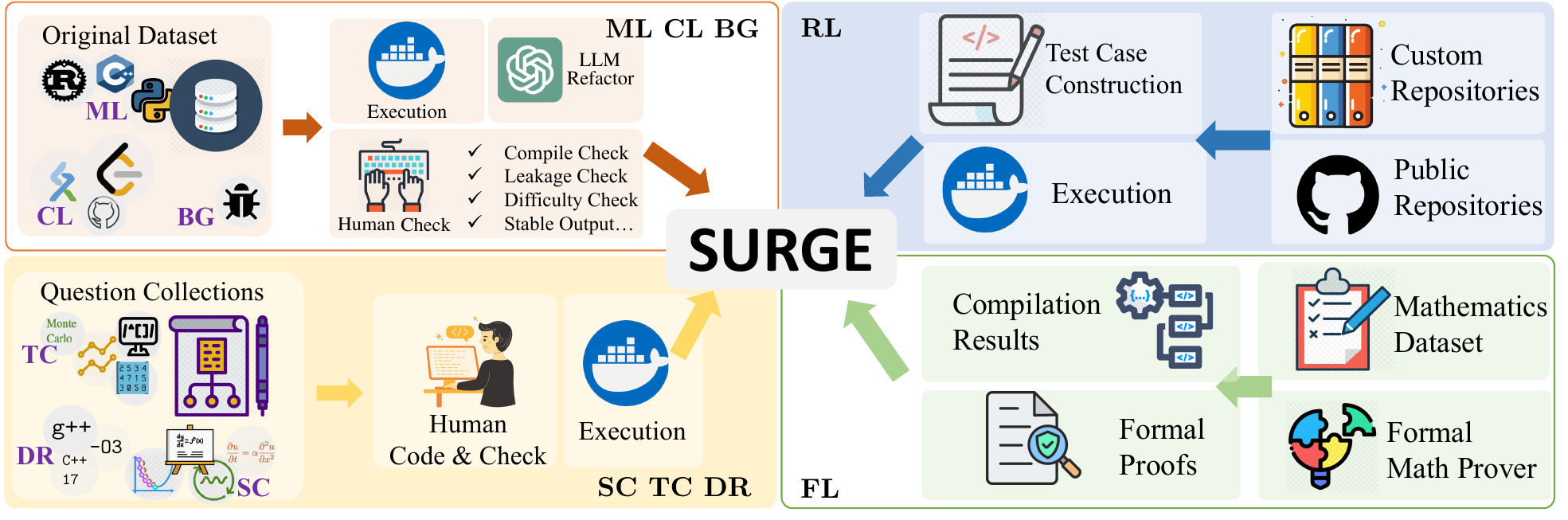}
\caption{The Construction of \bench employs $4$ methodologies: 1. Iterative Refactor, 2. Repository Sampling, 3. Manual Implementation, and 4. Inference \& Verification.}
\label{fig:construction}
\end{figure*}

\paragraph{Neural Surrogate Models.} Neural surrogate models are neural network-based approximations used to replace computationally expensive simulations in various scientific and engineering domains~\citep{zhang2024artificialintelligencesciencequantum, sun2019review}. These models act as domain-specific emulators by learning complex input-output relationships from high-fidelity data\citep{raissi2020hidden, sun2020surrogate, bessa2017framework, thuerey2020deep, raissi2019physics, willard2022integrating}. Recently, generative models have been incorporated into surrogate modeling. Some equip language models with traditional surrogate models to facilitate iterative optimization~\citep{ma2024llmsimulationbileveloptimizers, lyu2025adaptinglearninggroundingllms}, and some use generative models to realize the end-to-end surrogate process~\citep{gruver2024largelanguagemodelszeroshot, hao2024largelanguagemodelssurrogate, wimmer2023leveragingvisionlanguagemodelsgranular, che2024gamegenxinteractiveopenworldgame}. While these studies primarily focus on natural sciences, time series, and multimodal gaming, the application of surrogate modeling to code execution, where both input and output exist in the modality of language, remains unexplored.

\paragraph{LLMs for Code.} LLMs are widely used in code-related tasks~\citep{lu2021codexglue, luo2023wizardcoder, codeqwen, guo2024deepseek}, which can be fundamentally categorized into code understanding and code generation. Code understanding tasks include code summarization~\cite{hu2018deep, harer2019tree, ahmad2020transformer}, bug detection~\cite{li2018vuldeepecker, russell2018automated, zhou2019devign, chakraborty2020deep}, duplication detection~\cite{zhang2019novel, yu2019neural, wang2020detecting}, code retrieval~\citep{husain2020codesearchnetchallengeevaluatingstate, lu2021codexglue}, etc. Code generation tasks include code completion~\cite{li2017code, parvez2018building}, code repair~\cite{chen2019sequencer, chakraborty2020codit, lutellier2020coconut}, test generation~\citep{Watson_2020, Siddiq_2024, schfer2023empiricalevaluationusinglarge}, etc.
%
%
However, while the potential execution result of code is important for both code understanding and generation, this aspect remains largely unexplored~\citep{weber2024learningcompileprogramsneural}. 
\section{\bench}

\bench assesses the model’s ability to approximate execution results across multiple dimensions, including multi-lingual diversity, repository-level complexity, computational intensity, error handling, and scenario-dependent variability. Below, we describe each component of \bench, including motivation and construction methods.

\begin{table*}[!t]
\centering

\resizebox{\textwidth}{!}{\begin{tabular}{@{}cccccc@{}}
\toprule
Subset & Construction Method & Source & Quantity & Metric & Categories \\ \midrule
ML & Iterative Refactor & McEval~\citep{chai2024mcevalmassivelymultilingualcode} & 150 & Exact Match & Languages \\
CL & Iterative Refactor & GitHub & 150 & Exact Match & Difficulties \& Languages \\
RL & Repository Sampling & GitHub \& Custom & 60 & Mixed & - \\
SC & Manual Implementation & Custom & 150 & Mixed & Scenarios \\
TC & Manual Implementation & Custom & 150 & Mixed & Scenarios, CPU Time \\
BG & Iterative Refactor & DebugBench~\citep{tian2024debugbenchevaluatingdebuggingcapability} & 150 & Jaccard similarity & Error Type \& Languages \\
DR & Manual Implementation & Custom & 200 & Jaccard similarity & Variable Type \\
FL & Inference \& Verification & Lean-Workbook~\citep{ying2024lean} & 150 & Custom & - \\ \bottomrule
\end{tabular}}
\caption{Statistics of \bench, including construction methods, problem sources, quantities, evaluation metrics, and criteria for further classification. In the table, ``custom" refers to customized approaches, and ``mixed" indicates multiple methods, which are elaborated in detail in the text.}
\label{tab:1}
\end{table*}

\subsection{Dataset Construction}
\label{sec:dc}

As illustrated in Figure~\ref{fig:construction}, the construction of \bench involves four distinct methodologies, each applied to specific components of our eight subsets. For \textbf{ML, CL, BG} components, we employ the Iterative Refactor methodology, where we refine the code through an interactive process involving LLM assistance and human verification. The \textbf{RL} component is constructed through Repository Sampling, where we extract and construct test cases from both public and custom code repositories. For \textbf{SC, TC, DR} components, we utilize Manual Implementation, carefully handcrafting code based on selected textbook materials and question collections. Finally, the \textbf{FL} component is developed using the Inference \& Verification approach, leveraging formal mathematical provers to generate proofs and validate them through compiler verification.

To mitigate the risk of data contamination and answer leakage, we implemented a robust two-fold sanitization strategy across the dataset: (1) We applied \textbf{automated filtering} scripts to systematically remove comments from all code snippets before presenting them to the models for evaluation. (2)We \textbf{manually inspected} all generated or collected code to identify and remove any comments, `assert` statements, or other annotations that could inadvertently provide hints or reveal the expected output.

\subsection{Dataset Components}

\paragraph{\underline{M}ulti-\underline{l}ingual Code~(ML).}

A fundamental nature of a general-purpose surrogate executor is its ability to handle multiple programming languages, especially computational languages. Our dataset covers 7 such languages, including \codefont{C}, \codefont{C++}, \codefont{C\#}, \codefont{Java}, \codefont{Rust}, \codefont{Python}, and \codefont{Julia}. Our dataset is adapted from McEval~\citep{chai2024mcevalmassivelymultilingualcode}. The original dataset does not provide executable code, so we used an LLM to generate executable code by providing it with prompts, ground truth, and test cases in the original dataset. This generated code was then manually processed; specifically, we (1) modified code that failed to compile (e.g., by adding missing headers) and (2) carefully reviewed the code to prevent answer leakage through assert statements or comments.

\paragraph{\underline{C}ompetition-\underline{l}evel Code~(CL).}

Next, we consider competition-level code, which presents a higher level of coding difficulty. We collect these tasks from 2 public repositories~\footnote{\url{https://github.com/azl397985856/leetcode}}\footnote{\url{https://gitee.com/shanire/OJCode}}, which contain problems from open coding platforms~(e.g. LeetCode, Luogu). The dataset includes problems in 3 languages, \codefont{C++}, \codefont{Java}, and \codefont{JavaScript}.
Since the original repositories only provide partial solutions, we first use an LLM to generate complete, executable code that prints the expected output. This generated code is then manually verified. During this process, if issues such as package mismatches or syntax errors (e.g., mixing language versions) arose, we manually revised the code to ensure successful compilation and execution. To investigate whether problem difficulty affects the performance of surrogate models, we classify problems into $5$ different difficulty levels.

\paragraph{\underline{R}epository-\underline{l}evel Code~(RL).}

In real-world scenarios, most code exists at the repository level, making repository-level code execution prediction equally important for a general-purpose surrogate model. We manually collect computational repositories that fit within the input length constraints of LLMs. These repositories include tasks such as solving the 24-point problem, Sudoku solving, and converting Python code to LaTeX. These tasks exhibit complex logic but do not rely on sophisticated models or external inputs. To assess the model's ability to understand multi-file structures, we also manually construct two repositories containing advanced C++ syntax and multiple files.

\paragraph{\underline{S}cientific \underline{C}omputing~(SC).}

Scientific computing has long been adopting neural surrogate models. We introduced tasks ranging from solving ordinary differential equations~(ODEs) to optimization problems and signal processing. These tasks are motivated by and widely used in real-world scientific challenges, including areas where increasing research has been done on solving these computational tasks through building efficient surrogate models~\citep{DBLP:journals/natmi/WuZLW23, journals/nature/0004LCXJJ023}.  A comprehensive overview of the setup for each task, along with the corresponding algorithms, can be found in Appendix~\ref{sec:appendix1}.

\paragraph{\underline{T}ime-\underline{C}onsuming Algorithms~(TC).}
Surrogate models were originally motivated by real-world applications where program execution is high-cost and time-consuming. It's a necessity for LLMs to generalize well to strongly computation-power-dependent and time-consuming tasks. We include examples from linear algebra, sorting, searching, Monte Carlo simulations, and string matching programs, ensuring a broad representation of computationally intense tasks. These tasks cover various complexity classes, including P (e.g., sorting an array), NP (e.g., Hamiltonian Cycle), and NP-Hard (e.g., Traveling Salesman's Problem). Additionally, we record the CPU execution time for each program under consistent environments individually to support subsequent studies. A detailed description of this subset is provided in Appendix ~\ref{sec:appendix2}.

\paragraph{\underline{B}u\underline{g}gy Code~(BG).}

Real-world code execution often encounters errors, which pose risks in sensitive scenarios. Therefore, code surrogate models should recognize the presence of bugs. This dataset is adapted from DebugBench~\citep{tian2024debugbenchevaluatingdebuggingcapability}, which extracted Java, Python, and C++ code from LeetCode and manually inserted errors from 4 major bug categories and 18 minor types.
Since DebugBench only provides code snippets rather than complete executable programs, we first used an LLM to automatically complete the error-free code into fully runnable versions. After verifying their correctness, we replaced relevant parts with buggy code and executed them again to capture the corresponding error outputs. Some errors resulted in infinite loops, causing timeouts, so we set a $30$-second execution filter to avoid such cases.

\paragraph{Code with \underline{D}ifferential \underline{R}esults under Different Scenarios~(DR).}

Various contextual factors, such as compiler versions, optimization levels, and language standards, often influence code execution. These variations can lead to different outputs for the same code snippet.
We focus specifically on C++ and manually collect code snippets from textbooks and online sources~\citep{10.5555/1841497, 10.5555/2423877} that exhibit different behaviors under varying compilation settings. We consider multiple compilers (\codefont{g++}, \codefont{clang++}), C++ standards (\codefont{03}, \codefont{11}, \codefont{14}, \codefont{17}), and optimization levels (\codefont{-O0}, \codefont{-O1}, \codefont{-O2}, \codefont{-O3}, \codefont{-Os}). Each snippet is executed across these different settings, and we retain only those that produce varying outputs through different configurations while discarding cases that yield identical results across all settings.

\paragraph{Mathematics \underline{F}ormal \underline{L}anguage~(FL).}

Math-Proving Formal Languages are specialized programming languages designed for mathematical proof verification through compilers~\citep{de2015lean, moura2021lean, paulson1994isabelle, barras1997coq}. These compilers can determine whether a proof is correct and identify specific errors. The verification is very time-consuming.
In this study, we focus on Lean4 which is the most widely used proof language. To build our dataset, we use Goedel-Prover~\citep{lin2025goedelproverfrontiermodelopensource} to conduct large-scale reasoning on Lean-Workbook~\citep{ying2024lean} and extract an equal proportion of correct and incorrect proofs. This balanced dataset allows us to evaluate the surrogate model's ability to assess proof validity effectively.

\subsection{Evaluation Metrics}

We design different evaluation metrics tailored to each subset of \bench to ensure accurate evaluation. 

In \textbf{ML} and \textbf{CL}, the outputs are simple numerical values or formatted strings, we employ exact string matching to measure correctness.

For \textbf{RL}, we employ different evaluation methods for different tasks. For structured C repositories, we use exact character matching to compare outputs. For Sudoku and 24-point problems, we use edit distance to compare results. For other types of repositories, we apply the Ratcliff/Obershelp~\citep{ratcliff1988pattern} algorithm.

For \textbf{SC} and \textbf{TC}, various tasks necessitate distinct evaluation methods. Specifically, (1) numerical simulations are evaluated using the average Relative Absolute Error (RAE); (2) position-based tasks, such as binary search, are assessed through exact string matching; and (3) sorting tasks are evaluated by the rank correlation coefficient~\citep{spearman1904proof}. Details regarding the evaluation metrics can be found in Appendix~\ref{app:metric} and Appendix~\ref{app:metric2}.

For \textbf{BG}, we use the Jaccard similarity~\citep{jaccard1901etude} between predicted and ground truth error messages. For \textbf{DR}, since the same code can produce different outputs in varying settings, which sometimes include warnings or errors, we again utilize Jaccard similarity.

For \textbf{FL}, the results consist of two parts: (1) whether the proof passes or not, and (2) if it fails, we evaluate the accuracy of the error message. The error message consists of a list containing the error locations and descriptions. We compute the score of a prediction as $\frac{1}{N} \sum_{j=1}^{N} \mathbbm{1}[{\hat{p}_j \in P}] \cdot J(\hat{m}_j, m_j)$, where \( N \) is the number of errors in the ground truth, \( P \) is the set of predicted error positions, \( p_j \) represents the \( j \)-th ground truth error position, \( \hat{p}_j \) represents the predicted error position corresponding to \( p_j \), $\mathbbm{1}[{\hat{p}_j \in P}]$ is the indicator function which equals to $1$ only when there exists $\hat{p}_j \in P$, \( m_j \) is the ground truth error message for position \( p_j \), \( \hat{m}_j \) is the predicted error message for position \( \hat{p}_j \), and \( J\) is the Jaccard similarity function.

\begin{table*}[]
\centering

\resizebox{\textwidth}{!}{\begin{tabular}{lcccccccccccccccc}
\toprule
\multirow{2}{*}{\textbf{Model}} & \multicolumn{6}{c}{\textbf{ML}} & \multirow{2}{*}{\textbf{CL}} & \multirow{2}{*}{\textbf{RL}} & \multirow{2}{*}{\textbf{SC}} & \multirow{2}{*}{\textbf{TC}} & \multicolumn{3}{c}{\textbf{BG}} & \multirow{2}{*}{\textbf{DR}} & \multirow{2}{*}{\textbf{FL}} & \multirow{2}{*}{\textbf{Avg.}} \\ \cline{2-7} \cline{12-14}
 & \textbf{CPP} & \textbf{Rust} & \textbf{Python} & \textbf{Julia} & \textbf{Java} & \textbf{Others} &  &  &  &  & \textbf{CPP} & \textbf{Java} & \textbf{Python} &  &  &  \\ \midrule
\multicolumn{17}{c}{\textit{Zero-shot}}                                                                                                                                                           \\ \midrule
\texttt{Claude-3.5-Sonnet} & \cellcolor{cyan!45}$72.73$ & \cellcolor{cyan!34}$55.00$ & \cellcolor{cyan!55}$88.00$ & \cellcolor{cyan!41}$66.67$ & \cellcolor{cyan!48}$76.92$ & \cellcolor{cyan!46}$75.00$ & \cellcolor{cyan!50}$81.58$ & \cellcolor{cyan!35}$57.31$ & \cellcolor{cyan!38}$61.55$ & \cellcolor{cyan!22}$35.27$ & \cellcolor{cyan!5}$9.09$ & \cellcolor{cyan!7}$12.55$ & \cellcolor{cyan!32}$51.40$ & \cellcolor{cyan!8}$12.92$ & \cellcolor{cyan!11}$17.92$ & \cellcolor{cyan!32}$51.59$ \\
\texttt{DeepSeek-V3} & \cellcolor{cyan!34}$54.55$ & \cellcolor{cyan!37}$60.00$ & \cellcolor{cyan!47}$76.00$ & \cellcolor{cyan!38}$61.11$ & \cellcolor{cyan!28}$46.15$ & \cellcolor{cyan!45}$72.50$ & \cellcolor{cyan!35}$56.58$ & \cellcolor{cyan!27}$44.19$ & \cellcolor{cyan!37}$59.65$ & \cellcolor{cyan!22}$35.31$ & \cellcolor{cyan!2}$4.05$ & \cellcolor{cyan!1}$3.03$ & \cellcolor{cyan!13}$21.50$ & \cellcolor{cyan!6}$10.92$ & \cellcolor{cyan!20}$32.46$ & \cellcolor{cyan!26}$42.53$ \\
\texttt{GPT-4o} & \cellcolor{cyan!25}$40.91$ & \cellcolor{cyan!28}$45.00$ & \cellcolor{cyan!37}$60.00$ & \cellcolor{cyan!34}$55.56$ & \cellcolor{cyan!36}$57.69$ & \cellcolor{cyan!34}$55.00$ & \cellcolor{cyan!41}$66.45$ & \cellcolor{cyan!30}$49.13$ & \cellcolor{cyan!33}$53.16$ & \cellcolor{cyan!21}$34.44$ & \cellcolor{cyan!2}$4.28$ & \cellcolor{cyan!4}$7.48$ & \cellcolor{cyan!21}$34.59$ & \cellcolor{cyan!9}$14.75$ & \cellcolor{cyan!13}$21.99$ & \cellcolor{cyan!25}$40.03$ \\
\texttt{Qwen-Max} & \cellcolor{cyan!31}$50.00$ & \cellcolor{cyan!28}$45.00$ & \cellcolor{cyan!27}$44.00$ & \cellcolor{cyan!17}$27.78$ & \cellcolor{cyan!16}$26.92$ & \cellcolor{cyan!31}$50.00$ & \cellcolor{cyan!24}$38.82$ & \cellcolor{cyan!23}$37.15$ & \cellcolor{cyan!35}$56.98$ & \cellcolor{cyan!22}$35.44$ & \cellcolor{cyan!1}$2.82$ & \cellcolor{cyan!2}$3.20$ & \cellcolor{cyan!19}$30.52$ & \cellcolor{cyan!8}$14.03$ & \cellcolor{cyan!18}$29.89$ & \cellcolor{cyan!20}$32.84$ \\
\texttt{Qwen-2.5-7B-Instruct} & \cellcolor{cyan!8}$13.64$ & \cellcolor{cyan!3}$5.00$ & \cellcolor{cyan!7}$12.00$ & \cellcolor{cyan!3}$5.56$ & \cellcolor{cyan!7}$11.54$ & \cellcolor{cyan!10}$17.50$ & \cellcolor{cyan!17}$27.63$ & \cellcolor{cyan!17}$27.98$ & \cellcolor{cyan!14}$22.90$ & \cellcolor{cyan!18}$29.43$ & \cellcolor{cyan!1}$2.21$ & \cellcolor{cyan!2}$3.66$ & \cellcolor{cyan!5}$9.46$ & \cellcolor{cyan!4}$7.24$ & \cellcolor{cyan!22}$36.42$ & \cellcolor{cyan!9}$15.48$ \\
\texttt{Qwen-2.5-32B-Instruct} & \cellcolor{cyan!28}$45.45$ & \cellcolor{cyan!12}$20.00$ & \cellcolor{cyan!30}$48.00$ & \cellcolor{cyan!20}$33.33$ & \cellcolor{cyan!26}$42.31$ & \cellcolor{cyan!12}$20.00$ & \cellcolor{cyan!31}$50.66$ & \cellcolor{cyan!14}$22.61$ & \cellcolor{cyan!20}$32.50$ & \cellcolor{cyan!19}$30.99$ & \cellcolor{cyan!0}$1.21$ & \cellcolor{cyan!1}$2.09$ & \cellcolor{cyan!7}$12.16$ & \cellcolor{cyan!4}$7.26$ & \cellcolor{cyan!4}$7.65$ & \cellcolor{cyan!15}$25.08$ \\
\texttt{Qwen-2.5-Coder-7B-Instruct} & \cellcolor{cyan!28}$45.45$ & \cellcolor{cyan!18}$30.00$ & \cellcolor{cyan!32}$52.00$ & \cellcolor{cyan!27}$44.44$ & \cellcolor{cyan!31}$50.00$ & \cellcolor{cyan!26}$42.50$ & \cellcolor{cyan!46}$75.00$ & \cellcolor{cyan!13}$20.86$ & \cellcolor{cyan!28}$45.50$ & \cellcolor{cyan!17}$28.00$ & \cellcolor{cyan!0}$0.86$ & \cellcolor{cyan!2}$3.47$ & \cellcolor{cyan!8}$13.53$ & \cellcolor{cyan!8}$14.23$ & \cellcolor{cyan!25}$40.40$ & \cellcolor{cyan!21}$33.75$ \\
\texttt{Qwen-2.5-Coder-32B-Instruct} & \cellcolor{cyan!36}$59.09$ & \cellcolor{cyan!34}$55.00$ & \cellcolor{cyan!37}$60.00$ & \cellcolor{cyan!27}$44.44$ & \cellcolor{cyan!43}$69.23$ & \cellcolor{cyan!37}$60.00$ & \cellcolor{cyan!50}$80.26$ & \cellcolor{cyan!30}$48.43$ & \cellcolor{cyan!35}$57.18$ & \cellcolor{cyan!11}$18.84$ & \cellcolor{cyan!0}$1.49$ & \cellcolor{cyan!1}$1.95$ & \cellcolor{cyan!11}$17.96$ & \cellcolor{cyan!8}$13.42$ & \cellcolor{cyan!18}$29.19$ & \cellcolor{cyan!25}$41.10$ \\
\texttt{LLaMA-3.1-8B-Instruct} & \cellcolor{cyan!0}$0.00$ & \cellcolor{cyan!0}$0.00$ & \cellcolor{cyan!2}$4.00$ & \cellcolor{cyan!3}$5.56$ & \cellcolor{cyan!9}$15.38$ & \cellcolor{cyan!3}$5.00$ & \cellcolor{cyan!8}$13.16$ & \cellcolor{cyan!12}$20.41$ & \cellcolor{cyan!2}$4.41$ & \cellcolor{cyan!9}$15.46$ & \cellcolor{cyan!2}$4.41$ & \cellcolor{cyan!2}$4.12$ & \cellcolor{cyan!3}$5.98$ & \cellcolor{cyan!2}$3.97$ & \cellcolor{cyan!0}$0.00$ & \cellcolor{cyan!5}$8.49$ \\
\texttt{LLaMA-3.1-70B-Instruct} & \cellcolor{cyan!34}$54.55$ & \cellcolor{cyan!25}$40.00$ & \cellcolor{cyan!32}$52.00$ & \cellcolor{cyan!20}$33.33$ & \cellcolor{cyan!40}$65.38$ & \cellcolor{cyan!29}$47.50$ & \cellcolor{cyan!48}$78.29$ & \cellcolor{cyan!19}$31.60$ & \cellcolor{cyan!30}$48.65$ & \cellcolor{cyan!18}$30.19$ & \cellcolor{cyan!0}$1.59$ & \cellcolor{cyan!2}$4.18$ & \cellcolor{cyan!8}$14.27$ & \cellcolor{cyan!9}$15.73$ & \cellcolor{cyan!24}$39.16$ & \cellcolor{cyan!23}$37.10$ \\
\midrule
\multicolumn{17}{c}{\textit{Zero-shot Chain-of-Thought}}                                                                                                                                      \\ \midrule
\texttt{Claude-3.5-Sonnet} & \cellcolor{cyan!56}$90.91$ & \cellcolor{cyan!40}$65.00$ & \cellcolor{cyan!60}$96.00$ & \cellcolor{cyan!48}$77.78$ & \cellcolor{cyan!43}$69.23$ & \cellcolor{cyan!57}$92.50$ & \cellcolor{cyan!51}$82.24$ & \cellcolor{cyan!38}$62.31$ & \cellcolor{cyan!39}$63.38$ & \cellcolor{cyan!25}$40.70$ & \cellcolor{cyan!10}$16.91$ & \cellcolor{cyan!12}$20.69$ & \cellcolor{cyan!38}$62.23$ & \cellcolor{cyan!11}$18.19$ & \cellcolor{cyan!21}$33.98$ & \cellcolor{cyan!37}$59.47$ \\
\texttt{DeepSeek-V3} & \cellcolor{cyan!51}$81.82$ & \cellcolor{cyan!53}$85.00$ & \cellcolor{cyan!55}$88.00$ & \cellcolor{cyan!45}$72.22$ & \cellcolor{cyan!43}$69.23$ & \cellcolor{cyan!53}$85.00$ & \cellcolor{cyan!47}$76.32$ & \cellcolor{cyan!39}$62.70$ & \cellcolor{cyan!35}$57.57$ & \cellcolor{cyan!22}$36.71$ & \cellcolor{cyan!2}$4.45$ & \cellcolor{cyan!4}$7.85$ & \cellcolor{cyan!28}$46.26$ & \cellcolor{cyan!10}$16.21$ & \cellcolor{cyan!21}$35.19$ & \cellcolor{cyan!34}$54.97$ \\
\texttt{GPT-4o} & \cellcolor{cyan!42}$68.18$ & \cellcolor{cyan!40}$65.00$ & \cellcolor{cyan!57}$92.00$ & \cellcolor{cyan!45}$72.22$ & \cellcolor{cyan!48}$76.92$ & \cellcolor{cyan!48}$77.50$ & \cellcolor{cyan!49}$79.61$ & \cellcolor{cyan!33}$53.74$ & \cellcolor{cyan!30}$48.56$ & \cellcolor{cyan!17}$28.36$ & \cellcolor{cyan!5}$8.19$ & \cellcolor{cyan!6}$9.97$ & \cellcolor{cyan!27}$44.29$ & \cellcolor{cyan!8}$14.21$ & \cellcolor{cyan!17}$27.91$ & \cellcolor{cyan!31}$51.11$ \\
\texttt{Qwen-Max} & \cellcolor{cyan!53}$86.36$ & \cellcolor{cyan!46}$75.00$ & \cellcolor{cyan!50}$80.00$ & \cellcolor{cyan!45}$72.22$ & \cellcolor{cyan!48}$76.92$ & \cellcolor{cyan!50}$80.00$ & \cellcolor{cyan!44}$71.05$ & \cellcolor{cyan!31}$50.49$ & \cellcolor{cyan!38}$61.78$ & \cellcolor{cyan!22}$36.71$ & \cellcolor{cyan!1}$2.65$ & \cellcolor{cyan!4}$7.73$ & \cellcolor{cyan!29}$46.85$ & \cellcolor{cyan!10}$16.16$ & \cellcolor{cyan!12}$20.74$ & \cellcolor{cyan!32}$52.31$ \\
\texttt{Qwen-2.5-7B-Instruct} & \cellcolor{cyan!25}$40.91$ & \cellcolor{cyan!9}$15.00$ & \cellcolor{cyan!20}$32.00$ & \cellcolor{cyan!20}$33.33$ & \cellcolor{cyan!16}$26.92$ & \cellcolor{cyan!29}$47.50$ & \cellcolor{cyan!32}$52.63$ & \cellcolor{cyan!17}$27.51$ & \cellcolor{cyan!16}$25.68$ & \cellcolor{cyan!18}$28.95$ & \cellcolor{cyan!0}$1.12$ & \cellcolor{cyan!2}$3.76$ & \cellcolor{cyan!9}$14.94$ & \cellcolor{cyan!5}$9.41$ & \cellcolor{cyan!22}$36.46$ & \cellcolor{cyan!16}$26.41$ \\
\texttt{Qwen-2.5-32B-Instruct} & \cellcolor{cyan!31}$50.00$ & \cellcolor{cyan!25}$40.00$ & \cellcolor{cyan!25}$40.00$ & \cellcolor{cyan!34}$55.56$ & \cellcolor{cyan!24}$38.46$ & \cellcolor{cyan!35}$57.50$ & \cellcolor{cyan!33}$53.29$ & \cellcolor{cyan!20}$32.71$ & \cellcolor{cyan!27}$43.29$ & \cellcolor{cyan!19}$30.59$ & \cellcolor{cyan!1}$3.10$ & \cellcolor{cyan!4}$6.96$ & \cellcolor{cyan!14}$23.86$ & \cellcolor{cyan!7}$11.49$ & \cellcolor{cyan!4}$7.39$ & \cellcolor{cyan!20}$32.95$ \\
\texttt{Qwen-2.5-Coder-7B-Instruct} & \cellcolor{cyan!42}$68.18$ & \cellcolor{cyan!25}$40.00$ & \cellcolor{cyan!25}$40.00$ & \cellcolor{cyan!24}$38.89$ & \cellcolor{cyan!33}$53.85$ & \cellcolor{cyan!35}$57.50$ & \cellcolor{cyan!28}$46.05$ & \cellcolor{cyan!12}$19.70$ & \cellcolor{cyan!25}$40.91$ & \cellcolor{cyan!18}$30.19$ & \cellcolor{cyan!1}$2.29$ & \cellcolor{cyan!2}$4.71$ & \cellcolor{cyan!7}$12.77$ & \cellcolor{cyan!9}$15.04$ & \cellcolor{cyan!23}$37.12$ & \cellcolor{cyan!21}$33.81$ \\
\texttt{Qwen-2.5-Coder-32B-Instruct} & \cellcolor{cyan!48}$77.27$ & \cellcolor{cyan!40}$65.00$ & \cellcolor{cyan!50}$80.00$ & \cellcolor{cyan!34}$55.56$ & \cellcolor{cyan!45}$73.08$ & \cellcolor{cyan!42}$67.50$ & \cellcolor{cyan!44}$71.71$ & \cellcolor{cyan!34}$54.58$ & \cellcolor{cyan!34}$55.69$ & \cellcolor{cyan!21}$34.36$ & \cellcolor{cyan!1}$2.05$ & \cellcolor{cyan!2}$4.74$ & \cellcolor{cyan!14}$22.43$ & \cellcolor{cyan!11}$17.62$ & \cellcolor{cyan!17}$28.55$ & \cellcolor{cyan!29}$47.34$ \\
\texttt{LLaMA-3.1-8B-Instruct} & \cellcolor{cyan!25}$40.91$ & \cellcolor{cyan!9}$15.00$ & \cellcolor{cyan!15}$24.00$ & \cellcolor{cyan!13}$22.22$ & \cellcolor{cyan!16}$26.92$ & \cellcolor{cyan!18}$30.00$ & \cellcolor{cyan!25}$41.45$ & \cellcolor{cyan!11}$17.87$ & \cellcolor{cyan!20}$32.65$ & \cellcolor{cyan!11}$18.22$ & \cellcolor{cyan!0}$1.52$ & \cellcolor{cyan!2}$4.23$ & \cellcolor{cyan!8}$13.38$ & \cellcolor{cyan!6}$10.12$ & \cellcolor{cyan!0}$0.66$ & \cellcolor{cyan!12}$19.94$ \\
\texttt{LLaMA-3.1-70B-Instruct} & \cellcolor{cyan!36}$59.09$ & \cellcolor{cyan!31}$50.00$ & \cellcolor{cyan!45}$72.00$ & \cellcolor{cyan!38}$61.11$ & \cellcolor{cyan!36}$57.69$ & \cellcolor{cyan!32}$52.50$ & \cellcolor{cyan!36}$58.55$ & \cellcolor{cyan!21}$34.44$ & \cellcolor{cyan!27}$43.93$ & \cellcolor{cyan!18}$29.76$ & \cellcolor{cyan!1}$1.71$ & \cellcolor{cyan!2}$3.49$ & \cellcolor{cyan!9}$15.02$ & \cellcolor{cyan!10}$16.86$ & \cellcolor{cyan!16}$25.85$ & \cellcolor{cyan!24}$38.80$ \\
\midrule
\multicolumn{17}{c}{\textit{Few-shot Chain-of-Thought}}                                                                                                                                        \\ \midrule
\texttt{Claude-3.5-Sonnet} & \cellcolor{cyan!53}$86.36$ & \cellcolor{cyan!43}$70.00$ & \cellcolor{cyan!60}$96.00$ & \cellcolor{cyan!45}$72.22$ & \cellcolor{cyan!40}$65.38$ & \cellcolor{cyan!51}$82.50$ & \cellcolor{cyan!51}$82.24$ & \cellcolor{cyan!44}$70.65$ & \cellcolor{cyan!39}$63.58$ & \cellcolor{cyan!25}$41.00$ & \cellcolor{cyan!13}$22.04$ & \cellcolor{cyan!14}$23.61$ & \cellcolor{cyan!27}$44.15$ & \cellcolor{cyan!16}$25.70$ & \cellcolor{cyan!19}$31.99$ & \cellcolor{cyan!36}$58.49$ \\
\texttt{DeepSeek-V3} & \cellcolor{cyan!56}$90.91$ & \cellcolor{cyan!40}$65.00$ & \cellcolor{cyan!52}$84.00$ & \cellcolor{cyan!48}$77.78$ & \cellcolor{cyan!45}$73.08$ & \cellcolor{cyan!59}$95.00$ & \cellcolor{cyan!50}$80.26$ & \cellcolor{cyan!49}$78.64$ & \cellcolor{cyan!41}$66.00$ & \cellcolor{cyan!24}$38.60$ & \cellcolor{cyan!13}$21.98$ & \cellcolor{cyan!9}$15.14$ & \cellcolor{cyan!25}$40.27$ & \cellcolor{cyan!15}$24.38$ & \cellcolor{cyan!21}$35.17$ & \cellcolor{cyan!36}$59.08$ \\
\texttt{GPT-4o} & \cellcolor{cyan!42}$68.18$ & \cellcolor{cyan!37}$60.00$ & \cellcolor{cyan!55}$88.00$ & \cellcolor{cyan!48}$77.78$ & \cellcolor{cyan!45}$73.08$ & \cellcolor{cyan!46}$75.00$ & \cellcolor{cyan!47}$75.66$ & \cellcolor{cyan!48}$76.86$ & \cellcolor{cyan!37}$59.65$ & \cellcolor{cyan!23}$37.12$ & \cellcolor{cyan!8}$12.91$ & \cellcolor{cyan!4}$7.74$ & \cellcolor{cyan!18}$29.52$ & \cellcolor{cyan!13}$22.08$ & \cellcolor{cyan!16}$26.65$ & \cellcolor{cyan!32}$52.68$ \\
\texttt{Qwen-Max} & \cellcolor{cyan!51}$81.82$ & \cellcolor{cyan!43}$70.00$ & \cellcolor{cyan!55}$88.00$ & \cellcolor{cyan!48}$77.78$ & \cellcolor{cyan!45}$73.08$ & \cellcolor{cyan!50}$80.00$ & \cellcolor{cyan!51}$82.24$ & \cellcolor{cyan!45}$72.53$ & \cellcolor{cyan!38}$62.32$ & \cellcolor{cyan!23}$37.88$ & \cellcolor{cyan!12}$19.68$ & \cellcolor{cyan!12}$19.78$ & \cellcolor{cyan!23}$37.57$ & \cellcolor{cyan!14}$23.91$ & \cellcolor{cyan!15}$24.76$ & \cellcolor{cyan!35}$56.76$ \\
\texttt{Qwen-2.5-7B-Instruct} & \cellcolor{cyan!17}$27.27$ & \cellcolor{cyan!15}$25.00$ & \cellcolor{cyan!22}$36.00$ & \cellcolor{cyan!24}$38.89$ & \cellcolor{cyan!16}$26.92$ & \cellcolor{cyan!26}$42.50$ & \cellcolor{cyan!30}$48.68$ & \cellcolor{cyan!28}$45.19$ & \cellcolor{cyan!27}$43.42$ & \cellcolor{cyan!18}$28.97$ & \cellcolor{cyan!3}$4.92$ & \cellcolor{cyan!2}$4.70$ & \cellcolor{cyan!8}$12.94$ & \cellcolor{cyan!6}$10.66$ & \cellcolor{cyan!21}$34.53$ & \cellcolor{cyan!17}$28.71$ \\
\texttt{Qwen-2.5-32B-Instruct} & \cellcolor{cyan!36}$59.09$ & \cellcolor{cyan!34}$55.00$ & \cellcolor{cyan!32}$52.00$ & \cellcolor{cyan!41}$66.67$ & \cellcolor{cyan!33}$53.85$ & \cellcolor{cyan!37}$60.00$ & \cellcolor{cyan!39}$63.16$ & \cellcolor{cyan!40}$64.10$ & \cellcolor{cyan!39}$63.12$ & \cellcolor{cyan!20}$32.53$ & \cellcolor{cyan!3}$5.41$ & \cellcolor{cyan!4}$6.94$ & \cellcolor{cyan!17}$28.66$ & \cellcolor{cyan!8}$13.81$ & \cellcolor{cyan!14}$22.87$ & \cellcolor{cyan!26}$43.15$ \\
\texttt{Qwen-2.5-Coder-7B-Instruct} & \cellcolor{cyan!34}$54.55$ & \cellcolor{cyan!18}$30.00$ & \cellcolor{cyan!22}$36.00$ & \cellcolor{cyan!27}$44.44$ & \cellcolor{cyan!31}$50.00$ & \cellcolor{cyan!26}$42.50$ & \cellcolor{cyan!36}$58.55$ & \cellcolor{cyan!31}$50.48$ & \cellcolor{cyan!33}$53.91$ & \cellcolor{cyan!19}$30.69$ & \cellcolor{cyan!2}$3.90$ & \cellcolor{cyan!3}$4.93$ & \cellcolor{cyan!9}$14.44$ & \cellcolor{cyan!8}$14.02$ & \cellcolor{cyan!15}$25.25$ & \cellcolor{cyan!21}$34.24$ \\
\texttt{Qwen-2.5-Coder-32B-Instruct} & \cellcolor{cyan!42}$68.18$ & \cellcolor{cyan!46}$75.00$ & \cellcolor{cyan!45}$72.00$ & \cellcolor{cyan!34}$55.56$ & \cellcolor{cyan!40}$65.38$ & \cellcolor{cyan!43}$70.00$ & \cellcolor{cyan!48}$76.97$ & \cellcolor{cyan!40}$64.16$ & \cellcolor{cyan!35}$56.37$ & \cellcolor{cyan!21}$34.34$ & \cellcolor{cyan!2}$3.93$ & \cellcolor{cyan!4}$6.49$ & \cellcolor{cyan!12}$20.22$ & \cellcolor{cyan!11}$19.09$ & \cellcolor{cyan!14}$22.57$ & \cellcolor{cyan!29}$47.35$ \\
\texttt{LLaMA-3.1-8B-Instruct} & \cellcolor{cyan!8}$13.64$ & \cellcolor{cyan!12}$20.00$ & \cellcolor{cyan!12}$20.00$ & \cellcolor{cyan!3}$5.56$ & \cellcolor{cyan!16}$26.92$ & \cellcolor{cyan!14}$22.50$ & \cellcolor{cyan!18}$30.26$ & \cellcolor{cyan!21}$34.15$ & \cellcolor{cyan!29}$47.89$ & \cellcolor{cyan!13}$22.27$ & \cellcolor{cyan!2}$4.44$ & \cellcolor{cyan!2}$4.55$ & \cellcolor{cyan!6}$9.66$ & \cellcolor{cyan!7}$12.05$ & \cellcolor{cyan!8}$13.25$ & \cellcolor{cyan!11}$19.14$ \\
\texttt{LLaMA-3.1-70B-Instruct} & \cellcolor{cyan!34}$54.55$ & \cellcolor{cyan!21}$35.00$ & \cellcolor{cyan!25}$40.00$ & \cellcolor{cyan!13}$22.22$ & \cellcolor{cyan!33}$53.85$ & \cellcolor{cyan!21}$35.00$ & \cellcolor{cyan!42}$68.42$ & \cellcolor{cyan!31}$50.83$ & \cellcolor{cyan!38}$60.96$ & \cellcolor{cyan!18}$30.29$ & \cellcolor{cyan!5}$8.00$ & \cellcolor{cyan!3}$5.95$ & \cellcolor{cyan!11}$18.32$ & \cellcolor{cyan!8}$13.27$ & \cellcolor{cyan!20}$33.12$ & \cellcolor{cyan!22}$35.32$ \\
\bottomrule
\end{tabular}}
\caption{Performance of different models under different prompting strategies on \bench.}
\label{tab:main}
\end{table*}

\subsection{Dataset Statistics}

Table \ref{tab:1} presents detailed statistics of \bench, including the construction methods, problem sources, dataset quantities, number of examples in few-shot scenarios, evaluation metrics, and classification criteria for each subset.
\section{Experiments}

\subsection{Setup}

\paragraph{Models.} We tested \bench on $17$ open-source and $4$ closed-source models of different sizes, including both chat models and code models. The closed-source models include \codefont{GPT-4o}~(2024-08-06)~\citep{openai2024gpt4o}, \codefont{GPT-4o-mini}~(2024-07-18)~\citep{openai2024gpt4omini}, \codefont{Claude-3.5-Sonnet}~(2024-10-22)~\citep{anthropic2024claude35}, and \codefont{Qwen-Max}~(2025-01-25)~\citep{qwen25}. The open-source models include \codefont{LLaMA-3.1-\{8, 70\}B-Instruct}, \codefont{LLaMA-3.3-70B-Instruct}, \codefont{Qwen-2.5-\{0.5, 1.5, 3, 7, 14, 32, 72\}B-Instruct}, \codefont{Qwen-2.5-Coder-\{0.5, 1.5, 3, 7, 14, 32\}B-Instruct} and \codefont{DeepSeek-V3}~(671B).

\paragraph{Settings.} We tested the above models on \bench under 3 settings: 0-shot w/o CoT, 0-shot w/ CoT, and few-shot w/ CoT. CoT here means whether we use Chain-of-Thought~\citep{wei2022chain} prompting, allowing the models to think step by step, or ask the models to answer directly. We set the temperature to $0$, i.e., employing greedy decoding.

\begin{table*}[!t]

\resizebox{\textwidth}{!}{
\begin{tabular}{lcccccccccc}
\toprule
\multirow{2}{*}{Model} & \multicolumn{3}{c}{Compiler} & \multicolumn{3}{c}{Standard} & \multicolumn{3}{c}{Optimization} \\
\cmidrule(lr){2-4} \cmidrule(lr){5-7} \cmidrule(lr){8-10}
& Zero-shot & Zero-shot CoT & Few-shot CoT & Zero-shot & Zero-shot CoT & Few-shot CoT & Zero-shot & Zero-shot CoT & Few-shot CoT \\
\midrule
\texttt{Claude-3.5} & 12.92 & 18.19 & 25.70 & 14.21 & 16.75 & 23.59 & 16.06 & 1.23 & 12.90 \\
\texttt{GPT-4o} & 14.75 & 14.21 & 22.08 & 15.33 & 14.61 & 20.02 & 7.66 & 1.63 & 1.22 \\
\texttt{LLaMA-3.1-8B} & 3.97 & 10.12 & 12.05 & 4.29 & 10.60 & 13.17 & 1.96 & 3.91 & 8.04 \\
\texttt{LLaMA-3.1-70B} & 15.73 & 16.86 & 13.27 & 16.36 & 16.27 & 13.24 & 15.04 & 2.24 & 3.73 \\
\bottomrule
\end{tabular}}

\caption{Model Performance Across Different Variables in DR}
\label{tab:dr}

\end{table*}

\subsection{Results}

Table~\ref{tab:main} presents the performance of 10 selected models across 8 sub-datasets of \bench under 3 different settings. In this table, we provide a detailed breakdown of the models' performance on the ML and BG sub-datasets. We present the complete results of all 21 models in Table~\ref{tab:main2} in Appendix~\ref{app:main2}. From the results, several notable findings emerge:

\textbf{\bench demonstrates strong discriminative ability}. Even the strongest models perform only moderately well, highlighting the value of our benchmark. The models exhibit significant performance differences across different subsets, reflecting the comprehensiveness of our dataset. Additionally, models that perform well in other tasks, such as \texttt{Claude-3.5-Sonnet}, also achieve substantial overall results on \bench. This demonstrates the reasonableness of our dataset and its effectiveness in benchmarking LLMs as general-purpose surrogate code executors.

\textbf{Different prompting strategies lead to varying model performance and have different effects across subsets.} We found that for the task of code execution surrogacy, both Chain-of-Thought prompting and few-shot learning can enhance model performance.

\textbf{Larger model sizes tend to yield better performance on \bench.} From the results, we observe that regardless of whether it is a Qwen-Chat model or a Qwen-Coder model, performance improves as the parameter size increases.

\textbf{Chat models and coder models exhibit different performance patterns and are affected differently by prompting strategies}. We observed that for chat and code models of the same size, code models outperform chat models in the zero-shot setting \bench. However, in the other two settings, chat models perform better. This suggests that code models have stronger zero-shot surrogate capabilities, whereas chat models excel in reasoning and imitation abilities.

\textbf{On some sub-datasets of \bench, stronger models perform worse than smaller, weaker ones.} For example, in the FL dataset, we found that this occurs because stronger models tend to actively look for errors in the code, often misidentifying correct code as incorrect. In contrast, smaller models are more inclined to assume that all code is error-free. Since half of the samples in this subset are indeed correct, the smaller models end up achieving better performance.

\subsection{Anomaly Analysis}
We conducted case studies on anomalous results to better understand model behavior.

\textbf{Larger Models Underperforming.} In some instances, larger models performed worse than their smaller counterparts. For example, on ML tasks, \texttt{Qwen-2.5-14B-Instruct} was outperformed by \texttt{Qwen-2.5-0.5B-Instruct}. Our case study revealed that while the 0.5B model's reasoning was less detailed, it correctly grasped the code's core logic. In contrast, the 14B model attempted more elaborate reasoning but often made critical misjudgments (e.g., in conditionals like $\leq$), leading to incorrect results. This suggests that on certain complex reasoning tasks, larger models may be prone to "overthinking" or following incorrect logical paths.

\textbf{Zero-Shot Failures.} We investigated cases of complete failure in the zero-shot setting, such as \texttt{LLaMA-3.1-8B-Instruct} scoring 0 on ML tasks for C++ and Rust, and on all FL tasks. For ML, the failures were attributed to the inherent difficulty of the test cases and specific language features (e.g., Rust's strict type system). For FL tasks, the model incorrectly identified errors in all problems. For half of the correct problems, it wrongly predicted errors. For the other half that contained errors, it failed to predict the specific error messages accurately, leading to incorrect predictions in all cases.

\section{Analysis}

\subsection{The Impact of Language Type}

In Table~\ref{tab:main}, we compare model performance across different programming languages.  

In the ML subset, LLMs perform best in Python, followed by C++. Python’s simple syntax makes it easier to process, while C++ benefits from its strong presence in programming and system-level tasks. Models also perform well in Julia, likely due to its clean syntax and similarity to Python. However, performance drops significantly in Rust, where strict ownership and lifetime rules introduce complexity, making it hard to predict.  

In the BG subset, when predicting the output of buggy code, models excel in Python but struggle with C++. Python’s clear error messages aid prediction, whereas C++'s static typing, manual memory management, and potential for undefined behavior make error handling more difficult.

\begin{table}[!h]
\centering
\resizebox{0.48\textwidth}{!}{
\begin{tabular}{lccccc}
\toprule
Difficulty & 1 & 2 & 3 & 4 & 5 \\
\midrule
\multicolumn{6}{c}{\textit{Zero-shot}} \\ \midrule
\texttt{Claude-3.5-Sonnet} & $90.91$ & $81.25$ & $82.05$ & $72.00$ & $79.49$ \\
\texttt{GPT-4o} & $72.73$ & $56.25$ & $58.97$ & $84.00$ & $61.54$ \\
\texttt{LLaMA-3.1-8B-Instruct} & $24.24$ & $18.75$ & $7.69$ & $4.00$ & $12.82$ \\
\texttt{LLaMA-3.1-70B-Instruct} & $96.97$ & $93.75$ & $64.10$ & $64.00$ & $79.49$ \\
\midrule
\multicolumn{6}{c}{\textit{Zero-shot  Chain-of-Thought}} \\ \midrule
\texttt{Claude-3.5-Sonnet} & $96.97$ & $87.50$ & $76.92$ & $72.00$ & $79.49$ \\
\texttt{GPT-4o} & $96.97$ & $81.25$ & $74.36$ & $80.00$ & $69.23$ \\
\texttt{LLaMA-3.1-8B-Instruct} & $57.58$ & $43.75$ & $30.77$ & $40.00$ & $38.46$ \\
\texttt{LLaMA-3.1-70B-Instruct} & $90.91$ & $56.25$ & $48.72$ & $36.00$ & $56.41$ \\
\midrule
\multicolumn{6}{c}{\textit{Few-shot Chain-of-Thought}} \\ \midrule
\texttt{Claude-3.5-Sonnet} & $96.97$ & $81.25$ & $76.92$ & $68.00$ & $84.62$ \\
\texttt{GPT-4o} & $93.94$ & $81.25$ & $71.79$ & $60.00$ & $71.79$ \\
\texttt{LLaMA-3.1-8B-Instruct} & $39.39$ & $18.75$ & $33.33$ & $16.00$ & $33.33$ \\
\texttt{LLaMA-3.1-70B-Instruct} & $87.88$ & $87.50$ & $48.72$ & $48.00$ & $76.92$ \\
\bottomrule
\end{tabular}}
\caption{Model Performance by Difficulty Level}
\label{tab:cl}
\end{table}

\subsection{The Impact of Problem Difficulty}

To analyze how problem difficulty affects model performance, we examine results in the CL subset, as shown in Table~\ref{tab:cl}.

Across all settings, model performance generally declines as problem difficulty increases. However, at the highest difficulty level, performance improves slightly. This anomaly arises because difficulty levels are categorized based on coding complexity, but the hardest problems often involve implementing simple functionality using complex, optimized algorithms. In such cases, models can generate correct answers through end-to-end reasoning without fully understanding the underlying code logic. In contrast, levels 1–4 require logical reasoning based on code execution, making prediction harder as complexity increases.

Furthermore, while Chain-of-Thought prompting improves performance, few-shot learning does not and may even degrade results. This is likely because buggy code varies widely, causing few-shot examples to mislead rather than aid the model.

\begin{figure}[!h]
    \centering
    \includegraphics[width=\linewidth]{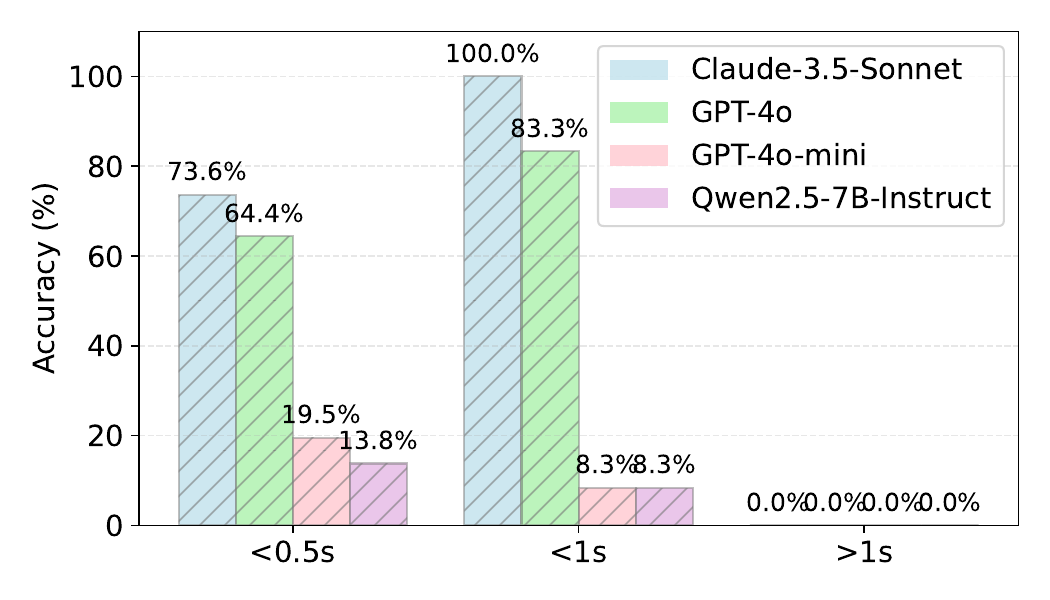}
    \caption{Model's Performance on \textbf{TC} subset across programs with different run time on CPU.}
    \label{fig:time}
\end{figure}

\begin{figure*}[!h]
    \centering
    \includegraphics[width=\linewidth]{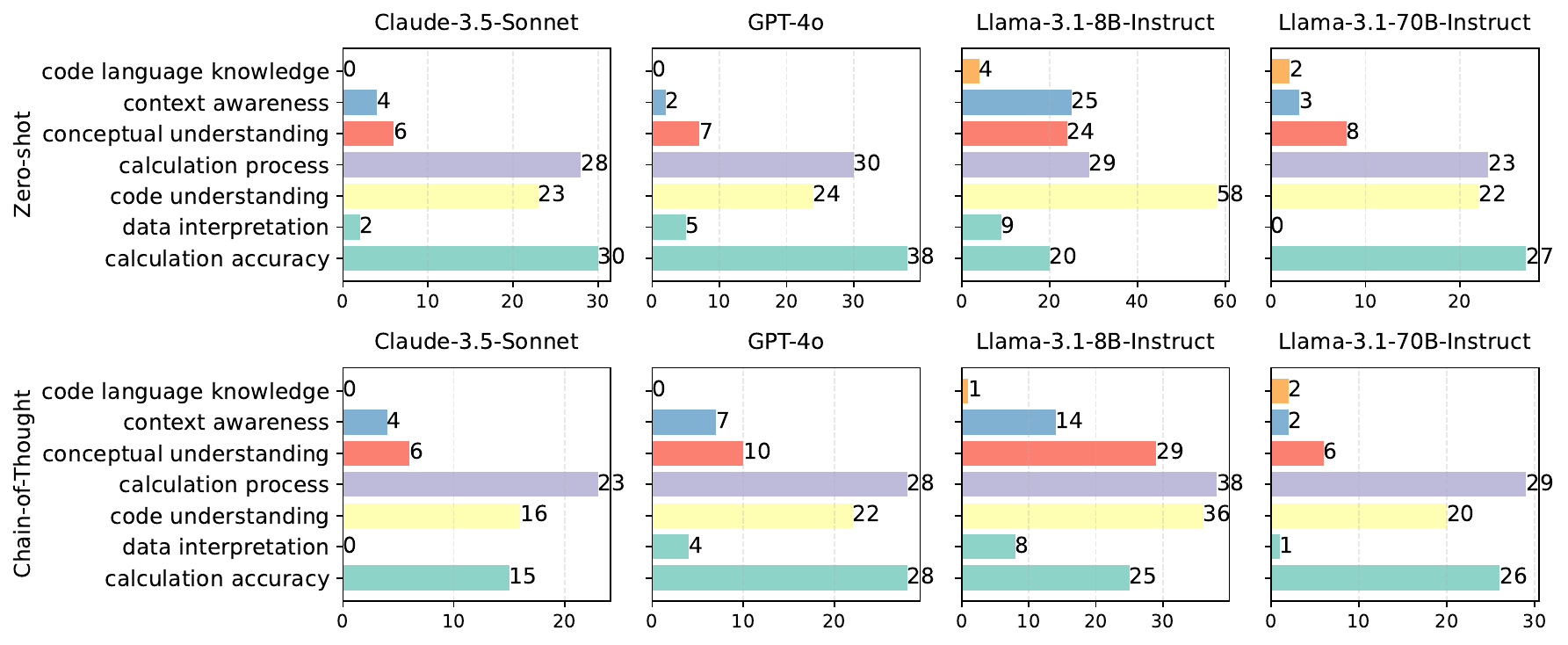}
    \caption{Breakdown of error types across different language models and prompting methods.}
    \label{fig:ea}
\end{figure*}

\subsection{Prediction Accuracy \& CPU Time}

We explore the relationship between the execution time of the given code through running on CPUs and the accuracy of using LLMs as surrogate models to acquire the output. We categorize problems in TC into distinct bins based on their execution times and calculate the average accuracy for each model across all samples within the same bin, as shown in Figure~\ref{fig:time}. We observed the trend of prediction accuracy of the model falling as the actual execution time required for the corresponding program prolonged. It's especially worth noting that for computational tasks that require execution time longer than 1 second, even state-of-the-art models struggle to obtain even one correct answer.

\subsection{The Impact of Variable Type}

The DR subset in \bench examines model performance in predicting code behavior under different environmental factors. Specifically, DR includes variations in C++ compiler version, C++ standard, and compilation optimization settings. Table~\ref{tab:dr} presents model performance across different prompting strategies in DR.

In the zero-shot setting, model accuracy improves sequentially across the three factors, suggesting that compiler version differences are harder to predict than standard variations, which in turn are harder than optimization settings. However, with Chain-of-Thought reasoning, performance declines across all factors, with the sharpest drop for optimization settings. This indicates that while CoT aids reasoning for compiler versions and standards, it adds unnecessary complexity for optimizations, ultimately reducing accuracy.

\subsection{In-depth Look at Model Behavior}

We looked into LLMs' surrogation process and summarized three key observations:

\textbf{Deep Semantic and Logical Reasoning.} LLMs demonstrate a strong ability to interpret complex, language-specific semantics. For instance, when analyzing C++ code with bit fields, models correctly reasoned about the effects of \texttt{signed} vs. \texttt{unsigned} types, bit-width limitations, and how value overflows are handled under two's complement representation. This indicates an understanding of low-level computational logic.

\textbf{Stateful Algorithmic Tracing.} For algorithmic tasks, LLMs effectively simulate the execution process by tracing the program's state. The models' chain-of-thought process shows them iterating through loops, updating variables step-by-step, and applying conditional logic as an interpreter would.

\textbf{Identification of Computational Boundaries.} In a brute-force Traveling Salesman Problem (TSP) solver, models could correctly understand the overall goal and algorithm, but often failed to predict the precise final floating-point value. This suggests their "execution" is a form of sophisticated logical inference, which falters when faced with complex combinatorial calculations that exceed their internal reasoning capacity.

\subsection{Error Analysis}

To further understand the model's performance and limitations regarding serving as general surrogate models, we categorize the Errors made by \codefont{Claude-3.5-Sonnet}, \codefont{GPT-4o}, and \codefont{Llama-3.1-8B-Instruct} and \codefont{Llama-3.1-70B-Instruct}in the \textbf{CL} subset of \bench.

We first combined LLM-assisted annotation and manual verification to identify $7$ most typical error types: 1. Code Language Knowledge: evaluating foundational programming language proficiency, 2. Context Awareness: measuring understanding of long text and repository-level code, 3. Conceptual Understanding: assessing comprehension of programming concepts, 4. Calculation Process: verifying computational step accuracy, 5. Code Understanding: testing comprehension of code logic and structure, 6. Data Interpretation: evaluating data processing and analysis capabilities, and 7. Calculation Accuracy: Measuring precision in scientific computations. We then use LLM to label errors in the model's responses. The categorized error statistics are demonstrated in Figure ~\ref{fig:ea}.

As models prompted with CoT exhibit fewer instances of most error types compared to zero-shot, especially for the Code Understanding capability of Llama. The main error types are Code Understanding, Calculation Process, and Calculation Accuracy. For zero-shot, the primary error is accuracy, but for CoT, the most frequent error is Calculation Process. This suggests that CoT can better grasp the overall code logic and produce more correct results, but it may still make mistakes in the chain of thought process. In general, CoT has fewer and smaller errors. From the model perspective, Llama has a clear lead in Conceptual Understanding errors, indicating its weaker ability to understand concepts.

\subsection{Training Scale Analysis}

We also investigate whether training scaling can affect LLMs' surrogate execution capabilities on the FL task. We trained models of varying sizes on different amounts of training data sampled from the same distribution, and tested them to predict the error feedbacks for incorrect proofs.

As demonstrated in Figure~\ref{fig:case2}, both model size and training steps are crucial factors in determining surrogate execution accuracy. As we scale from 0.5B to 7B parameters, models consistently show improved learning efficiency and higher performance ceilings throughout the training process. Larger models learn faster in the early stages and continue to improve for longer before plateauing, suggesting better utilization of the training data.

These empirical observations align with established scaling laws in language modeling, indicating that surrogate execution capabilities follow similar scaling patterns as other language tasks.
\section{Conclusion}
We introduce \bench, a holistic benchmark for evaluating LLMs as general-purpose surrogate code executors. Through extensive empirical study, we argue that there remains significant room for further improvements on grounding LLMs to facilitate general surrogate models. Our findings not only chart the current landscape but also illuminate a clear path for future research.

\section*{Acknowledgements}

This work was independently conducted by the authors and is self-funded by BL without institutional affiliation.

\section*{Limitations}

Despite its comprehensive evaluation, our study has several limitations. LLMs remain approximators rather than exact code executors, often struggling with edge cases, intricate runtime behaviors, and execution-dependent state changes. While SURGE covers diverse execution scenarios, it does not encompass all specialized environments, such as hardware-dependent simulations or real-time systems. Additionally, LLMs may generate plausible but incorrect outputs, particularly in complex logical dependencies or undefined behaviors, making error detection challenging. Our scaling study is constrained by computational resources, limiting the assessment of extremely large models or extensive training data distributions. Furthermore, security risks remain, as LLMs may fail to recognize vulnerabilities, potentially misjudging harmful code. Finally, our benchmark operates in a controlled setting, whereas real-world software development involves dynamic interactions and iterative debugging, which are not fully captured in our study. Future work should focus on improving LLMs' reasoning abilities, enhancing robustness in execution prediction, and integrating them with traditional program analysis techniques for practical deployment.

\bibliography{reference}

@String{Computing = "Computing" }

@String{Computer = "{IEEE} Computer" }

@String{Springer = "Springer-Verlag" }

@misc{husain2020codesearchnetchallengeevaluatingstate,
      title={CodeSearchNet Challenge: Evaluating the State of Semantic Code Search}, 
      author={Hamel Husain and Ho-Hsiang Wu and Tiferet Gazit and Miltiadis Allamanis and Marc Brockschmidt},
      year={2020},
      eprint={1909.09436},
      archivePrefix={arXiv},
      primaryClass={cs.LG},
      url={https://arxiv.org/abs/1909.09436}, 
}

@inproceedings{lu2021codexglue,
title={Code{XGLUE}: A Machine Learning Benchmark Dataset for Code Understanding and Generation},
author={Shuai Lu and Daya Guo and Shuo Ren and Junjie Huang and Alexey Svyatkovskiy and Ambrosio Blanco and Colin Clement and Dawn Drain and Daxin Jiang and Duyu Tang and Ge Li and Lidong Zhou and Linjun Shou and Long Zhou and Michele Tufano and MING GONG and Ming Zhou and Nan Duan and Neel Sundaresan and Shao Kun Deng and Shengyu Fu and Shujie LIU},
booktitle={Thirty-fifth Conference on Neural Information Processing Systems Datasets and Benchmarks Track (Round 1)},
year={2021},
url={https://openreview.net/forum?id=6lE4dQXaUcb}
}

@article{harer2019tree,
  title={Tree-Transformer: A Transformer-Based Method for Correction of Tree-Structured Data},
  author={Jacob Harer and Chris Reale and Peter Chin},
  journal={arXiv preprint arXiv:1908.00449},
  year={2019},
  url={https://arxiv.org/abs/1908.00449},
}

@article{journals/nature/0004LCXJJ023,
  author = {Zhang, Yuchen and Long, Mingsheng and Chen, Kaiyuan and Xing, Lanxiang and Jin, Ronghua and Jordan, Michael I. and Wang, Jianmin},
  ee = {https://www.wikidata.org/entity/Q130467973},
  journal = {Nat.},
  number = 7970,
  pages = {526-532},
  title = {Skilful nowcasting of extreme precipitation with NowcastNet.},
  url = {http://dblp.uni-trier.de/db/journals/nature/nature619.html#0004LCXJJ023},
  volume = 619,
  year = 2023
}

@article{DBLP:journals/natmi/WuZLW23,
  author={Haixu Wu and Hang Zhou and Mingsheng Long and Jianmin Wang},
  title={Interpretable weather forecasting for worldwide stations with a unified deep model},
  year={2023},
  month={June},
  cdate={1685577600000},
  journal={Nat. Mac. Intell.},
  volume={5},
  number={6},
  pages={602-611},
  url={https://doi.org/10.1038/s42256-023-00667-9}
}

@article{spearman1904proof,
  author = {Spearman, C.},
  journal = {The American Journal of Psychology},
  number = 1,
  pages = {72-101},
  publisher = {University of Illinois Press},
  title = {The Proof and Measurement of Association between Two Things},
  url = {http://www.jstor.org/stable/1412159},
  volume = 15,
  year = 1904
}

@inproceedings{lutellier2020coconut,
    title={CoCoNuT: combining context-aware neural translation models using ensemble for program repair},
    author={Lutellier, Thibaud and Pham, Hung Viet and Pang, Lawrence and Li, Yitong and Wei, Moshi and Tan, Lin},
    booktitle={Proceedings of the 29th ACM SIGSOFT International Symposium on Software Testing and Analysis},
    pages={101--114},
    year={2020},
    publisher = {Association for Computing Machinery},
    address = {New York, NY, USA},
    url = {https://doi.org/10.1145/3395363.3397369},
    doi = {10.1145/3395363.3397369},
}

@article{chen2019sequencer,
  title={Sequencer: Sequence-to-sequence learning for end-to-end program repair},
  author={Chen, Zimin and Kommrusch, Steve James and Tufano, Michele and Pouchet, Louis-No{\"e}l and Poshyvanyk, Denys and Monperrus, Martin},
  journal={IEEE Transactions on Software Engineering},
  year={2019},
  publisher={IEEE},
  url = {https://doi.org/10.1109/TSE.2019.2940179},
  doi = {10.1109/TSE.2019.2940179},
}

@inproceedings{russell2018automated,
  title={Automated vulnerability detection in source code using deep representation learning},
  author={Russell, Rebecca and Kim, Louis and Hamilton, Lei and Lazovich, Tomo and Harer, Jacob and Ozdemir, Onur and Ellingwood, Paul and McConley, Marc},
  booktitle={2018 17th IEEE International Conference on Machine Learning and Applications (ICMLA)},
  pages={757--762},
  year={2018},
  organization={IEEE},
  doi={10.1109/ICMLA.2018.00120},
  url={https://doi.org/10.1109/ICMLA.2018.00120},
}

@article{li2018vuldeepecker,
  title={Vuldeepecker: A deep learning-based system for vulnerability detection},
  author={Li, Zhen and Zou, Deqing and Xu, Shouhuai and Ou, Xinyu and Jin, Hai and Wang, Sujuan and Deng, Zhijun and Zhong, Yuyi},
  journal={arXiv preprint arXiv:1801.01681},
  year={2018},
  url={https://arxiv.org/abs/1801.01681},
}

@ARTICLE{chakraborty2020codit,
  author={Saikat {Chakraborty} and Yangruibo {Ding} and Miltiadis {Allamanis} and Baishakhi {Ray}},
  journal={IEEE Transactions on Software Engineering}, 
  title={CODIT: Code Editing with Tree-Based Neural Models}, 
  year={2020},
  volume={},
  number={},
  pages={1-1},
  doi={10.1109/TSE.2020.3020502},
  url={https://arxiv.org/abs/1810.00314},
}

@article{chakraborty2020deep,
  title={Deep Learning based Vulnerability Detection: Are We There Yet?},
  author={Chakraborty, Saikat and Krishna, Rahul and Ding, Yangruibo and Ray, Baishakhi},
  journal={arXiv preprint arXiv:2009.07235},
  year={2020},
  url={https://arxiv.org/abs/2009.07235},
}

@inproceedings{hu2018deep,
  author = {Hu, Xing and Li, Ge and Xia, Xin and Lo, David and Jin, Zhi},
    title = {Deep Code Comment Generation},
    year = {2018},
    isbn = {9781450357142},
    publisher = {Association for Computing Machinery},
    address = {New York, NY, USA},
    url = {https://doi.org/10.1145/3196321.3196334},
    doi = {10.1145/3196321.3196334},
    booktitle = {Proceedings of the 26th Conference on Program Comprehension},
    pages = {200–210},
}

@article{CodeXGLUE,
  title={CodeXGLUE: A Machine Learning Benchmark Dataset for Code Understanding and Generation},
  author={Lu, Shuai and Guo, Daya and Ren, Shuo and Huang, Junjie and Svyatkovskiy, Alexey and Blanco, Ambrosio and Clement, Colin and Drain, Dawn and Jiang, Daxin and Tang, Duyu and others},
  journal={arXiv preprint arXiv:2102.04664},
  year={2021},
  url={https://arxiv.org/abs/2102.04664},
}

@inproceedings{li2017code,
  title     = {Code Completion with Neural Attention and Pointer Networks},
  author    = {Jian Li and Yue Wang and Michael R. Lyu and Irwin King},
  booktitle = {Proceedings of the Twenty-Seventh International Joint Conference on
               Artificial Intelligence, {IJCAI-18}},
  publisher = {International Joint Conferences on Artificial Intelligence Organization},  
  pages     = {4159--4165},
  year      = {2018},
  month     = {7},
  doi       = {10.24963/ijcai.2018/578},
  url       = {https://doi.org/10.24963/ijcai.2018/578},
}

@inproceedings{yu2019neural,
  title={Neural detection of semantic code clones via tree-based convolution},
  author={Yu, Hao and Lam, Wing and Chen, Long and Li, Ge and Xie, Tao and Wang, Qianxiang},
  booktitle={2019 IEEE/ACM 27th International Conference on Program Comprehension (ICPC)},
  pages={70--80},
  year={2019},
  publisher = {IEEE Press},
  url = {https://doi.org/10.1109/ICPC.2019.00021},
  doi = {10.1109/ICPC.2019.00021},
}

@inproceedings{wang2020detecting,
  title={Detecting Code Clones with Graph Neural Network and Flow-Augmented Abstract Syntax Tree},
  author={Wang, Wenhan and Li, Ge and Ma, Bo and Xia, Xin and Jin, Zhi},
  booktitle={2020 IEEE 27th International Conference on Software Analysis, Evolution and Reengineering (SANER)},
  pages={261--271},
  year={2020},
  organization={IEEE},
  doi={10.1109/SANER48275.2020.9054857},
  url={https://doi.org/10.1109/SANER48275.2020.9054857},
}

@inproceedings{zhang2019novel,
    author = {Zhang, Jian and Wang, Xu and Zhang, Hongyu and Sun, Hailong and Wang, Kaixuan and Liu, Xudong},
    title = {A Novel Neural Source Code Representation Based on Abstract Syntax Tree},
    year = {2019},
    publisher = {IEEE Press},
    url = {https://doi.org/10.1109/ICSE.2019.00086},
    doi = {10.1109/ICSE.2019.00086},
    booktitle = {Proceedings of the 41st International Conference on Software Engineering},
    pages = {783–794},
    numpages = {12},
    location = {Montreal, Quebec, Canada},
    series = {ICSE ’19}
}

@inproceedings{zhou2019devign,
     author = {Zhou, Yaqin and Liu, Shangqing and Siow, Jingkai and Du, Xiaoning and Liu, Yang},
     booktitle = {Advances in Neural Information Processing Systems},
     editor = {H. Wallach and H. Larochelle and A. Beygelzimer and F. d\textquotesingle Alch\'{e}-Buc and E. Fox and R. Garnett},
     pages = {10197--10207},
     publisher = {Curran Associates, Inc.},
     title = {Devign: Effective Vulnerability Identification by Learning Comprehensive Program Semantics via Graph Neural Networks},
     url = {https://proceedings.neurips.cc/paper/2019/file/49265d2447bc3bbfe9e76306ce40a31f-Paper.pdf},
     volume = {32},
     year = {2019}
}

@inproceedings{Watson_2020, series={ICSE ’20},
   title={On learning meaningful assert statements for unit test cases},
   url={http://dx.doi.org/10.1145/3377811.3380429},
   DOI={10.1145/3377811.3380429},
   booktitle={Proceedings of the ACM/IEEE 42nd International Conference on Software Engineering},
   publisher={ACM},
   author={Watson, Cody and Tufano, Michele and Moran, Kevin and Bavota, Gabriele and Poshyvanyk, Denys},
   year={2020},
   month=jun, collection={ICSE ’20} }

@inproceedings{Siddiq_2024, series={EASE 2024},
   title={Using Large Language Models to Generate JUnit Tests: An Empirical Study},
   url={http://dx.doi.org/10.1145/3661167.3661216},
   DOI={10.1145/3661167.3661216},
   booktitle={Proceedings of the 28th International Conference on Evaluation and Assessment in Software Engineering},
   publisher={ACM},
   author={Siddiq, Mohammed Latif and Da Silva Santos, Joanna Cecilia and Tanvir, Ridwanul Hasan and Ulfat, Noshin and Al Rifat, Fahmid and Carvalho Lopes, Vinícius},
   year={2024},
   month=jun, pages={313–322},
   collection={EASE 2024} }

@misc{schfer2023empiricalevaluationusinglarge,
      title={An Empirical Evaluation of Using Large Language Models for Automated Unit Test Generation}, 
      author={Max Schäfer and Sarah Nadi and Aryaz Eghbali and Frank Tip},
      year={2023},
      eprint={2302.06527},
      archivePrefix={arXiv},
      primaryClass={cs.SE},
      url={https://arxiv.org/abs/2302.06527}, 
}

@article{wei2022chain,
  title={Chain-of-thought prompting elicits reasoning in large language models},
  author={Wei, Jason and Wang, Xuezhi and Schuurmans, Dale and Bosma, Maarten and Xia, Fei and Chi, Ed and Le, Quoc V and Zhou, Denny and others},
  journal={Advances in neural information processing systems},
  volume={35},
  pages={24824--24837},
  year={2022}
}

@article{zheng2023codegeex,
  title={Codegeex: A pre-trained model for code generation with multilingual evaluations on humaneval-x},
  author={Zheng, Qinkai and Xia, Xiao and Zou, Xu and Dong, Yuxiao and Wang, Shan and Xue, Yufei and Wang, Zihan and Shen, Lei and Wang, Andi and Li, Yang and others},
  journal={arXiv preprint arXiv:2303.17568},
  year={2023}
}

@inproceedings{luo2023wizardcoder,
  title={WizardCoder: Empowering Code Large Language Models with Evol-Instruct},
  author={Luo, Ziyang and Xu, Can and Zhao, Pu and Sun, Qingfeng and Geng, Xiubo and Hu, Wenxiang and Tao, Chongyang and Ma, Jing and Lin, Qingwei and Jiang, Daxin},
  booktitle={The Twelfth International Conference on Learning Representations},
  year={2023}
}

@article{bi2024deepseek,
  title={Deepseek llm: Scaling open-source language models with longtermism},
  author={Bi, Xiao and Chen, Deli and Chen, Guanting and Chen, Shanhuang and Dai, Damai and Deng, Chengqi and Ding, Honghui and Dong, Kai and Du, Qiushi and Fu, Zhe and others},
  journal={arXiv preprint arXiv:2401.02954},
  year={2024}
}

@article{guo2024deepseek,
  title={DeepSeek-Coder: When the Large Language Model Meets Programming--The Rise of Code Intelligence},
  author={Guo, Daya and Zhu, Qihao and Yang, Dejian and Xie, Zhenda and Dong, Kai and Zhang, Wentao and Chen, Guanting and Bi, Xiao and Wu, Y and Li, YK and others},
  journal={arXiv preprint arXiv:2401.14196},
  year={2024}
}

@misc{claude3,
  author = {Anthropic},
  title = {{The Claude 3 Model Family: Opus, Sonnet, Haiku}},
  publisher = {Anthropic},
  howpublished = "\url{https://www-cdn.anthropic.com/de8ba9b01c9ab7cbabf5c33b80b7bbc618857627/Model_Card_Claude_3.pdf}",
  year = {2024}
}

@misc{llama3,
  author = {Meta},
  title = {{Introducing Meta Llama 3: The most capable openly available LLM to date}},
  publisher = {Meta},
  journal = {Meta Blog},
  howpublished = "\url{https://ai.meta.com/blog/meta-llama-3/}",
  year = {2024}
}

@misc{codealpaca,
  author = {Sahil Chaudhary},
  title = {Code Alpaca: An Instruction-following LLaMA model for code generation},
  year = {2023},
  publisher = {GitHub},
  journal = {GitHub repository},
  howpublished = {\url{https://github.com/sahil280114/codealpaca}},
}

@misc{codeqwen,
  author = {Qwen Team},
  title = {Code with CodeQwen1.5},
  year = {2024},
  publisher = {Qwen},
  journal = {Qwen Blog},
  howpublished = {\url{https://qwenlm.github.io/blog/codeqwen1.5}},
}

@article{hui2024qwen2,
  title={Qwen2. 5-coder technical report},
  author={Hui, Binyuan and Yang, Jian and Cui, Zeyu and Yang, Jiaxi and Liu, Dayiheng and Zhang, Lei and Liu, Tianyu and Zhang, Jiajun and Yu, Bowen and Dang, Kai and others},
  journal={arXiv preprint arXiv:2409.12186},
  year={2024}
}

@article{reid2024gemini,
  title={Gemini 1.5: Unlocking multimodal understanding across millions of tokens of context},
  author={Reid, Machel and Savinov, Nikolay and Teplyashin, Denis and Lepikhin, Dmitry and Lillicrap, Timothy and Alayrac, Jean-baptiste and Soricut, Radu and Lazaridou, Angeliki and Firat, Orhan and Schrittwieser, Julian and others},
  journal={arXiv preprint arXiv:2403.05530},
  year={2024}
}

@misc{zhang2024artificialintelligencesciencequantum,
      title={Artificial Intelligence for Science in Quantum, Atomistic, and Continuum Systems}, 
      author={Xuan Zhang and Limei Wang and Jacob Helwig and Youzhi Luo and Cong Fu and Yaochen Xie and Meng Liu and Yuchao Lin and Zhao Xu and Keqiang Yan and Keir Adams and Maurice Weiler and Xiner Li and Tianfan Fu and Yucheng Wang and Haiyang Yu and YuQing Xie and Xiang Fu and Alex Strasser and Shenglong Xu and Yi Liu and Yuanqi Du and Alexandra Saxton and Hongyi Ling and Hannah Lawrence and Hannes Stärk and Shurui Gui and Carl Edwards and Nicholas Gao and Adriana Ladera and Tailin Wu and Elyssa F. Hofgard and Aria Mansouri Tehrani and Rui Wang and Ameya Daigavane and Montgomery Bohde and Jerry Kurtin and Qian Huang and Tuong Phung and Minkai Xu and Chaitanya K. Joshi and Simon V. Mathis and Kamyar Azizzadenesheli and Ada Fang and Alán Aspuru-Guzik and Erik Bekkers and Michael Bronstein and Marinka Zitnik and Anima Anandkumar and Stefano Ermon and Pietro Liò and Rose Yu and Stephan Günnemann and Jure Leskovec and Heng Ji and Jimeng Sun and Regina Barzilay and Tommi Jaakkola and Connor W. Coley and Xiaoning Qian and Xiaofeng Qian and Tess Smidt and Shuiwang Ji},
      year={2024},
      eprint={2307.08423},
      archivePrefix={arXiv},
      primaryClass={cs.LG},
      url={https://arxiv.org/abs/2307.08423}, 
}

@article{sun2019review,
  title={A review of the artificial neural network surrogate modeling in aerodynamic design},
  author={Sun, Gang and Wang, Shuyue},
  journal={Proceedings of the Institution of Mechanical Engineers, Part G: Journal of Aerospace Engineering},
  volume={233},
  number={16},
  pages={5863--5872},
  year={2019},
  publisher={SAGE Publications Sage UK: London, England}
}

@article{raissi2020hidden,
  title={Hidden fluid mechanics: Learning velocity and pressure fields from flow visualizations},
  author={Raissi, Maziar and Yazdani, Alireza and Karniadakis, George Em},
  journal={Science},
  volume={367},
  number={6481},
  pages={1026--1030},
  year={2020},
  publisher={American Association for the Advancement of Science}
}

@article{sun2020surrogate,
  title={Surrogate modeling for fluid flows based on physics-constrained deep learning without simulation data},
  author={Sun, Luning and Gao, Han and Pan, Shaowu and Wang, Jian-Xun},
  journal={Computer Methods in Applied Mechanics and Engineering},
  volume={361},
  pages={112732},
  year={2020},
  publisher={Elsevier}
}

@article{bessa2017framework,
  title={A framework for data-driven analysis of materials under uncertainty: Countering the curse of dimensionality},
  author={Bessa, Miguel A and Bostanabad, Ramin and Liu, Zeliang and Hu, Anqi and Apley, Daniel W and Brinson, Catherine and Chen, Wei and Liu, Wing Kam},
  journal={Computer Methods in Applied Mechanics and Engineering},
  volume={320},
  pages={633--667},
  year={2017},
  publisher={Elsevier}
}

@article{thuerey2020deep,
  title={Deep learning methods for Reynolds-averaged Navier--Stokes simulations of airfoil flows},
  author={Thuerey, Nils and Wei{\ss}enow, Konstantin and Prantl, Lukas and Hu, Xiangyu},
  journal={AIAA Journal},
  volume={58},
  number={1},
  pages={25--36},
  year={2020},
  publisher={American Institute of Aeronautics and Astronautics}
}

@article{raissi2019physics,
  title={Physics-informed neural networks: A deep learning framework for solving forward and inverse problems involving nonlinear partial differential equations},
  author={Raissi, Maziar and Perdikaris, Paris and Karniadakis, George E},
  journal={Journal of Computational physics},
  volume={378},
  pages={686--707},
  year={2019},
  publisher={Elsevier}
}

@article{willard2022integrating,
  title={Integrating scientific knowledge with machine learning for engineering and environmental systems},
  author={Willard, Jared and Jia, Xiaowei and Xu, Shaoming and Steinbach, Michael and Kumar, Vipin},
  journal={ACM Computing Surveys},
  volume={55},
  number={4},
  pages={1--37},
  year={2022},
  publisher={ACM New York, NY}
}

@misc{gruver2024largelanguagemodelszeroshot,
      title={Large Language Models Are Zero-Shot Time Series Forecasters}, 
      author={Nate Gruver and Marc Finzi and Shikai Qiu and Andrew Gordon Wilson},
      year={2024},
      eprint={2310.07820},
      archivePrefix={arXiv},
      primaryClass={cs.LG},
      url={https://arxiv.org/abs/2310.07820}, 
}

@misc{hao2024largelanguagemodelssurrogate,
      title={Large Language Models as Surrogate Models in Evolutionary Algorithms: A Preliminary Study}, 
      author={Hao Hao and Xiaoqun Zhang and Aimin Zhou},
      year={2024},
      eprint={2406.10675},
      archivePrefix={arXiv},
      primaryClass={cs.NE},
      url={https://arxiv.org/abs/2406.10675}, 
}

@misc{ma2024llmsimulationbileveloptimizers,
      title={LLM and Simulation as Bilevel Optimizers: A New Paradigm to Advance Physical Scientific Discovery}, 
      author={Pingchuan Ma and Tsun-Hsuan Wang and Minghao Guo and Zhiqing Sun and Joshua B. Tenenbaum and Daniela Rus and Chuang Gan and Wojciech Matusik},
      year={2024},
      eprint={2405.09783},
      archivePrefix={arXiv},
      primaryClass={cs.LG},
      url={https://arxiv.org/abs/2405.09783}, 
}

@misc{wimmer2023leveragingvisionlanguagemodelsgranular,
      title={Leveraging Vision-Language Models for Granular Market Change Prediction}, 
      author={Christopher Wimmer and Navid Rekabsaz},
      year={2023},
      eprint={2301.10166},
      archivePrefix={arXiv},
      primaryClass={q-fin.ST},
      url={https://arxiv.org/abs/2301.10166}, 
}

@misc{che2024gamegenxinteractiveopenworldgame,
      title={GameGen-X: Interactive Open-world Game Video Generation}, 
      author={Haoxuan Che and Xuanhua He and Quande Liu and Cheng Jin and Hao Chen},
      year={2024},
      eprint={2411.00769},
      archivePrefix={arXiv},
      primaryClass={cs.CV},
      url={https://arxiv.org/abs/2411.00769}, 
}

@misc{lu2019efficientsurrogatemodelingmethods,
      title={Efficient surrogate modeling methods for large-scale Earth system models based on machine learning techniques}, 
      author={Dan Lu and Daniel Ricciuto},
      year={2019},
      eprint={1901.05125},
      archivePrefix={arXiv},
      primaryClass={stat.ML},
      url={https://arxiv.org/abs/1901.05125}, 
}

@article{nebbione2023methodological,
  title={A Methodological Framework for AI-Assisted Security Assessments of Active Directory Environments},
  author={Nebbione, Giuseppe and Calzarossa, Maria Carla},
  journal={Ieee Access},
  volume={11},
  pages={15119--15130},
  year={2023},
  publisher={IEEE}
}

@misc{li2025codeiocondensingreasoningpatterns,
      title={CodeI/O: Condensing Reasoning Patterns via Code Input-Output Prediction}, 
      author={Junlong Li and Daya Guo and Dejian Yang and Runxin Xu and Yu Wu and Junxian He},
      year={2025},
      eprint={2502.07316},
      archivePrefix={arXiv},
      primaryClass={cs.CL},
      url={https://arxiv.org/abs/2502.07316}, 
}

@article{10.1145/360248.360252, author = {King, James C.}, title = {Symbolic execution and program testing}, year = {1976}, issue_date = {July 1976}, publisher = {Association for Computing Machinery}, address = {New York, NY, USA}, volume = {19}, number = {7}, issn = {0001-0782}, url = {https://doi.org/10.1145/360248.360252}, doi = {10.1145/360248.360252}, journal = {Commun. ACM}, month = jul, pages = {385–394}, numpages = {10} }

@misc{yan2020neuralexecutionengineslearning,
      title={Neural Execution Engines: Learning to Execute Subroutines}, 
      author={Yujun Yan and Kevin Swersky and Danai Koutra and Parthasarathy Ranganathan and Milad Hashemi},
      year={2020},
      eprint={2006.08084},
      archivePrefix={arXiv},
      primaryClass={cs.LG},
      url={https://arxiv.org/abs/2006.08084}, 
}

@article{Non-intrusive,
author = {Hesthaven, Jan and Ubbiali, Stefano},
year = {2018},
month = {02},
pages = {},
title = {Non-intrusive reduced order modeling of nonlinear problems using neural networks},
volume = {363},
journal = {Journal of Computational Physics},
doi = {10.1016/j.jcp.2018.02.037}
}

@article{doi:10.1137/130932715,
author = {Benner, Peter and Gugercin, Serkan and Willcox, Karen},
title = {A Survey of Projection-Based Model Reduction Methods for Parametric Dynamical Systems},
journal = {SIAM Review},
volume = {57},
number = {4},
pages = {483-531},
year = {2015},
doi = {10.1137/130932715},

URL = { 
    
        https://doi.org/10.1137/130932715
    
    

},
eprint = { 
    
        https://doi.org/10.1137/130932715
    
    

}
}

@inproceedings{ccs,
author = {Shirazi, Farid and Seddighi, Adnan and Iqbal, Amna},
year = {2017},
month = {05},
pages = {534-549},
title = {Cloud Computing Security and Privacy: An Empirical Study},
isbn = {978-3-319-58076-0},
doi = {10.1007/978-3-319-58077-7_43}
}

@misc{wang2024uniquesecurityprivacythreats,
      title={Unique Security and Privacy Threats of Large Language Model: A Comprehensive Survey}, 
      author={Shang Wang and Tianqing Zhu and Bo Liu and Ming Ding and Xu Guo and Dayong Ye and Wanlei Zhou and Philip S. Yu},
      year={2024},
      eprint={2406.07973},
      archivePrefix={arXiv},
      primaryClass={cs.CR},
      url={https://arxiv.org/abs/2406.07973}, 
}

@article{10.1145/2408776.2408795, author = {Cadar, Cristian and Sen, Koushik}, title = {Symbolic execution for software testing: three decades later}, year = {2013}, issue_date = {February 2013}, publisher = {Association for Computing Machinery}, address = {New York, NY, USA}, volume = {56}, number = {2}, issn = {0001-0782}, url = {https://doi.org/10.1145/2408776.2408795}, doi = {10.1145/2408776.2408795}, journal = {Commun. ACM}, month = feb, pages = {82–90}, numpages = {9} }

@article{10.5555/2600239.2600241, author = {Merkel, Dirk}, title = {Docker: lightweight Linux containers for consistent development and deployment}, year = {2014}, issue_date = {March 2014}, publisher = {Belltown Media}, address = {Houston, TX}, volume = {2014}, number = {239}, issn = {1075-3583}, journal = {Linux J.}, month = mar, articleno = {2} }

@misc{tian2024debugbenchevaluatingdebuggingcapability,
      title={DebugBench: Evaluating Debugging Capability of Large Language Models}, 
      author={Runchu Tian and Yining Ye and Yujia Qin and Xin Cong and Yankai Lin and Yinxu Pan and Yesai Wu and Haotian Hui and Weichuan Liu and Zhiyuan Liu and Maosong Sun},
      year={2024},
      eprint={2401.04621},
      archivePrefix={arXiv},
      primaryClass={cs.SE},
      url={https://arxiv.org/abs/2401.04621}, 
}

@article{ying2024lean,
  title={Lean Workbook: A large-scale Lean problem set formalized from natural language math problems},
  author={Ying, Huaiyuan and Wu, Zijian and Geng, Yihan and Wang, Jiayu and Lin, Dahua and Chen, Kai},
  journal={arXiv preprint arXiv:2406.03847},
  year={2024}
}

@misc{lin2025goedelproverfrontiermodelopensource,
          title={Goedel-Prover: A Frontier Model for Open-Source Automated Theorem Proving}, 
          author={Yong Lin and Shange Tang and Bohan Lyu and Jiayun Wu and Hongzhou Lin and Kaiyu Yang and Jia Li and Mengzhou Xia and Danqi Chen and Sanjeev Arora and Chi Jin},
          year={2025},
          eprint={2502.07640},
          archivePrefix={arXiv},
          primaryClass={cs.LG},
          url={https://arxiv.org/abs/2502.07640}, 
        }

@book{paulson1994isabelle,
  title={{Isabelle}: A generic theorem prover},
  author={Paulson, Lawrence C},
  year={1994}
}

@inproceedings{de2015lean,
  title={The Lean theorem prover (system description)},
  author={De Moura, Leonardo and Kong, Soonho and Avigad, Jeremy and Van Doorn, Floris and von Raumer, Jakob},
  booktitle={Automated Deduction-CADE-25: 25th International Conference on Automated Deduction, Berlin, Germany, August 1-7, 2015, Proceedings 25},
  pages={378--388},
  year={2015},
  organization={Springer}
}

@inproceedings{moura2021lean,
  title={The {Lean} 4 theorem prover and programming language},
  author={Moura, Leonardo de and Ullrich, Sebastian},
  year={2021}
}

@phdthesis{barras1997coq,
  title={The {Coq} proof assistant reference manual: Version 6.1},
  author={Barras, Bruno and Boutin, Samuel and Cornes, Cristina and Courant, Judica{\"e}l and Filliatre, Jean-Christophe and Gimenez, Eduardo and Herbelin, Hugo and Huet, Gerard and Munoz, Cesar and Murthy, Chetan and others},
  year={1997},
  school={Inria}
}

@book{10.5555/2423877, author = {Lippman, Stanley B. and Lajoie, Jose and Moo, Barbara E.}, title = {C++ Primer}, year = {2012}, isbn = {0321714113}, publisher = {Addison-Wesley Professional}, edition = {5th} }

@book{10.5555/1841497, author = {Bryant, Randal E. and O'Hallaron, David R.}, title = {Computer Systems: A Programmer's Perspective}, year = {2010}, isbn = {0136108040}, publisher = {Addison-Wesley Publishing Company}, address = {USA}, edition = {2nd} }

@misc{lyu2025adaptinglearninggroundingllms,
      title={Adapting While Learning: Grounding LLMs for Scientific Problems with Intelligent Tool Usage Adaptation}, 
      author={Bohan Lyu and Yadi Cao and Duncan Watson-Parris and Leon Bergen and Taylor Berg-Kirkpatrick and Rose Yu},
      year={2025},
      eprint={2411.00412},
      archivePrefix={arXiv},
      primaryClass={cs.LG},
      url={https://arxiv.org/abs/2411.00412}, 
}

@misc{weber2024learningcompileprogramsneural,
      title={Learning to Compile Programs to Neural Networks}, 
      author={Logan Weber and Jesse Michel and Alex Renda and Michael Carbin},
      year={2024},
      eprint={2407.15078},
      archivePrefix={arXiv},
      primaryClass={cs.LG},
      url={https://arxiv.org/abs/2407.15078}, 
}

@article{10.1145/3617591, author = {Queiroz, Rui and Cruz, Tiago and Mendes, J\'{e}r\^{o}me and Sousa, Pedro and Sim\~{o}es, Paulo}, title = {Container-based Virtualization for Real-time Industrial Systems—A Systematic Review}, year = {2023}, issue_date = {March 2024}, publisher = {Association for Computing Machinery}, address = {New York, NY, USA}, volume = {56}, number = {3}, issn = {0360-0300}, url = {https://doi.org/10.1145/3617591}, doi = {10.1145/3617591}, journal = {ACM Comput. Surv.}, month = oct, articleno = {59}, numpages = {38} }

@misc{gu2025softwaretestingextendedreality,
      title={Software Testing for Extended Reality Applications: A Systematic Mapping Study}, 
      author={Ruizhen Gu and José Miguel Rojas and Donghwan Shin},
      year={2025},
      eprint={2501.08909},
      archivePrefix={arXiv},
      primaryClass={cs.SE},
      url={https://arxiv.org/abs/2501.08909}, 
}

@misc{chai2024mcevalmassivelymultilingualcode,
      title={McEval: Massively Multilingual Code Evaluation}, 
      author={Linzheng Chai and Shukai Liu and Jian Yang and Yuwei Yin and Ke Jin and Jiaheng Liu and Tao Sun and Ge Zhang and Changyu Ren and Hongcheng Guo and Zekun Wang and Boyang Wang and Xianjie Wu and Bing Wang and Tongliang Li and Liqun Yang and Sufeng Duan and Zhoujun Li},
      year={2024},
      eprint={2406.07436},
      archivePrefix={arXiv},
      primaryClass={cs.PL},
      url={https://arxiv.org/abs/2406.07436}, 
}

@article{jaccard1901etude,
  author = {Jaccard, Paul},
  journal = {Bulletin del la Soci\'{e}t\'{e} Vaudoise des Sciences Naturelles},
  pages = {547--579},
  title = {\'{E}tude comparative de la distribution florale dans une portion des Alpes et des Jura},
  volume = 37,
  year = 1901
}

@article{qwen25,
  title={Qwen2.5 technical report},
  author={Qwen Team},
  journal={arXiv preprint arXiv:2412.15115},
  year={2024}
}

@online{anthropic2024claude35,
    author = {Anthropic},
    title = {Introducing {Claude} 3.5 {Sonnet}},
    year = {2024},
    month = {6},
    day = {21},
    url = {https://www.anthropic.com/news/claude-3-5-sonnet},
    organization = {Anthropic PBC},
    urldate = {2025-02-16}
}

@misc{openai2024gpt4o,
  author       = {OpenAI},
  title        = {Hello GPT-4o},
  howpublished = {OpenAI Blog},
  year         = {2024},
  url          = {https://openai.com/index/hello-gpt-4o/},
  note         = {Accessed: 2025-02-16}
}

@misc{openai2024gpt4omini,
  author       = {OpenAI},
  title        = {GPT-4o Mini: Advancing Cost-Efficient Intelligence},
  howpublished = {OpenAI Blog},
  year         = {2024},
  url          = {https://openai.com/index/gpt-4o-mini-advancing-cost-efficient-intelligence/},
  note         = {Accessed: 2025-02-16}
}

@misc{lyu2024largelanguagemodelscode,
      title={Large Language Models as Code Executors: An Exploratory Study}, 
      author={Chenyang Lyu and Lecheng Yan and Rui Xing and Wenxi Li and Younes Samih and Tianbo Ji and Longyue Wang},
      year={2024},
      eprint={2410.06667},
      archivePrefix={arXiv},
      primaryClass={cs.CL},
      url={https://arxiv.org/abs/2410.06667}, 
}

@inproceedings{zheng2024llamafactory,
  title={LlamaFactory: Unified Efficient Fine-Tuning of 100+ Language Models},
  author={Yaowei Zheng and Richong Zhang and Junhao Zhang and Yanhan Ye and Zheyan Luo and Zhangchi Feng and Yongqiang Ma},
  booktitle={Proceedings of the 62nd Annual Meeting of the Association for Computational Linguistics (Volume 3: System Demonstrations)},
  address={Bangkok, Thailand},
  publisher={Association for Computational Linguistics},
  year={2024},
  url={http://arxiv.org/abs/2403.13372}
}

@misc{ouyang2022traininglanguagemodelsfollow,
      title={Training language models to follow instructions with human feedback}, 
      author={Long Ouyang and Jeff Wu and Xu Jiang and Diogo Almeida and Carroll L. Wainwright and Pamela Mishkin and Chong Zhang and Sandhini Agarwal and Katarina Slama and Alex Ray and John Schulman and Jacob Hilton and Fraser Kelton and Luke Miller and Maddie Simens and Amanda Askell and Peter Welinder and Paul Christiano and Jan Leike and Ryan Lowe},
      year={2022},
      eprint={2203.02155},
      archivePrefix={arXiv},
      primaryClass={cs.CL},
      url={https://arxiv.org/abs/2203.02155}, 
}

@article{ratcliff1988pattern,
  title={Pattern Matching: The Gestalt Approach},
  author={Ratcliff, John W. and Metzener, David E.},
  journal={Dr. Dobb's Journal},
  volume={13},
  number={7},
  pages={46--51},
  year={1988},
  publisher={M\&T Publishing Inc.}
}

@article{ahmad2020transformer,
  title={A transformer-based approach for source code summarization},
  author={Ahmad, Wasi Uddin and Chakraborty, Saikat and Ray, Baishakhi and Chang, Kai-Wei},
  journal={arXiv preprint arXiv:2005.00653},
  year={2020}
}

@article{parvez2018building,
  title={Building language models for text with named entities},
  author={Parvez, Md Rizwan and Chakraborty, Saikat and Ray, Baishakhi and Chang, Kai-Wei},
  journal={arXiv preprint arXiv:1805.04836},
  year={2018}
}

\newpage
\section*{Appendix}
\appendix
\label{sec:appendix}

\section{Complete Main Results}
\label{app:main2}

\begin{table*}[]
\centering
\resizebox{\textwidth}{!}{\begin{tabular}{lcccccccccccccccc}
\toprule
\multirow{2}{*}{\textbf{Model}} & \multicolumn{6}{c}{\textbf{ML}} & \multirow{2}{*}{\textbf{CL}} & \multirow{2}{*}{\textbf{RL}} & \multirow{2}{*}{\textbf{SC}} & \multirow{2}{*}{\textbf{TC}} & \multicolumn{3}{c}{\textbf{BG}} & \multirow{2}{*}{\textbf{DR}} & \multirow{2}{*}{\textbf{FL}} & \multirow{2}{*}{\textbf{Avg.}} \\ \cline{2-7} \cline{12-14}
 & \textbf{CPP} & \textbf{Rust} & \textbf{Python} & \textbf{Julia} & \textbf{Java} & \textbf{Others} &  &  &  &  & \textbf{CPP} & \textbf{Java} & \textbf{Python} &  &  &  \\ \midrule
\multicolumn{17}{c}{\textit{Zero-shot}}                                                                                                                                                           \\ \midrule
\texttt{Claude-3.5-Sonnet} & \cellcolor{cyan!45}$72.73$ & \cellcolor{cyan!34}$55.00$ & \cellcolor{cyan!55}$88.00$ & \cellcolor{cyan!41}$66.67$ & \cellcolor{cyan!48}$76.92$ & \cellcolor{cyan!46}$75.00$ & \cellcolor{cyan!50}$81.58$ & \cellcolor{cyan!35}$57.31$ & \cellcolor{cyan!38}$61.55$ & \cellcolor{cyan!22}$35.27$ & \cellcolor{cyan!5}$9.09$ & \cellcolor{cyan!7}$12.55$ & \cellcolor{cyan!32}$51.40$ & \cellcolor{cyan!8}$12.92$ & \cellcolor{cyan!11}$17.92$ & \cellcolor{cyan!32}$51.59$ \\
\texttt{DeepSeek-V3} & \cellcolor{cyan!34}$54.55$ & \cellcolor{cyan!37}$60.00$ & \cellcolor{cyan!47}$76.00$ & \cellcolor{cyan!38}$61.11$ & \cellcolor{cyan!28}$46.15$ & \cellcolor{cyan!45}$72.50$ & \cellcolor{cyan!35}$56.58$ & \cellcolor{cyan!27}$44.19$ & \cellcolor{cyan!37}$59.65$ & \cellcolor{cyan!22}$35.31$ & \cellcolor{cyan!2}$4.05$ & \cellcolor{cyan!1}$3.03$ & \cellcolor{cyan!13}$21.50$ & \cellcolor{cyan!6}$10.92$ & \cellcolor{cyan!20}$32.46$ & \cellcolor{cyan!26}$42.53$ \\
\texttt{GPT-4o} & \cellcolor{cyan!25}$40.91$ & \cellcolor{cyan!28}$45.00$ & \cellcolor{cyan!37}$60.00$ & \cellcolor{cyan!34}$55.56$ & \cellcolor{cyan!36}$57.69$ & \cellcolor{cyan!34}$55.00$ & \cellcolor{cyan!41}$66.45$ & \cellcolor{cyan!30}$49.13$ & \cellcolor{cyan!33}$53.16$ & \cellcolor{cyan!21}$34.44$ & \cellcolor{cyan!2}$4.28$ & \cellcolor{cyan!4}$7.48$ & \cellcolor{cyan!21}$34.59$ & \cellcolor{cyan!9}$14.75$ & \cellcolor{cyan!13}$21.99$ & \cellcolor{cyan!25}$40.03$ \\
\texttt{GPT-4o-Mini} & \cellcolor{cyan!36}$59.09$ & \cellcolor{cyan!28}$45.00$ & \cellcolor{cyan!40}$64.00$ & \cellcolor{cyan!20}$33.33$ & \cellcolor{cyan!43}$69.23$ & \cellcolor{cyan!37}$60.00$ & \cellcolor{cyan!52}$84.21$ & \cellcolor{cyan!22}$35.33$ & \cellcolor{cyan!25}$40.00$ & \cellcolor{cyan!19}$31.85$ & \cellcolor{cyan!0}$1.39$ & \cellcolor{cyan!2}$4.28$ & \cellcolor{cyan!8}$13.86$ & \cellcolor{cyan!7}$11.65$ & \cellcolor{cyan!24}$39.07$ & \cellcolor{cyan!24}$39.49$ \\
\texttt{Qwen-Max} & \cellcolor{cyan!31}$50.00$ & \cellcolor{cyan!28}$45.00$ & \cellcolor{cyan!27}$44.00$ & \cellcolor{cyan!17}$27.78$ & \cellcolor{cyan!16}$26.92$ & \cellcolor{cyan!31}$50.00$ & \cellcolor{cyan!24}$38.82$ & \cellcolor{cyan!23}$37.15$ & \cellcolor{cyan!35}$56.98$ & \cellcolor{cyan!22}$35.44$ & \cellcolor{cyan!1}$2.82$ & \cellcolor{cyan!2}$3.20$ & \cellcolor{cyan!19}$30.52$ & \cellcolor{cyan!8}$14.03$ & \cellcolor{cyan!18}$29.89$ & \cellcolor{cyan!20}$32.84$ \\
\texttt{Qwen-2.5-0.5B-Instruct} & \cellcolor{cyan!8}$13.64$ & \cellcolor{cyan!6}$10.00$ & \cellcolor{cyan!2}$4.00$ & \cellcolor{cyan!0}$0.00$ & \cellcolor{cyan!7}$11.54$ & \cellcolor{cyan!4}$7.50$ & \cellcolor{cyan!11}$19.08$ & \cellcolor{cyan!3}$5.02$ & \cellcolor{cyan!5}$9.43$ & \cellcolor{cyan!5}$8.61$ & \cellcolor{cyan!0}$1.20$ & \cellcolor{cyan!2}$3.77$ & \cellcolor{cyan!6}$11.04$ & \cellcolor{cyan!2}$4.62$ & \cellcolor{cyan!26}$42.38$ & \cellcolor{cyan!6}$10.85$ \\
\texttt{Qwen-2.5-1.5B-Instruct} & \cellcolor{cyan!22}$36.36$ & \cellcolor{cyan!9}$15.00$ & \cellcolor{cyan!15}$24.00$ & \cellcolor{cyan!10}$16.67$ & \cellcolor{cyan!9}$15.38$ & \cellcolor{cyan!21}$35.00$ & \cellcolor{cyan!25}$40.79$ & \cellcolor{cyan!3}$5.89$ & \cellcolor{cyan!7}$12.08$ & \cellcolor{cyan!8}$14.15$ & \cellcolor{cyan!0}$1.15$ & \cellcolor{cyan!2}$3.52$ & \cellcolor{cyan!6}$11.10$ & \cellcolor{cyan!5}$9.24$ & \cellcolor{cyan!26}$41.72$ & \cellcolor{cyan!11}$18.80$ \\
\texttt{Qwen-2.5-3B-Instruct} & \cellcolor{cyan!22}$36.36$ & \cellcolor{cyan!6}$10.00$ & \cellcolor{cyan!17}$28.00$ & \cellcolor{cyan!13}$22.22$ & \cellcolor{cyan!21}$34.62$ & \cellcolor{cyan!14}$22.50$ & \cellcolor{cyan!22}$35.53$ & \cellcolor{cyan!8}$13.08$ & \cellcolor{cyan!12}$20.04$ & \cellcolor{cyan!9}$14.61$ & \cellcolor{cyan!0}$1.58$ & \cellcolor{cyan!2}$3.92$ & \cellcolor{cyan!8}$13.27$ & \cellcolor{cyan!3}$5.75$ & \cellcolor{cyan!9}$15.89$ & \cellcolor{cyan!11}$18.49$ \\
\texttt{Qwen-2.5-7B-Instruct} & \cellcolor{cyan!8}$13.64$ & \cellcolor{cyan!3}$5.00$ & \cellcolor{cyan!7}$12.00$ & \cellcolor{cyan!3}$5.56$ & \cellcolor{cyan!7}$11.54$ & \cellcolor{cyan!10}$17.50$ & \cellcolor{cyan!17}$27.63$ & \cellcolor{cyan!17}$27.98$ & \cellcolor{cyan!14}$22.90$ & \cellcolor{cyan!18}$29.43$ & \cellcolor{cyan!1}$2.21$ & \cellcolor{cyan!2}$3.66$ & \cellcolor{cyan!5}$9.46$ & \cellcolor{cyan!4}$7.24$ & \cellcolor{cyan!22}$36.42$ & \cellcolor{cyan!9}$15.48$ \\
\texttt{Qwen-2.5-14B-Instruct} & \cellcolor{cyan!5}$9.09$ & \cellcolor{cyan!3}$5.00$ & \cellcolor{cyan!5}$8.00$ & \cellcolor{cyan!3}$5.56$ & \cellcolor{cyan!14}$23.08$ & \cellcolor{cyan!3}$5.00$ & \cellcolor{cyan!6}$11.18$ & \cellcolor{cyan!24}$39.08$ & \cellcolor{cyan!21}$33.79$ & \cellcolor{cyan!17}$28.18$ & \cellcolor{cyan!1}$2.11$ & \cellcolor{cyan!1}$2.24$ & \cellcolor{cyan!10}$16.30$ & \cellcolor{cyan!2}$4.24$ & \cellcolor{cyan!4}$7.28$ & \cellcolor{cyan!8}$13.34$ \\
\texttt{Qwen-2.5-32B-Instruct} & \cellcolor{cyan!28}$45.45$ & \cellcolor{cyan!12}$20.00$ & \cellcolor{cyan!30}$48.00$ & \cellcolor{cyan!20}$33.33$ & \cellcolor{cyan!26}$42.31$ & \cellcolor{cyan!12}$20.00$ & \cellcolor{cyan!31}$50.66$ & \cellcolor{cyan!14}$22.61$ & \cellcolor{cyan!20}$32.50$ & \cellcolor{cyan!19}$30.99$ & \cellcolor{cyan!0}$1.21$ & \cellcolor{cyan!1}$2.09$ & \cellcolor{cyan!7}$12.16$ & \cellcolor{cyan!4}$7.26$ & \cellcolor{cyan!4}$7.65$ & \cellcolor{cyan!15}$25.08$ \\
\texttt{Qwen-2.5-72B-Instruct} & \cellcolor{cyan!28}$45.45$ & \cellcolor{cyan!25}$40.00$ & \cellcolor{cyan!32}$52.00$ & \cellcolor{cyan!34}$55.56$ & \cellcolor{cyan!43}$69.23$ & \cellcolor{cyan!32}$52.50$ & \cellcolor{cyan!45}$72.37$ & \cellcolor{cyan!21}$33.66$ & \cellcolor{cyan!33}$53.70$ & \cellcolor{cyan!20}$32.46$ & \cellcolor{cyan!1}$1.99$ & \cellcolor{cyan!1}$2.65$ & \cellcolor{cyan!6}$10.69$ & \cellcolor{cyan!8}$13.35$ & \cellcolor{cyan!21}$34.44$ & \cellcolor{cyan!23}$38.00$ \\
\texttt{Qwen-2.5-Coder-0.5B-Instruct} & \cellcolor{cyan!14}$22.73$ & \cellcolor{cyan!6}$10.00$ & \cellcolor{cyan!12}$20.00$ & \cellcolor{cyan!3}$5.56$ & \cellcolor{cyan!7}$11.54$ & \cellcolor{cyan!14}$22.50$ & \cellcolor{cyan!16}$26.97$ & \cellcolor{cyan!5}$8.41$ & \cellcolor{cyan!4}$6.55$ & \cellcolor{cyan!3}$6.39$ & \cellcolor{cyan!0}$0.73$ & \cellcolor{cyan!1}$2.92$ & \cellcolor{cyan!6}$10.05$ & \cellcolor{cyan!3}$5.76$ & \cellcolor{cyan!25}$41.06$ & \cellcolor{cyan!8}$13.41$ \\
\texttt{Qwen-2.5-Coder-1.5B-Instruct} & \cellcolor{cyan!28}$45.45$ & \cellcolor{cyan!18}$30.00$ & \cellcolor{cyan!20}$32.00$ & \cellcolor{cyan!17}$27.78$ & \cellcolor{cyan!16}$26.92$ & \cellcolor{cyan!21}$35.00$ & \cellcolor{cyan!34}$54.61$ & \cellcolor{cyan!10}$17.58$ & \cellcolor{cyan!12}$20.58$ & \cellcolor{cyan!8}$13.19$ & \cellcolor{cyan!0}$0.36$ & \cellcolor{cyan!1}$2.83$ & \cellcolor{cyan!3}$5.43$ & \cellcolor{cyan!5}$8.08$ & \cellcolor{cyan!25}$41.06$ & \cellcolor{cyan!15}$24.06$ \\
\texttt{Qwen-2.5-Coder-3B-Instruct} & \cellcolor{cyan!28}$45.45$ & \cellcolor{cyan!12}$20.00$ & \cellcolor{cyan!25}$40.00$ & \cellcolor{cyan!13}$22.22$ & \cellcolor{cyan!28}$46.15$ & \cellcolor{cyan!25}$40.00$ & \cellcolor{cyan!42}$68.42$ & \cellcolor{cyan!13}$21.58$ & \cellcolor{cyan!23}$37.43$ & \cellcolor{cyan!14}$22.43$ & \cellcolor{cyan!0}$1.06$ & \cellcolor{cyan!2}$3.61$ & \cellcolor{cyan!8}$13.41$ & \cellcolor{cyan!5}$8.96$ & \cellcolor{cyan!25}$41.06$ & \cellcolor{cyan!17}$28.79$ \\
\texttt{Qwen-2.5-Coder-7B-Instruct} & \cellcolor{cyan!28}$45.45$ & \cellcolor{cyan!18}$30.00$ & \cellcolor{cyan!32}$52.00$ & \cellcolor{cyan!27}$44.44$ & \cellcolor{cyan!31}$50.00$ & \cellcolor{cyan!26}$42.50$ & \cellcolor{cyan!46}$75.00$ & \cellcolor{cyan!13}$20.86$ & \cellcolor{cyan!28}$45.50$ & \cellcolor{cyan!17}$28.00$ & \cellcolor{cyan!0}$0.86$ & \cellcolor{cyan!2}$3.47$ & \cellcolor{cyan!8}$13.53$ & \cellcolor{cyan!8}$14.23$ & \cellcolor{cyan!25}$40.40$ & \cellcolor{cyan!21}$33.75$ \\
\texttt{Qwen-2.5-Coder-14B-Instruct} & \cellcolor{cyan!36}$59.09$ & \cellcolor{cyan!21}$35.00$ & \cellcolor{cyan!30}$48.00$ & \cellcolor{cyan!34}$55.56$ & \cellcolor{cyan!43}$69.23$ & \cellcolor{cyan!39}$62.50$ & \cellcolor{cyan!51}$82.89$ & \cellcolor{cyan!22}$35.26$ & \cellcolor{cyan!32}$51.93$ & \cellcolor{cyan!20}$32.20$ & \cellcolor{cyan!0}$1.39$ & \cellcolor{cyan!2}$3.66$ & \cellcolor{cyan!10}$16.12$ & \cellcolor{cyan!7}$11.35$ & \cellcolor{cyan!26}$41.72$ & \cellcolor{cyan!25}$40.39$ \\
\texttt{Qwen-2.5-Coder-32B-Instruct} & \cellcolor{cyan!36}$59.09$ & \cellcolor{cyan!34}$55.00$ & \cellcolor{cyan!37}$60.00$ & \cellcolor{cyan!27}$44.44$ & \cellcolor{cyan!43}$69.23$ & \cellcolor{cyan!37}$60.00$ & \cellcolor{cyan!50}$80.26$ & \cellcolor{cyan!30}$48.43$ & \cellcolor{cyan!35}$57.18$ & \cellcolor{cyan!11}$18.84$ & \cellcolor{cyan!0}$1.49$ & \cellcolor{cyan!1}$1.95$ & \cellcolor{cyan!11}$17.96$ & \cellcolor{cyan!8}$13.42$ & \cellcolor{cyan!18}$29.19$ & \cellcolor{cyan!25}$41.10$ \\
\texttt{LLaMA-3.1-8B-Instruct} & \cellcolor{cyan!0}$0.00$ & \cellcolor{cyan!0}$0.00$ & \cellcolor{cyan!2}$4.00$ & \cellcolor{cyan!3}$5.56$ & \cellcolor{cyan!9}$15.38$ & \cellcolor{cyan!3}$5.00$ & \cellcolor{cyan!8}$13.16$ & \cellcolor{cyan!12}$20.41$ & \cellcolor{cyan!2}$4.41$ & \cellcolor{cyan!9}$15.46$ & \cellcolor{cyan!2}$4.41$ & \cellcolor{cyan!2}$4.12$ & \cellcolor{cyan!3}$5.98$ & \cellcolor{cyan!2}$3.97$ & \cellcolor{cyan!0}$0.00$ & \cellcolor{cyan!5}$8.49$ \\
\texttt{LLaMA-3.1-70B-Instruct} & \cellcolor{cyan!34}$54.55$ & \cellcolor{cyan!25}$40.00$ & \cellcolor{cyan!32}$52.00$ & \cellcolor{cyan!20}$33.33$ & \cellcolor{cyan!40}$65.38$ & \cellcolor{cyan!29}$47.50$ & \cellcolor{cyan!48}$78.29$ & \cellcolor{cyan!19}$31.60$ & \cellcolor{cyan!30}$48.65$ & \cellcolor{cyan!18}$30.19$ & \cellcolor{cyan!0}$1.59$ & \cellcolor{cyan!2}$4.18$ & \cellcolor{cyan!8}$14.27$ & \cellcolor{cyan!9}$15.73$ & \cellcolor{cyan!24}$39.16$ & \cellcolor{cyan!23}$37.10$ \\
\texttt{LLaMA-3.3-70B-Instruct} & \cellcolor{cyan!42}$68.18$ & \cellcolor{cyan!31}$50.00$ & \cellcolor{cyan!37}$60.00$ & \cellcolor{cyan!27}$44.44$ & \cellcolor{cyan!36}$57.69$ & \cellcolor{cyan!39}$62.50$ & \cellcolor{cyan!41}$66.45$ & \cellcolor{cyan!27}$43.53$ & \cellcolor{cyan!24}$38.45$ & \cellcolor{cyan!18}$30.35$ & \cellcolor{cyan!1}$2.06$ & \cellcolor{cyan!2}$3.23$ & \cellcolor{cyan!6}$11.01$ & \cellcolor{cyan!7}$11.22$ & \cellcolor{cyan!24}$39.13$ & \cellcolor{cyan!24}$39.22$ \\
\midrule
\multicolumn{17}{c}{\textit{Zero-shot Chain-of-Thought}}                                                                                                                                      \\ \midrule
\texttt{Claude-3.5-Sonnet} & \cellcolor{cyan!56}$90.91$ & \cellcolor{cyan!40}$65.00$ & \cellcolor{cyan!60}$96.00$ & \cellcolor{cyan!48}$77.78$ & \cellcolor{cyan!43}$69.23$ & \cellcolor{cyan!57}$92.50$ & \cellcolor{cyan!51}$82.24$ & \cellcolor{cyan!38}$62.31$ & \cellcolor{cyan!39}$63.38$ & \cellcolor{cyan!25}$40.70$ & \cellcolor{cyan!10}$16.91$ & \cellcolor{cyan!12}$20.69$ & \cellcolor{cyan!38}$62.23$ & \cellcolor{cyan!11}$18.19$ & \cellcolor{cyan!21}$33.98$ & \cellcolor{cyan!37}$59.47$ \\
\texttt{DeepSeek-V3} & \cellcolor{cyan!51}$81.82$ & \cellcolor{cyan!53}$85.00$ & \cellcolor{cyan!55}$88.00$ & \cellcolor{cyan!45}$72.22$ & \cellcolor{cyan!43}$69.23$ & \cellcolor{cyan!53}$85.00$ & \cellcolor{cyan!47}$76.32$ & \cellcolor{cyan!39}$62.70$ & \cellcolor{cyan!35}$57.57$ & \cellcolor{cyan!22}$36.71$ & \cellcolor{cyan!2}$4.45$ & \cellcolor{cyan!4}$7.85$ & \cellcolor{cyan!28}$46.26$ & \cellcolor{cyan!10}$16.21$ & \cellcolor{cyan!21}$35.19$ & \cellcolor{cyan!34}$54.97$ \\
\texttt{GPT-4o} & \cellcolor{cyan!42}$68.18$ & \cellcolor{cyan!40}$65.00$ & \cellcolor{cyan!57}$92.00$ & \cellcolor{cyan!45}$72.22$ & \cellcolor{cyan!48}$76.92$ & \cellcolor{cyan!48}$77.50$ & \cellcolor{cyan!49}$79.61$ & \cellcolor{cyan!33}$53.74$ & \cellcolor{cyan!30}$48.56$ & \cellcolor{cyan!17}$28.36$ & \cellcolor{cyan!5}$8.19$ & \cellcolor{cyan!6}$9.97$ & \cellcolor{cyan!27}$44.29$ & \cellcolor{cyan!8}$14.21$ & \cellcolor{cyan!17}$27.91$ & \cellcolor{cyan!31}$51.11$ \\
\texttt{GPT-4o-Mini} & \cellcolor{cyan!48}$77.27$ & \cellcolor{cyan!37}$60.00$ & \cellcolor{cyan!55}$88.00$ & \cellcolor{cyan!31}$50.00$ & \cellcolor{cyan!43}$69.23$ & \cellcolor{cyan!50}$80.00$ & \cellcolor{cyan!47}$75.66$ & \cellcolor{cyan!21}$34.46$ & \cellcolor{cyan!25}$40.60$ & \cellcolor{cyan!18}$29.59$ & \cellcolor{cyan!1}$1.77$ & \cellcolor{cyan!3}$5.40$ & \cellcolor{cyan!13}$21.04$ & \cellcolor{cyan!8}$13.16$ & \cellcolor{cyan!20}$33.11$ & \cellcolor{cyan!28}$45.29$ \\
\texttt{Qwen-Max} & \cellcolor{cyan!53}$86.36$ & \cellcolor{cyan!46}$75.00$ & \cellcolor{cyan!50}$80.00$ & \cellcolor{cyan!45}$72.22$ & \cellcolor{cyan!48}$76.92$ & \cellcolor{cyan!50}$80.00$ & \cellcolor{cyan!44}$71.05$ & \cellcolor{cyan!31}$50.49$ & \cellcolor{cyan!38}$61.78$ & \cellcolor{cyan!22}$36.71$ & \cellcolor{cyan!1}$2.65$ & \cellcolor{cyan!4}$7.73$ & \cellcolor{cyan!29}$46.85$ & \cellcolor{cyan!10}$16.16$ & \cellcolor{cyan!12}$20.74$ & \cellcolor{cyan!32}$52.31$ \\
\texttt{Qwen-2.5-0.5B-Instruct} & \cellcolor{cyan!17}$27.27$ & \cellcolor{cyan!6}$10.00$ & \cellcolor{cyan!10}$16.00$ & \cellcolor{cyan!0}$0.00$ & \cellcolor{cyan!2}$3.85$ & \cellcolor{cyan!10}$17.50$ & \cellcolor{cyan!13}$22.37$ & \cellcolor{cyan!3}$6.29$ & \cellcolor{cyan!0}$1.11$ & \cellcolor{cyan!3}$5.28$ & \cellcolor{cyan!0}$1.22$ & \cellcolor{cyan!2}$3.41$ & \cellcolor{cyan!5}$9.15$ & \cellcolor{cyan!3}$5.21$ & \cellcolor{cyan!23}$37.09$ & \cellcolor{cyan!7}$11.84$ \\
\texttt{Qwen-2.5-1.5B-Instruct} & \cellcolor{cyan!19}$31.82$ & \cellcolor{cyan!9}$15.00$ & \cellcolor{cyan!12}$20.00$ & \cellcolor{cyan!6}$11.11$ & \cellcolor{cyan!12}$19.23$ & \cellcolor{cyan!17}$27.50$ & \cellcolor{cyan!16}$26.32$ & \cellcolor{cyan!7}$12.47$ & \cellcolor{cyan!5}$8.14$ & \cellcolor{cyan!7}$12.77$ & \cellcolor{cyan!0}$1.22$ & \cellcolor{cyan!2}$3.23$ & \cellcolor{cyan!6}$9.81$ & \cellcolor{cyan!3}$4.93$ & \cellcolor{cyan!24}$39.74$ & \cellcolor{cyan!10}$16.22$ \\
\texttt{Qwen-2.5-3B-Instruct} & \cellcolor{cyan!25}$40.91$ & \cellcolor{cyan!25}$40.00$ & \cellcolor{cyan!20}$32.00$ & \cellcolor{cyan!13}$22.22$ & \cellcolor{cyan!19}$30.77$ & \cellcolor{cyan!21}$35.00$ & \cellcolor{cyan!15}$25.00$ & \cellcolor{cyan!9}$15.01$ & \cellcolor{cyan!12}$20.78$ & \cellcolor{cyan!8}$12.84$ & \cellcolor{cyan!1}$2.14$ & \cellcolor{cyan!1}$2.86$ & \cellcolor{cyan!3}$5.29$ & \cellcolor{cyan!4}$7.12$ & \cellcolor{cyan!8}$13.93$ & \cellcolor{cyan!12}$20.39$ \\
\texttt{Qwen-2.5-7B-Instruct} & \cellcolor{cyan!25}$40.91$ & \cellcolor{cyan!9}$15.00$ & \cellcolor{cyan!20}$32.00$ & \cellcolor{cyan!20}$33.33$ & \cellcolor{cyan!16}$26.92$ & \cellcolor{cyan!29}$47.50$ & \cellcolor{cyan!32}$52.63$ & \cellcolor{cyan!17}$27.51$ & \cellcolor{cyan!16}$25.68$ & \cellcolor{cyan!18}$28.95$ & \cellcolor{cyan!0}$1.12$ & \cellcolor{cyan!2}$3.76$ & \cellcolor{cyan!9}$14.94$ & \cellcolor{cyan!5}$9.41$ & \cellcolor{cyan!22}$36.46$ & \cellcolor{cyan!16}$26.41$ \\
\texttt{Qwen-2.5-14B-Instruct} & \cellcolor{cyan!42}$68.18$ & \cellcolor{cyan!34}$55.00$ & \cellcolor{cyan!47}$76.00$ & \cellcolor{cyan!38}$61.11$ & \cellcolor{cyan!40}$65.38$ & \cellcolor{cyan!45}$72.50$ & \cellcolor{cyan!38}$61.18$ & \cellcolor{cyan!27}$43.89$ & \cellcolor{cyan!22}$36.49$ & \cellcolor{cyan!20}$32.07$ & \cellcolor{cyan!0}$1.57$ & \cellcolor{cyan!1}$3.15$ & \cellcolor{cyan!11}$19.13$ & \cellcolor{cyan!7}$11.97$ & \cellcolor{cyan!5}$8.67$ & \cellcolor{cyan!25}$41.09$ \\
\texttt{Qwen-2.5-32B-Instruct} & \cellcolor{cyan!31}$50.00$ & \cellcolor{cyan!25}$40.00$ & \cellcolor{cyan!25}$40.00$ & \cellcolor{cyan!34}$55.56$ & \cellcolor{cyan!24}$38.46$ & \cellcolor{cyan!35}$57.50$ & \cellcolor{cyan!33}$53.29$ & \cellcolor{cyan!20}$32.71$ & \cellcolor{cyan!27}$43.29$ & \cellcolor{cyan!19}$30.59$ & \cellcolor{cyan!1}$3.10$ & \cellcolor{cyan!4}$6.96$ & \cellcolor{cyan!14}$23.86$ & \cellcolor{cyan!7}$11.49$ & \cellcolor{cyan!4}$7.39$ & \cellcolor{cyan!20}$32.95$ \\
\texttt{Qwen-2.5-72B-Instruct} & \cellcolor{cyan!42}$68.18$ & \cellcolor{cyan!50}$80.00$ & \cellcolor{cyan!60}$96.00$ & \cellcolor{cyan!34}$55.56$ & \cellcolor{cyan!48}$76.92$ & \cellcolor{cyan!48}$77.50$ & \cellcolor{cyan!40}$65.13$ & \cellcolor{cyan!28}$45.30$ & \cellcolor{cyan!33}$53.39$ & \cellcolor{cyan!20}$32.95$ & \cellcolor{cyan!1}$1.72$ & \cellcolor{cyan!1}$3.19$ & \cellcolor{cyan!10}$16.32$ & \cellcolor{cyan!10}$16.61$ & \cellcolor{cyan!18}$29.80$ & \cellcolor{cyan!29}$47.90$ \\
\texttt{Qwen-2.5-Coder-0.5B-Instruct} & \cellcolor{cyan!8}$13.64$ & \cellcolor{cyan!0}$0.00$ & \cellcolor{cyan!2}$4.00$ & \cellcolor{cyan!3}$5.56$ & \cellcolor{cyan!2}$3.85$ & \cellcolor{cyan!4}$7.50$ & \cellcolor{cyan!1}$1.97$ & \cellcolor{cyan!1}$3.07$ & \cellcolor{cyan!1}$2.14$ & \cellcolor{cyan!2}$3.98$ & \cellcolor{cyan!1}$2.14$ & \cellcolor{cyan!1}$2.44$ & \cellcolor{cyan!1}$2.83$ & \cellcolor{cyan!1}$2.23$ & \cellcolor{cyan!21}$33.77$ & \cellcolor{cyan!3}$6.36$ \\
\texttt{Qwen-2.5-Coder-1.5B-Instruct} & \cellcolor{cyan!19}$31.82$ & \cellcolor{cyan!6}$10.00$ & \cellcolor{cyan!17}$28.00$ & \cellcolor{cyan!3}$5.56$ & \cellcolor{cyan!9}$15.38$ & \cellcolor{cyan!14}$22.50$ & \cellcolor{cyan!11}$19.08$ & \cellcolor{cyan!16}$26.39$ & \cellcolor{cyan!12}$20.08$ & \cellcolor{cyan!9}$15.42$ & \cellcolor{cyan!0}$1.16$ & \cellcolor{cyan!2}$3.60$ & \cellcolor{cyan!7}$11.46$ & \cellcolor{cyan!4}$7.24$ & \cellcolor{cyan!24}$39.74$ & \cellcolor{cyan!10}$17.16$ \\
\texttt{Qwen-2.5-Coder-3B-Instruct} & \cellcolor{cyan!31}$50.00$ & \cellcolor{cyan!9}$15.00$ & \cellcolor{cyan!20}$32.00$ & \cellcolor{cyan!13}$22.22$ & \cellcolor{cyan!26}$42.31$ & \cellcolor{cyan!25}$40.00$ & \cellcolor{cyan!32}$52.63$ & \cellcolor{cyan!8}$13.13$ & \cellcolor{cyan!21}$33.96$ & \cellcolor{cyan!12}$20.14$ & \cellcolor{cyan!0}$1.24$ & \cellcolor{cyan!2}$3.72$ & \cellcolor{cyan!8}$13.36$ & \cellcolor{cyan!4}$7.63$ & \cellcolor{cyan!25}$40.40$ & \cellcolor{cyan!16}$25.85$ \\
\texttt{Qwen-2.5-Coder-7B-Instruct} & \cellcolor{cyan!42}$68.18$ & \cellcolor{cyan!25}$40.00$ & \cellcolor{cyan!25}$40.00$ & \cellcolor{cyan!24}$38.89$ & \cellcolor{cyan!33}$53.85$ & \cellcolor{cyan!35}$57.50$ & \cellcolor{cyan!28}$46.05$ & \cellcolor{cyan!12}$19.70$ & \cellcolor{cyan!25}$40.91$ & \cellcolor{cyan!18}$30.19$ & \cellcolor{cyan!1}$2.29$ & \cellcolor{cyan!2}$4.71$ & \cellcolor{cyan!7}$12.77$ & \cellcolor{cyan!9}$15.04$ & \cellcolor{cyan!23}$37.12$ & \cellcolor{cyan!21}$33.81$ \\
\texttt{Qwen-2.5-Coder-14B-Instruct} & \cellcolor{cyan!36}$59.09$ & \cellcolor{cyan!28}$45.00$ & \cellcolor{cyan!37}$60.00$ & \cellcolor{cyan!34}$55.56$ & \cellcolor{cyan!43}$69.23$ & \cellcolor{cyan!39}$62.50$ & \cellcolor{cyan!44}$71.71$ & \cellcolor{cyan!16}$26.09$ & \cellcolor{cyan!32}$52.12$ & \cellcolor{cyan!20}$33.09$ & \cellcolor{cyan!1}$3.19$ & \cellcolor{cyan!3}$5.21$ & \cellcolor{cyan!11}$18.58$ & \cellcolor{cyan!8}$13.41$ & \cellcolor{cyan!21}$34.44$ & \cellcolor{cyan!25}$40.61$ \\
\texttt{Qwen-2.5-Coder-32B-Instruct} & \cellcolor{cyan!48}$77.27$ & \cellcolor{cyan!40}$65.00$ & \cellcolor{cyan!50}$80.00$ & \cellcolor{cyan!34}$55.56$ & \cellcolor{cyan!45}$73.08$ & \cellcolor{cyan!42}$67.50$ & \cellcolor{cyan!44}$71.71$ & \cellcolor{cyan!34}$54.58$ & \cellcolor{cyan!34}$55.69$ & \cellcolor{cyan!21}$34.36$ & \cellcolor{cyan!1}$2.05$ & \cellcolor{cyan!2}$4.74$ & \cellcolor{cyan!14}$22.43$ & \cellcolor{cyan!11}$17.62$ & \cellcolor{cyan!17}$28.55$ & \cellcolor{cyan!29}$47.34$ \\
\texttt{LLaMA-3.1-8B-Instruct} & \cellcolor{cyan!25}$40.91$ & \cellcolor{cyan!9}$15.00$ & \cellcolor{cyan!15}$24.00$ & \cellcolor{cyan!13}$22.22$ & \cellcolor{cyan!16}$26.92$ & \cellcolor{cyan!18}$30.00$ & \cellcolor{cyan!25}$41.45$ & \cellcolor{cyan!11}$17.87$ & \cellcolor{cyan!20}$32.65$ & \cellcolor{cyan!11}$18.22$ & \cellcolor{cyan!0}$1.52$ & \cellcolor{cyan!2}$4.23$ & \cellcolor{cyan!8}$13.38$ & \cellcolor{cyan!6}$10.12$ & \cellcolor{cyan!0}$0.66$ & \cellcolor{cyan!12}$19.94$ \\
\texttt{LLaMA-3.1-70B-Instruct} & \cellcolor{cyan!36}$59.09$ & \cellcolor{cyan!31}$50.00$ & \cellcolor{cyan!45}$72.00$ & \cellcolor{cyan!38}$61.11$ & \cellcolor{cyan!36}$57.69$ & \cellcolor{cyan!32}$52.50$ & \cellcolor{cyan!36}$58.55$ & \cellcolor{cyan!21}$34.44$ & \cellcolor{cyan!27}$43.93$ & \cellcolor{cyan!18}$29.76$ & \cellcolor{cyan!1}$1.71$ & \cellcolor{cyan!2}$3.49$ & \cellcolor{cyan!9}$15.02$ & \cellcolor{cyan!10}$16.86$ & \cellcolor{cyan!16}$25.85$ & \cellcolor{cyan!24}$38.80$ \\
\texttt{LLaMA-3.3-70B-Instruct} & \cellcolor{cyan!39}$63.64$ & \cellcolor{cyan!28}$45.00$ & \cellcolor{cyan!35}$56.00$ & \cellcolor{cyan!34}$55.56$ & \cellcolor{cyan!36}$57.69$ & \cellcolor{cyan!42}$67.50$ & \cellcolor{cyan!35}$57.24$ & \cellcolor{cyan!22}$35.88$ & \cellcolor{cyan!24}$39.50$ & \cellcolor{cyan!19}$30.61$ & \cellcolor{cyan!2}$3.34$ & \cellcolor{cyan!2}$3.95$ & \cellcolor{cyan!8}$13.53$ & \cellcolor{cyan!10}$16.38$ & \cellcolor{cyan!21}$35.11$ & \cellcolor{cyan!24}$38.73$ \\
\midrule
\multicolumn{17}{c}{\textit{Few-shot Chain-of-Thought}}                                                                                                                                        \\ \midrule
\texttt{Claude-3.5-Sonnet} & \cellcolor{cyan!53}$86.36$ & \cellcolor{cyan!43}$70.00$ & \cellcolor{cyan!60}$96.00$ & \cellcolor{cyan!45}$72.22$ & \cellcolor{cyan!40}$65.38$ & \cellcolor{cyan!51}$82.50$ & \cellcolor{cyan!51}$82.24$ & \cellcolor{cyan!44}$70.65$ & \cellcolor{cyan!39}$63.58$ & \cellcolor{cyan!25}$41.00$ & \cellcolor{cyan!13}$22.04$ & \cellcolor{cyan!14}$23.61$ & \cellcolor{cyan!27}$44.15$ & \cellcolor{cyan!16}$25.70$ & \cellcolor{cyan!19}$31.99$ & \cellcolor{cyan!36}$58.49$ \\
\texttt{DeepSeek-V3} & \cellcolor{cyan!56}$90.91$ & \cellcolor{cyan!40}$65.00$ & \cellcolor{cyan!52}$84.00$ & \cellcolor{cyan!48}$77.78$ & \cellcolor{cyan!45}$73.08$ & \cellcolor{cyan!59}$95.00$ & \cellcolor{cyan!50}$80.26$ & \cellcolor{cyan!49}$78.64$ & \cellcolor{cyan!41}$66.00$ & \cellcolor{cyan!24}$38.60$ & \cellcolor{cyan!13}$21.98$ & \cellcolor{cyan!9}$15.14$ & \cellcolor{cyan!25}$40.27$ & \cellcolor{cyan!15}$24.38$ & \cellcolor{cyan!21}$35.17$ & \cellcolor{cyan!36}$59.08$ \\
\texttt{GPT-4o} & \cellcolor{cyan!42}$68.18$ & \cellcolor{cyan!37}$60.00$ & \cellcolor{cyan!55}$88.00$ & \cellcolor{cyan!48}$77.78$ & \cellcolor{cyan!45}$73.08$ & \cellcolor{cyan!46}$75.00$ & \cellcolor{cyan!47}$75.66$ & \cellcolor{cyan!48}$76.86$ & \cellcolor{cyan!37}$59.65$ & \cellcolor{cyan!23}$37.12$ & \cellcolor{cyan!8}$12.91$ & \cellcolor{cyan!4}$7.74$ & \cellcolor{cyan!18}$29.52$ & \cellcolor{cyan!13}$22.08$ & \cellcolor{cyan!16}$26.65$ & \cellcolor{cyan!32}$52.68$ \\
\texttt{GPT-4o-Mini} & \cellcolor{cyan!48}$77.27$ & \cellcolor{cyan!34}$55.00$ & \cellcolor{cyan!50}$80.00$ & \cellcolor{cyan!31}$50.00$ & \cellcolor{cyan!45}$73.08$ & \cellcolor{cyan!45}$72.50$ & \cellcolor{cyan!44}$71.05$ & \cellcolor{cyan!39}$63.89$ & \cellcolor{cyan!34}$55.87$ & \cellcolor{cyan!21}$34.20$ & \cellcolor{cyan!11}$17.89$ & \cellcolor{cyan!6}$9.95$ & \cellcolor{cyan!14}$23.75$ & \cellcolor{cyan!11}$18.24$ & \cellcolor{cyan!15}$24.68$ & \cellcolor{cyan!30}$48.49$ \\
\texttt{Qwen-Max} & \cellcolor{cyan!51}$81.82$ & \cellcolor{cyan!43}$70.00$ & \cellcolor{cyan!55}$88.00$ & \cellcolor{cyan!48}$77.78$ & \cellcolor{cyan!45}$73.08$ & \cellcolor{cyan!50}$80.00$ & \cellcolor{cyan!51}$82.24$ & \cellcolor{cyan!45}$72.53$ & \cellcolor{cyan!38}$62.32$ & \cellcolor{cyan!23}$37.88$ & \cellcolor{cyan!12}$19.68$ & \cellcolor{cyan!12}$19.78$ & \cellcolor{cyan!23}$37.57$ & \cellcolor{cyan!14}$23.91$ & \cellcolor{cyan!15}$24.76$ & \cellcolor{cyan!35}$56.76$ \\
\texttt{Qwen-2.5-0.5B-Instruct} & \cellcolor{cyan!11}$18.18$ & \cellcolor{cyan!3}$5.00$ & \cellcolor{cyan!2}$4.00$ & \cellcolor{cyan!0}$0.00$ & \cellcolor{cyan!4}$7.69$ & \cellcolor{cyan!6}$10.00$ & \cellcolor{cyan!10}$17.11$ & \cellcolor{cyan!11}$17.82$ & \cellcolor{cyan!20}$32.32$ & \cellcolor{cyan!3}$6.20$ & \cellcolor{cyan!2}$3.51$ & \cellcolor{cyan!3}$4.83$ & \cellcolor{cyan!3}$5.17$ & \cellcolor{cyan!3}$5.29$ & \cellcolor{cyan!12}$19.21$ & \cellcolor{cyan!6}$11.17$ \\
\texttt{Qwen-2.5-1.5B-Instruct} & \cellcolor{cyan!17}$27.27$ & \cellcolor{cyan!9}$15.00$ & \cellcolor{cyan!10}$16.00$ & \cellcolor{cyan!10}$16.67$ & \cellcolor{cyan!7}$11.54$ & \cellcolor{cyan!17}$27.50$ & \cellcolor{cyan!12}$19.74$ & \cellcolor{cyan!6}$9.88$ & \cellcolor{cyan!19}$31.44$ & \cellcolor{cyan!8}$13.20$ & \cellcolor{cyan!2}$3.76$ & \cellcolor{cyan!2}$3.66$ & \cellcolor{cyan!5}$8.15$ & \cellcolor{cyan!4}$7.17$ & \cellcolor{cyan!26}$41.72$ & \cellcolor{cyan!10}$16.85$ \\
\texttt{Qwen-2.5-3B-Instruct} & \cellcolor{cyan!28}$45.45$ & \cellcolor{cyan!18}$30.00$ & \cellcolor{cyan!17}$28.00$ & \cellcolor{cyan!6}$11.11$ & \cellcolor{cyan!7}$11.54$ & \cellcolor{cyan!17}$27.50$ & \cellcolor{cyan!17}$27.63$ & \cellcolor{cyan!10}$17.40$ & \cellcolor{cyan!24}$38.70$ & \cellcolor{cyan!11}$17.74$ & \cellcolor{cyan!4}$7.67$ & \cellcolor{cyan!4}$6.45$ & \cellcolor{cyan!5}$9.09$ & \cellcolor{cyan!6}$10.21$ & \cellcolor{cyan!22}$35.25$ & \cellcolor{cyan!13}$21.58$ \\
\texttt{Qwen-2.5-7B-Instruct} & \cellcolor{cyan!17}$27.27$ & \cellcolor{cyan!15}$25.00$ & \cellcolor{cyan!22}$36.00$ & \cellcolor{cyan!24}$38.89$ & \cellcolor{cyan!16}$26.92$ & \cellcolor{cyan!26}$42.50$ & \cellcolor{cyan!30}$48.68$ & \cellcolor{cyan!28}$45.19$ & \cellcolor{cyan!27}$43.42$ & \cellcolor{cyan!18}$28.97$ & \cellcolor{cyan!3}$4.92$ & \cellcolor{cyan!2}$4.70$ & \cellcolor{cyan!8}$12.94$ & \cellcolor{cyan!6}$10.66$ & \cellcolor{cyan!21}$34.53$ & \cellcolor{cyan!17}$28.71$ \\
\texttt{Qwen-2.5-14B-Instruct} & \cellcolor{cyan!39}$63.64$ & \cellcolor{cyan!34}$55.00$ & \cellcolor{cyan!37}$60.00$ & \cellcolor{cyan!41}$66.67$ & \cellcolor{cyan!38}$61.54$ & \cellcolor{cyan!43}$70.00$ & \cellcolor{cyan!35}$57.24$ & \cellcolor{cyan!33}$53.59$ & \cellcolor{cyan!31}$49.99$ & \cellcolor{cyan!20}$32.48$ & \cellcolor{cyan!2}$3.29$ & \cellcolor{cyan!2}$4.40$ & \cellcolor{cyan!11}$17.88$ & \cellcolor{cyan!8}$14.10$ & \cellcolor{cyan!6}$10.76$ & \cellcolor{cyan!25}$41.37$ \\
\texttt{Qwen-2.5-32B-Instruct} & \cellcolor{cyan!36}$59.09$ & \cellcolor{cyan!34}$55.00$ & \cellcolor{cyan!32}$52.00$ & \cellcolor{cyan!41}$66.67$ & \cellcolor{cyan!33}$53.85$ & \cellcolor{cyan!37}$60.00$ & \cellcolor{cyan!39}$63.16$ & \cellcolor{cyan!40}$64.10$ & \cellcolor{cyan!39}$63.12$ & \cellcolor{cyan!20}$32.53$ & \cellcolor{cyan!3}$5.41$ & \cellcolor{cyan!4}$6.94$ & \cellcolor{cyan!17}$28.66$ & \cellcolor{cyan!8}$13.81$ & \cellcolor{cyan!14}$22.87$ & \cellcolor{cyan!26}$43.15$ \\
\texttt{Qwen-2.5-72B-Instruct} & \cellcolor{cyan!39}$63.64$ & \cellcolor{cyan!46}$75.00$ & \cellcolor{cyan!55}$88.00$ & \cellcolor{cyan!45}$72.22$ & \cellcolor{cyan!43}$69.23$ & \cellcolor{cyan!50}$80.00$ & \cellcolor{cyan!43}$70.39$ & \cellcolor{cyan!42}$67.98$ & \cellcolor{cyan!38}$61.45$ & \cellcolor{cyan!21}$34.92$ & \cellcolor{cyan!1}$2.89$ & \cellcolor{cyan!1}$3.05$ & \cellcolor{cyan!5}$9.52$ & \cellcolor{cyan!11}$18.43$ & \cellcolor{cyan!20}$33.18$ & \cellcolor{cyan!31}$49.99$ \\
\texttt{Qwen-2.5-Coder-0.5B-Instruct} & \cellcolor{cyan!5}$9.09$ & \cellcolor{cyan!0}$0.00$ & \cellcolor{cyan!0}$0.00$ & \cellcolor{cyan!0}$0.00$ & \cellcolor{cyan!0}$0.00$ & \cellcolor{cyan!1}$2.50$ & \cellcolor{cyan!0}$1.32$ & \cellcolor{cyan!8}$13.88$ & \cellcolor{cyan!8}$13.18$ & \cellcolor{cyan!3}$5.62$ & \cellcolor{cyan!4}$6.66$ & \cellcolor{cyan!2}$4.42$ & \cellcolor{cyan!1}$2.15$ & \cellcolor{cyan!3}$5.19$ & \cellcolor{cyan!23}$37.75$ & \cellcolor{cyan!5}$9.25$ \\
\texttt{Qwen-2.5-Coder-1.5B-Instruct} & \cellcolor{cyan!2}$4.55$ & \cellcolor{cyan!3}$5.00$ & \cellcolor{cyan!10}$16.00$ & \cellcolor{cyan!0}$0.00$ & \cellcolor{cyan!9}$15.38$ & \cellcolor{cyan!3}$5.00$ & \cellcolor{cyan!1}$1.97$ & \cellcolor{cyan!17}$28.79$ & \cellcolor{cyan!24}$38.97$ & \cellcolor{cyan!9}$15.25$ & \cellcolor{cyan!0}$1.24$ & \cellcolor{cyan!2}$3.28$ & \cellcolor{cyan!6}$9.70$ & \cellcolor{cyan!6}$9.90$ & \cellcolor{cyan!21}$33.77$ & \cellcolor{cyan!8}$13.49$ \\
\texttt{Qwen-2.5-Coder-3B-Instruct} & \cellcolor{cyan!22}$36.36$ & \cellcolor{cyan!28}$45.00$ & \cellcolor{cyan!20}$32.00$ & \cellcolor{cyan!17}$27.78$ & \cellcolor{cyan!19}$30.77$ & \cellcolor{cyan!23}$37.50$ & \cellcolor{cyan!31}$50.00$ & \cellcolor{cyan!17}$27.43$ & \cellcolor{cyan!25}$40.84$ & \cellcolor{cyan!14}$23.31$ & \cellcolor{cyan!0}$1.28$ & \cellcolor{cyan!2}$4.06$ & \cellcolor{cyan!7}$12.00$ & \cellcolor{cyan!6}$9.64$ & \cellcolor{cyan!24}$38.41$ & \cellcolor{cyan!17}$27.76$ \\
\texttt{Qwen-2.5-Coder-7B-Instruct} & \cellcolor{cyan!34}$54.55$ & \cellcolor{cyan!18}$30.00$ & \cellcolor{cyan!22}$36.00$ & \cellcolor{cyan!27}$44.44$ & \cellcolor{cyan!31}$50.00$ & \cellcolor{cyan!26}$42.50$ & \cellcolor{cyan!36}$58.55$ & \cellcolor{cyan!31}$50.48$ & \cellcolor{cyan!33}$53.91$ & \cellcolor{cyan!19}$30.69$ & \cellcolor{cyan!2}$3.90$ & \cellcolor{cyan!3}$4.93$ & \cellcolor{cyan!9}$14.44$ & \cellcolor{cyan!8}$14.02$ & \cellcolor{cyan!15}$25.25$ & \cellcolor{cyan!21}$34.24$ \\
\texttt{Qwen-2.5-Coder-14B-Instruct} & \cellcolor{cyan!39}$63.64$ & \cellcolor{cyan!25}$40.00$ & \cellcolor{cyan!32}$52.00$ & \cellcolor{cyan!24}$38.89$ & \cellcolor{cyan!40}$65.38$ & \cellcolor{cyan!39}$62.50$ & \cellcolor{cyan!48}$77.63$ & \cellcolor{cyan!34}$55.01$ & \cellcolor{cyan!33}$52.87$ & \cellcolor{cyan!20}$32.29$ & \cellcolor{cyan!3}$5.99$ & \cellcolor{cyan!3}$5.01$ & \cellcolor{cyan!9}$15.76$ & \cellcolor{cyan!8}$13.39$ & \cellcolor{cyan!20}$32.62$ & \cellcolor{cyan!25}$40.87$ \\
\texttt{Qwen-2.5-Coder-32B-Instruct} & \cellcolor{cyan!42}$68.18$ & \cellcolor{cyan!46}$75.00$ & \cellcolor{cyan!45}$72.00$ & \cellcolor{cyan!34}$55.56$ & \cellcolor{cyan!40}$65.38$ & \cellcolor{cyan!43}$70.00$ & \cellcolor{cyan!48}$76.97$ & \cellcolor{cyan!40}$64.16$ & \cellcolor{cyan!35}$56.37$ & \cellcolor{cyan!21}$34.34$ & \cellcolor{cyan!2}$3.93$ & \cellcolor{cyan!4}$6.49$ & \cellcolor{cyan!12}$20.22$ & \cellcolor{cyan!11}$19.09$ & \cellcolor{cyan!14}$22.57$ & \cellcolor{cyan!29}$47.35$ \\
\texttt{LLaMA-3.1-8B-Instruct} & \cellcolor{cyan!8}$13.64$ & \cellcolor{cyan!12}$20.00$ & \cellcolor{cyan!12}$20.00$ & \cellcolor{cyan!3}$5.56$ & \cellcolor{cyan!16}$26.92$ & \cellcolor{cyan!14}$22.50$ & \cellcolor{cyan!18}$30.26$ & \cellcolor{cyan!21}$34.15$ & \cellcolor{cyan!29}$47.89$ & \cellcolor{cyan!13}$22.27$ & \cellcolor{cyan!2}$4.44$ & \cellcolor{cyan!2}$4.55$ & \cellcolor{cyan!6}$9.66$ & \cellcolor{cyan!7}$12.05$ & \cellcolor{cyan!8}$13.25$ & \cellcolor{cyan!11}$19.14$ \\
\texttt{LLaMA-3.1-70B-Instruct} & \cellcolor{cyan!34}$54.55$ & \cellcolor{cyan!21}$35.00$ & \cellcolor{cyan!25}$40.00$ & \cellcolor{cyan!13}$22.22$ & \cellcolor{cyan!33}$53.85$ & \cellcolor{cyan!21}$35.00$ & \cellcolor{cyan!42}$68.42$ & \cellcolor{cyan!31}$50.83$ & \cellcolor{cyan!38}$60.96$ & \cellcolor{cyan!18}$30.29$ & \cellcolor{cyan!5}$8.00$ & \cellcolor{cyan!3}$5.95$ & \cellcolor{cyan!11}$18.32$ & \cellcolor{cyan!8}$13.27$ & \cellcolor{cyan!20}$33.12$ & \cellcolor{cyan!22}$35.32$ \\
\texttt{LLaMA-3.3-70B-Instruct} & \cellcolor{cyan!39}$63.64$ & \cellcolor{cyan!21}$35.00$ & \cellcolor{cyan!25}$40.00$ & \cellcolor{cyan!17}$27.78$ & \cellcolor{cyan!28}$46.15$ & \cellcolor{cyan!21}$35.00$ & \cellcolor{cyan!30}$49.34$ & \cellcolor{cyan!37}$60.45$ & \cellcolor{cyan!39}$63.84$ & \cellcolor{cyan!20}$32.52$ & \cellcolor{cyan!3}$5.17$ & \cellcolor{cyan!3}$5.37$ & \cellcolor{cyan!7}$11.56$ & \cellcolor{cyan!8}$12.99$ & \cellcolor{cyan!20}$33.11$ & \cellcolor{cyan!21}$34.80$ \\
\bottomrule
\end{tabular}}
\caption{Performance of different models under different prompting strategies on \bench.}
\label{tab:main2}
\end{table*}

\clearpage

\section{Prompts}

\subsection{Prompts for Dataset Refactoring}

\lstset{
    numbers=none,
    keywordstyle= \color{ blue!70},
    commentstyle= \color{red!50!green!50!blue!50},
    frame=none,
    rulesepcolor= \color{ red!20!green!20!blue!20} ,
    framexleftmargin=2em,
    columns=fullflexible,
    breaklines=true,
    breakindent=0pt,
    basicstyle=\rmfamily
}

\paragraph{ML:}

\begin{tcolorbox}[left=0mm,right=0mm,top=0mm,bottom=0mm,boxsep=1mm,arc=0mm,boxrule=0pt, frame empty, breakable]
    \small
    \begin{lstlisting}
I will provide you with a code problem with a solution. You need to generate a complete, executable code based on the raw json data, including all necessary package imports, the original code, the test cases, and the main function. 
You need to generate the executable code and expected result.
Please choose a test case according to the 'test' field from raw json data, and the code should print the answer of the test case.
The output should be json format, with code and expected_result fields.
Please only generate the number or string answer in 'expected_result' field without any extra description.
\end{lstlisting}
\end{tcolorbox}

\paragraph{CL:}

\begin{tcolorbox}[left=0mm,right=0mm,top=0mm,bottom=0mm,boxsep=1mm,arc=0mm,boxrule=0pt, frame empty, breakable]
    \small
    \begin{lstlisting}
I will provide you with the solution to a code problem in cpp, python, and javascript. You need to score according to the difficulty of the problem from 1 to 5, while 5 means the hardest. And generate topic keywords for the problem.
The output should only be json format, with difficulty and keywords fields.
difficulty: 1-5, integer
keywords: two or three words to best describe the problem, string list
\end{lstlisting}
\end{tcolorbox}

\paragraph{BG:}

\begin{tcolorbox}[left=0mm,right=0mm,top=0mm,bottom=0mm,boxsep=1mm,arc=0mm,boxrule=0pt, frame empty, breakable]
    \small
    \begin{lstlisting}
I will provide you with a piece of code and some test cases. You need to generate a complete, executable code based on these, including all necessary package imports, the original code, the test cases, and the main function. You should wrap the original code with ORIGINAL_CODE_START and ORIGINAL_CODE_END comments. Additionally, the program should output the results of the test cases. Do not include expected output in your answer.
\end{lstlisting}
\end{tcolorbox}

\section{Details of \bench}

\subsection{ML}

\begin{table}[!ht]
  \centering

  \begin{tabular}{lccccccc}
      \toprule
      Java & C\# & Rust & Julia & Python & C++ & C \\
      \midrule
      25   & 20  & 20   & 26    & 18     & 21  & 20 \\
      \bottomrule
  \end{tabular}

  \caption{Language usage count across different categories in the ML subset.}
  \label{tab:ml_language_usage}
  
\end{table}

In ML, the usage distribution of various programming languages is shown in Table \ref{tab:ml_language_usage}. We selected a variety of languages, including Java, C\#, Rust, Julia, Python, C++, and C, to evaluate the model's ability to handle multilingual code. This diverse selection helps to comprehensively assess the model's performance across different languages.

\subsubsection{System Prompts}

\paragraph{Zero-shot Chain-of-Thought:}

\begin{tcolorbox}[left=0mm,right=0mm,top=0mm,bottom=0mm,boxsep=1mm,arc=0mm,boxrule=0pt, frame empty, breakable]
    \small
    \begin{lstlisting}
Given the following code, what is the execution result?
You should think step by step.  Your answer should be in the following format:
Thought: <your thought>
Output:
<execution result>
\end{lstlisting}
\end{tcolorbox}

\paragraph{Zero-shot:}

\begin{tcolorbox}[left=0mm,right=0mm,top=0mm,bottom=0mm,boxsep=1mm,arc=0mm,boxrule=0pt, frame empty, breakable]
    \small
    \begin{lstlisting}
Given the following code, what is the execution result?
Your answer should be in the following format:
Output:
<execution result>
\end{lstlisting}
\end{tcolorbox}

\paragraph{Few-shot Chain-of-Thought:}

\begin{tcolorbox}[left=0mm,right=0mm,top=0mm,bottom=0mm,boxsep=1mm,arc=0mm,boxrule=0pt, frame empty, breakable]
    \small
    \begin{lstlisting}
Given the following code, what is the execution result?
You should think step by step.  Your answer should be in the following format:
Thought: <your thought>
Output:
<execution result>
Following are 3 examples: 
{{examples here}}

\end{lstlisting}
\end{tcolorbox}

\subsubsection{Demo Questions}

\begin{tcolorbox}[left=0mm,right=0mm,top=0mm,bottom=0mm,boxsep=1mm,arc=0mm,boxrule=0pt, frame empty, breakable]
    \small
    \begin{lstlisting}
def catalan_number(n: int) -> int:
    # Initialize an array to store the intermediate catalan numbers
    catalan = [0] * (n + 1)
    catalan[0] = 1  # Base case

    # Calculate catalan numbers using the recursive formula
    for i in range(1, n + 1):
        for j in range(i):
            catalan[i] += catalan[j] * catalan[i - j - 1]

    return catalan[n]

if __name__ == "__main__":
    # Run the test function and print the result of a specific test case
    print(catalan_number(3))
\end{lstlisting}
\end{tcolorbox}

\begin{tcolorbox}[left=0mm,right=0mm,top=0mm,bottom=0mm,boxsep=1mm,arc=0mm,boxrule=0pt, frame empty, breakable]
    \small
    \begin{lstlisting}
import java.util.*;

class Solution {
    public static int countPrefixWords(List<String> wordList, String prefix) {

        int count = 0;
        for (String word : wordList) {
            if (word.startsWith(prefix)) {
                count++;
            }
        }
        return count;
    }

    public static void main(String[] args) {
        System.out.println(countPrefixWords(Arrays.asList("dog", "dodge", "dot", "dough"), "do"));
    }
}
\end{lstlisting}
\end{tcolorbox}

\begin{tcolorbox}[left=0mm,right=0mm,top=0mm,bottom=0mm,boxsep=1mm,arc=0mm,boxrule=0pt, frame empty, breakable]
    \small
    \begin{lstlisting}
#include <assert.h>
#include <stdio.h>

long long minTotalCost(int n, int *C)
{
   return (long long)(C[n-2]) * (n - 1) + C[n-1];
}

int main() {
    int costs3[] = {5, 4, 3, 2};
    printf("%lld\n", minTotalCost(4, costs3));
    return 0;
}
\end{lstlisting}
\end{tcolorbox}

\begin{tcolorbox}[left=0mm,right=0mm,top=0mm,bottom=0mm,boxsep=1mm,arc=0mm,boxrule=0pt, frame empty, breakable]
    \small
    \begin{lstlisting}
function merge_sorted_arrays(nums1::Vector{Int}, m::Int, nums2::Vector{Int}, n::Int) :: Vector{Int}
    i = m 
    j = n 
    k = m + n
    
    while j > 0
        if i > 0 && nums1[i] > nums2[j]
            nums1[k] = nums1[i]
            i -= 1
        else
            nums1[k] = nums2[j]
            j -= 1
        end
        k -= 1
    end
    
    nums1
end

# Test case
result = merge_sorted_arrays([1, 3, 5, 0, 0, 0], 3, [2, 4, 6], 3)
println(result)
\end{lstlisting}
\end{tcolorbox}

\begin{tcolorbox}[left=0mm,right=0mm,top=0mm,bottom=0mm,boxsep=1mm,arc=0mm,boxrule=0pt, frame empty, breakable]
    \small
    \begin{lstlisting}
public class Solution {

  public static int findSmallestInteger(int n) {
    char[] characters = Integer.toString(n).toCharArray();
    int i = characters.length - 2;

    // Find the first digit that is smaller than the digit next to it.
    while (i >= 0 && characters[i] >= characters[i + 1]) {
      i--;
    }

    if (i == -1) {
      return -1; // Digits are in descending order, no greater number possible.
    }

    // Find the smallest digit on right side of (i) which is greater than characters[i]
    int j = characters.length - 1;
    while (characters[j] <= characters[i]) {
      j--;
    }

    // Swap the digits at indices i and j
    swap(characters, i, j);

    // Reverse the digits from index i+1 to the end of the array
    reverse(characters, i + 1);

    try {
      return Integer.parseInt(new String(characters));
    } catch (NumberFormatException e) {
      return -1; // The number formed is beyond the range of int.
    }
  }

  private static void swap(char[] arr, int i, int j) {
    char temp = arr[i];
    arr[i] = arr[j];
    arr[j] = temp;
  }

  private static void reverse(char[] arr, int start) {
    int end = arr.length - 1;
    while (start < end) {
      swap(arr, start, end);
      start++;
      end--;
    }
  }

  public static void main(String[] args) {
    System.out.println(findSmallestInteger(123));
  }
}
\end{lstlisting}
\end{tcolorbox}
\subsection{CL}

\begin{table}[!ht]
  \centering

  \begin{tabular}{lccc}
      \toprule
      Python & C++ & JavaScript \\
      \midrule
       50     & 51  & 49         \\
      \bottomrule
  \end{tabular}
  \caption{Language usage count across different categories in the CL subset.}
  \label{tab:cl_language_usage}
\end{table}

\begin{table}[!ht]
  \centering

      \begin{tabular}{cccc}
\toprule
\textbf{Difficulty} & JavaScript & CPP & Python \\
\midrule
1 & 10 & 11 & 11 \\
2 & 6 & 4 & 6 \\
3 & 12 & 14 & 12 \\
4 & 8 & 8 & 9 \\
5 & 13 & 14 & 12 \\
\bottomrule
\end{tabular}

\caption{Details of problems in different languages and different difficulty levels.}
\label{tab:stat:2}

\end{table}

In CL, we selected competition problems of varying difficulty, each with solutions in Python, C++, and JavaScript. You can see the distribution of language in Table \ref{tab:cl_language_usage}, and the distribution of problem difficulty in Table \ref{tab:stat:2}. This selection allows us to test the model's cross-language capabilities and its ability to handle problems of different difficulty levels.

\subsubsection{System Prompts}

\paragraph{Zero-shot Chain-of-Thought:}

\begin{tcolorbox}[left=0mm,right=0mm,top=0mm,bottom=0mm,boxsep=1mm,arc=0mm,boxrule=0pt, frame empty, breakable]
    \small
    \begin{lstlisting}
Given the following code, what is the execution result?
You should think step by step.  Your answer should be in the following format:
Thought: <your thought>
Output:
<execution result>
\end{lstlisting}
\end{tcolorbox}

\paragraph{Zero-shot:}

\begin{tcolorbox}[left=0mm,right=0mm,top=0mm,bottom=0mm,boxsep=1mm,arc=0mm,boxrule=0pt, frame empty, breakable]
    \small
    \begin{lstlisting}
Given the following code, what is the execution result?
Your answer should be in the following format:
Output:
<execution result>
\end{lstlisting}
\end{tcolorbox}

\paragraph{Few-shot Chain-of-Thought:}

\begin{tcolorbox}[left=0mm,right=0mm,top=0mm,bottom=0mm,boxsep=1mm,arc=0mm,boxrule=0pt, frame empty, breakable]
    \small
    \begin{lstlisting}
Given the following code, what is the execution result?
You should think step by step.  Your answer should be in the following format:
Thought: <your thought>
Output:
<execution result>
Following are 3 examples: 
{{examples here}}

\end{lstlisting}
\end{tcolorbox}

\subsubsection{Demo Questions}

\begin{tcolorbox}[left=0mm,right=0mm,top=0mm,bottom=0mm,boxsep=1mm,arc=0mm,boxrule=0pt, frame empty, breakable]
    \small
    \begin{lstlisting}
class TreeNode {
  constructor(val) {
    this.val = val;
    this.left = this.right = null;
  }
}

function maxDepth(root) {
  if (!root) return 0;
  const queue = [root, null];
  let depth = 1;

  while (queue.length > 0) {
    const node = queue.shift();
    if (node === null) {
      if (queue.length === 0) return depth;
      depth++;
      queue.push(null);
      continue;
    }
    if (node.left) queue.push(node.left);
    if (node.right) queue.push(node.right);
  }

  return depth;
}

// Test case
const root = new TreeNode(3);
root.left = new TreeNode(9);
root.right = new TreeNode(20);
root.right.left = new TreeNode(15);
root.right.right = new TreeNode(7);
console.log(maxDepth(root));
\end{lstlisting}
\end{tcolorbox}

\begin{tcolorbox}[left=0mm,right=0mm,top=0mm,bottom=0mm,boxsep=1mm,arc=0mm,boxrule=0pt, frame empty, breakable]
    \small
    \begin{lstlisting}
from collections import Counter
class Solution:
    def maxScoreWords(self, words, letters, score):
        self.ans = 0
        words_score = [sum(score[ord(c)-ord('a')] for c in word) for word in words]
        words_counter = [Counter(word) for word in words]

        def backtrack(start, cur, counter):
            if start > len(words):
                return
            self.ans = max(self.ans, cur)
            for j, w_counter in enumerate(words_counter[start:], start):
                if all(n <= counter.get(c,0) for c,n in w_counter.items()):
                    backtrack(j+1, cur+words_score[j], counter-w_counter)

        backtrack(0, 0, Counter(letters))
        return self.ans

solution = Solution()
print(solution.maxScoreWords(["dog","cat","dad","good"], ["a","a","c","d","d","d","g","o","o"], [1,0,9,5,0,0,3,0,0,0,0,0,0,0,
2,0,0,0,0,0,0,0,0,0,0,0]))
\end{lstlisting}
\end{tcolorbox}

\begin{tcolorbox}[left=0mm,right=0mm,top=0mm,bottom=0mm,boxsep=1mm,arc=0mm,boxrule=0pt, frame empty, breakable]
    \small
    \begin{lstlisting}
#include <iostream>
#include <unordered_map>
#include <string>
using namespace std;

int findTheLongestSubstring(string s) {
    unordered_map<char, int> mapper = {{'a', 1}, {'e', 2}, {'i', 4}, {'o', 8}, {'u', 16}};
    unordered_map<int, int> seen;
    seen[0] = -1;
    int max_len = 0, cur = 0;

    for(int i = 0; i < s.size(); ++i){
        if(mapper.find(s[i]) != mapper.end()){
            cur ^= mapper[s[i]];
        }
        if(seen.find(cur) != seen.end()){
            max_len = max(max_len, i - seen[cur]);
        } else {
            seen[cur] = i;
        }
    }

    return max_len;
}

// Test case
class Solution {
public:
    void solve() {
        string input = "eleetminicoworoep";
        cout << findTheLongestSubstring(input) << endl; // Expected output: 13
    }
};

int main(){
    Solution sol;
    sol.solve();
    return 0;
}
\end{lstlisting}
\end{tcolorbox}

\begin{tcolorbox}[left=0mm,right=0mm,top=0mm,bottom=0mm,boxsep=1mm,arc=0mm,boxrule=0pt, frame empty, breakable]
    \small
    \begin{lstlisting}
class TreeNode {
    constructor(val) {
        this.val = val;
        this.left = this.right = null;
    }
}

function backtrack(root, sum, res, tempList) {
    if (root === null) return;
    if (root.left === null && root.right === null && sum === root.val)
        return res.push([...tempList, root.val]);

    tempList.push(root.val);
    backtrack(root.left, sum - root.val, res, tempList);
    backtrack(root.right, sum - root.val, res, tempList);
    tempList.pop();
}

function pathSum(root, sum) {
    if (root === null) return [];
    const res = [];
    backtrack(root, sum, res, []);
    return res;
}

// Test case setup
const root = new TreeNode(5);
root.left = new TreeNode(4);
root.right = new TreeNode(8);
root.left.left = new TreeNode(11);
root.right.left = new TreeNode(13);
root.right.right = new TreeNode(4);
root.left.left.left = new TreeNode(7);
root.left.left.right = new TreeNode(2);
root.right.right.left = new TreeNode(5);
root.right.right.right = new TreeNode(1);

console.log(pathSum(root, 22));
\end{lstlisting}
\end{tcolorbox}

\begin{tcolorbox}[left=0mm,right=0mm,top=0mm,bottom=0mm,boxsep=1mm,arc=0mm,boxrule=0pt, frame empty, breakable]
    \small
    \begin{lstlisting}
class TrieNode {
    constructor() {
        this.children = {};
        this.isEndOfWord = false;
    }
}

class Trie {
    constructor() {
        this.root = new TrieNode();
    }

    insert(word) {
        let node = this.root;
        for (let char of word) {
            if (!node.children[char]) {
                node.children[char] = new TrieNode();
            }
            node = node.children[char];
        }
        node.isEndOfWord = true;
    }

    search(stream) {
        let node = this.root;
        for (let char of stream) {
            if (!node.children[char]) {
                return false;
            }
            node = node.children[char];
            if (node.isEndOfWord) {
                return true;
            }
        }
        return false;
    }
}

class StreamChecker {
    constructor(words) {
        this.trie = new Trie();
        this.stream = [];

        for (let word of [...new Set(words)]) {
            this.trie.insert(word.split('').reverse().join(''));
        }
    }

    query(letter) {
        this.stream.unshift(letter);
        return this.trie.search(this.stream);
    }
}

// Test case
const streamChecker = new StreamChecker(["cd", "f", "kl"]);
console.log(streamChecker.query('a')); // false
console.log(streamChecker.query('b')); // false
console.log(streamChecker.query('c')); // false
console.log(streamChecker.query('d')); // true
console.log(streamChecker.query('e')); // false
console.log(streamChecker.query('f')); // true
console.log(streamChecker.query('g')); // false
console.log(streamChecker.query('h')); // false
console.log(streamChecker.query('i')); // false
console.log(streamChecker.query('j')); // false
console.log(streamChecker.query('k')); // false
console.log(streamChecker.query('l')); // true
\end{lstlisting}
\end{tcolorbox}
\subsection{RL}

\begin{table}[!ht]
  \centering

  \begin{tabular}{lcc}
      \toprule
       Python & C++  \\
      \midrule
       24     & 36   \\
      \bottomrule
  \end{tabular}

  \caption{Language usage count across different categories in the RL subset.}
  \label{tab:rl_language_usage}
  
\end{table}

In RL, the distribution of programming language usage is shown in Table \ref{tab:rl_language_usage}. We utilized five GitHub repositories for this study, consisting of two Python projects and three C++ projects. Each repository contains a set of ten or more test cases, providing a diverse set of data for evaluation across different programming languages.

\subsubsection{System Prompts}

\paragraph{Zero-shot Chain-of-Thought:}

\begin{tcolorbox}[left=0mm,right=0mm,top=0mm,bottom=0mm,boxsep=1mm,arc=0mm,boxrule=0pt, frame empty, breakable]
    \small
    \begin{lstlisting}
You will be given a github repository and a function that generates a latex file with this repo. Your task is to predict the content of the latex file generated by the function.
You should think step by step.  Your answer should be in the following format:
Thought: <your thought>
Output:
<file content>
\end{lstlisting}
\end{tcolorbox}

\paragraph{Zero-shot:}

\begin{tcolorbox}[left=0mm,right=0mm,top=0mm,bottom=0mm,boxsep=1mm,arc=0mm,boxrule=0pt, frame empty, breakable]
    \small
    \begin{lstlisting}
You will be given a github repository and a function that generates a latex file with this repo. Your task is to predict the content of the latex file generated by the function.
Your answer should be in the following format:
Output:
<file content>
\end{lstlisting}
\end{tcolorbox}

\paragraph{Few-shot Chain-of-Thought:}

\begin{tcolorbox}[left=0mm,right=0mm,top=0mm,bottom=0mm,boxsep=1mm,arc=0mm,boxrule=0pt, frame empty, breakable]
    \small
    \begin{lstlisting}
You will be given a github repository and a function that generates a latex file with this repo. Your task is to predict the content of the latex file generated by the function.
You should think step by step.  Your answer should be in the following format:
Thought: <your thought>
Output:
<file content>
Following is one example:  
{{examples here}}

\end{lstlisting}
\end{tcolorbox}

\subsubsection{Demo Questions}

\begin{tcolorbox}[left=0mm,right=0mm,top=0mm,bottom=0mm,boxsep=1mm,arc=0mm,boxrule=0pt, frame empty, breakable]
    \small
    \begin{lstlisting}
main.cpp:<start_file>#include <iostream>
#include <vector>
#include <utility>
#include <stdlib.h>
#include <time.h>
#include <unistd.h>
#include <fstream>
#include <map>
using namespace std;

typedef vector<vector<char> > Board;

const int N = 9;

class SudokuPlayer
{
private:
    int rowUsed[N];
    int columnUsed[N];
    int blockUsed[N];

public:
    vector<Board> result;
    vector<pair<int, int> > spaces;

public:
    SudokuPlayer()
    {
        initState();
    }

    void initState()
    {
        memset(rowUsed, 0, sizeof(rowUsed));
        memset(columnUsed, 0, sizeof(columnUsed));
        memset(blockUsed, 0, sizeof(blockUsed));
        spaces.clear();
        result.clear();
    }

    void addResult(Board &board)
    {
        vector<vector<char> > obj(board);
        result.push_back(obj);
    }

    void flip(int i, int j, int digit)
    {
        rowUsed[i] ^= (1 << digit);
        columnUsed[j] ^= (1 << digit);
        blockUsed[(i / 3) * 3 + j / 3] ^= (1 << digit);
    }

    vector<Board> solveSudoku(Board board)
    {
        initState();
        for (int i = 0; i < N; i++)
        {
            for (int j = 0; j < N; j++)
            {
                if (board[i][j] == '$')
                {
                    spaces.push_back(pair<int, int>(i, j));
                }
                else
                {
                    int digit = board[i][j] - '1';
                    flip(i, j, digit);
                }
            }
        }
        DFS(board, 0);
        return result;
    }

    void DFS(Board &board, int pos)
    {
        if (pos == spaces.size())
        {
            addResult(board);
            return;
        }
        int i = spaces[pos].first;
        int j = spaces[pos].second;
        int mask = ~(rowUsed[i] | columnUsed[j] | blockUsed[(i / 3) * 3 + j / 3]) & 0x1ff;
        int digit = 0;
        while (mask)
        {
            if (mask & 1)
            {
                flip(i, j, digit);
                board[i][j] = '1' + digit;
                DFS(board, pos + 1);
                flip(i, j, digit);
            }
            mask = mask >> 1;
            digit++;
        }
    }

    void getResult()
    {
        for (size_t i = 0; i < result.size(); i++)
        {
            Board board = result[i];
            printBoard(board);
        }
    }

    bool checkBoard(Board &board)
    {
        initState();
        for (int i = 0; i < 9; i++)
        {
            for (int j = 0; j < 9; j++)
            {
                if (board[i][j] != '$')
                {
                    int digit = board[i][j] - '1';
                    if ((rowUsed[i] | columnUsed[j] | blockUsed[(i / 3) * 3 + j / 3]) & (1 << digit))
                    {
                        return false;
                    }
                    flip(i, j, digit);
                }
            }
        }
        return true;
    }

    void printBoard(Board &board)
    {
        for (int i = 0; i < board.size(); i++)
        {
            for (int j = 0; j < board[i].size(); j++)
            {
                cout << board[i][j] << " ";
            }
            cout << "\n";
        }
    }

    Board generateBoard(int digCount)
    {
        vector<vector<char> > board(N, vector<char>(N, '$'));
        vector<int> row = getRand9();
        for (int i = 0; i < 3; i++)
        {
            board[3][i + 3] = row[i] + '1';
            board[4][i + 3] = row[i + 3] + '1';
            board[5][i + 3] = row[i + 6] + '1';
        }
        copySquare(board, 3, 3, true);
        copySquare(board, 3, 3, false);
        copySquare(board, 3, 0, false);
        copySquare(board, 3, 6, false);

        while (digCount)
        {
            int x = rand() % 9;
            int y = rand() % 9;
            if (board[x][y] == '$')
                continue;
            char tmp = board[x][y];
            board[x][y] = '$';

            solveSudoku(board);
            if (result.size() == 1)
            {
                digCount--;
            }
            else
            {
                board[x][y] = tmp;
            }
        }
        // printBoard(board);
        // cout << "spaces " << player.spaces.size() << "\n";
        if (!checkBoard(board))
        {
            cout << "wrong board" << endl;
        }

        return board;
    }

    vector<int> getRand9()
    {
        vector<int> result;
        int digit = 0;
        while (result.size() != 9)
        {
            int num = rand() % 9;
            if ((1 << num) & digit)
            {
                continue;
            }
            else
            {
                result.push_back(num);
                digit ^= (1 << num);
            }
        }
        return result;
    }

    void copySquare(Board &board, int src_x, int src_y, bool isRow)
    {
        int rand_tmp = rand() % 2 + 1;
        int order_first[3] = {1, 2, 0};
        int order_second[3] = {2, 0, 1};
        if (rand_tmp == 2)
        {
            order_first[0] = 2;
            order_first[1] = 0;
            order_first[2] = 1;
            order_second[0] = 1;
            order_second[1] = 2;
            order_second[2] = 0;
        }
        for (int i = 0; i < 3; i++)
        {
            if (isRow)
            {
                board[src_x][i] = board[src_x + order_first[0]][src_y + i];
                board[src_x + 1][i] = board[src_x + order_first[1]][src_y + i];
                board[src_x + 2][i] = board[src_x + order_first[2]][src_y + i];
                board[src_x][i + 6] = board[src_x + order_second[0]][src_y + i];
                board[src_x + 1][i + 6] = board[src_x + order_second[1]][src_y + i];
                board[src_x + 2][i + 6] = board[src_x + order_second[2]][src_y + i];
            }
            else
            {
                board[i][src_y] = board[src_x + i][src_y + order_first[0]];
                board[i][src_y + 1] = board[src_x + i][src_y + order_first[1]];
                board[i][src_y + 2] = board[src_x + i][src_y + order_first[2]];
                board[i + 6][src_y] = board[src_x + i][src_y + order_second[0]];
                board[i + 6][src_y + 1] = board[src_x + i][src_y + order_second[1]];
                board[i + 6][src_y + 2] = board[src_x + i][src_y + order_second[2]];
            }
        }
    }
};

char data[9][9] = {
    {'5', '3', '.', '.', '7', '.', '.', '.', '.'},
    {'6', '.', '.', '1', '9', '5', '.', '.', '.'},
    {'.', '9', '8', '.', '.', '.', '.', '6', '.'},
    {'8', '.', '.', '.', '6', '.', '.', '.', '3'},
    {'4', '.', '.', '8', '.', '3', '.', '.', '1'},
    {'7', '.', '.', '.', '2', '.', '.', '.', '6'},
    {'.', '6', '.', '.', '.', '.', '2', '8', '.'},
    {'.', '.', '.', '4', '1', '9', '.', '.', '5'},
    {'.', '.', '.', '.', '8', '.', '.', '7', '9'}};

void test()
{
    SudokuPlayer player;
    vector<vector<char> > board(N, vector<char>(N, '.'));

    for (int i = 0; i < board.size(); i++)
    {
        for (int j = 0; j < board[i].size(); j++)
        {
            board[i][j] = data[i][j];
        }
    }
    bool check = player.checkBoard(board);
    if (check)
        cout << "checked" << endl;

    player.solveSudoku(board);
    player.getResult();

    cout << endl;
}

vector<Board> readFile(string filePath)
{
    ifstream infile;
    vector<Board> boards;
    infile.open(filePath);
    char data[100];
    Board tmp;
    vector<char> row;
    while (!infile.eof())
    {
        infile.getline(data, 100);
        if (data[0] == '-')
        {
            boards.push_back(Board(tmp));
            tmp.clear();
            continue;
        }
        for (int i = 0; i < strlen(data); i++)
        {
            if (('1' <= data[i] && data[i] <= '9') || data[i] == '$')
            {
                row.push_back(data[i]);
            }
        }
        tmp.push_back(vector<char>(row));
        row.clear();
    }
    infile.close();
    return boards;
}

void writeFile(vector<Board> boards, ofstream &f)
{
    for (int k = 0; k < boards.size(); k++)
    {
        for (int i = 0; i < boards[k].size(); i++)
        {
            for (int j = 0; j < boards[k][i].size(); j++)
            {
                f << boards[k][i][j] << " ";
            }
            f << "\n";
        }
        f << "------- " << k << " -------" << endl;
    }
}

map<char, string> parse(int argc, char *argv[])
{
    map<char, string> params;
    int compeleteBoardCount, gameNumber, gameLevel;
    vector<int> range;
    string inputFile;
    char opt = 0;
    while ((opt = getopt(argc, argv, "c:s:n:m:r:u")) != -1)
    {
        switch (opt)
        {
        case 'c':
            compeleteBoardCount = atoi(optarg);
            if (compeleteBoardCount < 1 || compeleteBoardCount > 1000000)
            {
                exit(0);
            }
            params[opt] = string(optarg);
            break;
        case 's':
            inputFile = string(optarg);
            if (access(optarg, 0) == -1)
            {
                printf("file does not exist\n");
                exit(0);
            }
            params[opt] = string(optarg);
            break;
        case 'n':
            gameNumber = atoi(optarg);
            if (gameNumber < 1 || gameNumber > 10000)
            {
                exit(0);
            }
            params[opt] = string(optarg);
            break;
        case 'm':
            gameLevel = atoi(optarg);
            if (gameLevel < 1 || gameLevel > 3)
            {
                exit(0);
            }
            params[opt] = string(optarg);
            break;
        case 'r':
            char *p;
            p = strtok(optarg, "~");
            while (p)
            {
                range.push_back(atoi(p));
                p = strtok(NULL, "~");
            }
            if (range.size() != 2)
            {
                exit(0);
            }
            if ((range[0] >= range[1]) || range[0] < 20 || range[1] > 55)
            {
                exit(0);
            }
            params[opt] = string(optarg);
            break;
        case 'u':
            params[opt] = string();
            break;
        default:
            exit(0);
            break;
        }
    }
    return params;
}

void generateGame(int gameNumber, int gameLevel, vector<int> digCount, ofstream &outfile, SudokuPlayer &player)
{
    for (int i = 0; i < gameNumber; i++)
    {
        int cnt = 0;
        if (digCount.size() == 1)
        {
            cnt = digCount[0];
        }
        else
        {
            cnt = rand() % (digCount[1] - digCount[0] + 1) + digCount[0];
        }
        Board b = player.generateBoard(cnt);
        vector<Board> bs;
        bs.push_back(b);
        writeFile(bs, outfile);
    }
    outfile.close();
}

int main(int argc, char *argv[])
{
    srand((unsigned)time(NULL));
    SudokuPlayer player;

    map<char, string> params = parse(argc, argv);
    map<char, string>::iterator it, tmp;

    int opt = 0;

    vector<int> range;
    int gameNumber;
    int gameLevel = 0;
    int solution_count = 0;

    vector<Board> boards;
    ofstream outfile;

    it = params.begin();
    while (it != params.end())
    {
        switch (it->first)
        {
        case 'c':
            outfile.open("game.txt", ios::out | ios::trunc);
            range.push_back(0);
            generateGame(atoi(it->second.c_str()), 0, range, outfile, player);
            range.clear();
            break;

        case 's':
            outfile.open("sudoku.txt", ios::out | ios::trunc);
            boards = readFile(it->second);
            for (int i = 0; i < boards.size(); i++)
            {
                vector<Board> result = player.solveSudoku(boards[i]);
                writeFile(result, outfile);
            }
            outfile.close();
            break;

        case 'n':
        case 'm':
        case 'r':
        case 'u':
            tmp = params.find('n');
            if (tmp == params.end())
            {
                exit(0);
            }

            gameNumber = atoi(tmp->second.c_str());

            tmp = params.find('u');
            if (tmp != params.end())
            {
                solution_count = 1;
            }

            tmp = params.find('m');
            if (tmp != params.end())
            {
                gameLevel = atoi(tmp->second.c_str());
            }

            tmp = params.find('r');
            if (tmp != params.end())
            {
                char *p;
                char *pc = new char[100];
                strcpy(pc, tmp->second.c_str());
                p = strtok(pc, "~");
                while (p)
                {
                    range.push_back(atoi(p));
                    p = strtok(NULL, "~");
                }
            }
            else
            {
                if (gameLevel == 1)
                {
                    range.push_back(20);
                    range.push_back(30);
                }
                else if (gameLevel == 2)
                {
                    range.push_back(30);
                    range.push_back(40);
                }
                else if (gameLevel == 3)
                {
                    range.push_back(40);
                    range.push_back(55);
                }
                else
                {
                    range.push_back(20);
                    range.push_back(55);
                }
            }

            outfile.open("game.txt", ios::out | ios::trunc);
            generateGame(gameNumber, gameLevel, range, outfile, player);
            range.clear();
            break;
        }
        // cout << it->first << ' ' << it->second << endl;
        it++;
    }

    return 0;
}<end_file>;game.txt:<start_file><9 $ 5 $ 3 $ 7 1 2 
$ 1 2 $ $ 8 3 $ $ 
$ $ $ 2 7 $ 9 8 5 
8 $ 9 $ 6 $ 1 2 7 
1 $ $ $ 5 $ $ 6 3 
4 6 3 1 2 7 $ $ $ 
$ $ 8 3 4 6 2 7 1 
2 7 $ $ $ $ $ 3 $ 
$ 3 4 $ 1 $ $ $ 8 
------- 0 -------<endfile>
\end{lstlisting}
\end{tcolorbox}

\begin{tcolorbox}[left=0mm,right=0mm,top=0mm,bottom=0mm,boxsep=1mm,arc=0mm,boxrule=0pt, frame empty, breakable]
    \small
    \begin{lstlisting}
Here is the code repository:Cow.cpp:<start_file>#include "Cow.h"
Cow::Cow(std::string a,int b,int c,int d){
    name=a;
    l=b;
    u=c;
    m=d;
    in=0;
    state=0;
}<endfile>Cow.h:<start_file>#pragma once
#include <string>
class Cow{
    public:
    std::string name;
    int l,u,m;
    int in;
    int state;
    Cow(){}
    Cow(std::string a,int b,int c,int d);
};<endfile>Farm.cpp:<start_file>#include "Farm.h"
Farm::Farm(int a){
    n=a;
    num=0;
    cow=new Cow[a];
    milk=0;
}
void Farm::addCow(Cow a){
        cow[num]=a;
        num+=1;
    }
void Farm::supply(std::string a,int b){
    for(int i=0;i<n;i++){
        if(cow[i].name==a){
            cow[i].in+=b;
            break;
        }
    }
}
void Farm::startMeal(){
    for(int i=0;i<n;i++){
        if(cow[i].in==0)
        cow[i].state=0;
        if(cow[i].in>0&&cow[i].in<cow[i].l){
            cow[i].state=1;
            cow[i].in=0;
        }
        if(cow[i].in>=cow[i].l){
            cow[i].state=2;
            if(cow[i].in<=cow[i].u)
            cow[i].in=0;
            if(cow[i].in>cow[i].u)
            cow[i].in-=cow[i].u;
        }
    }
}
void Farm::produceMilk(){
    for(int i=0;i<n;i++){
        if(cow[i].state==0){
            milk+=0;
            continue;
        }
        if(cow[i].state==1){
            milk+=cow[i].m*0.5;
            continue;
        }
        if(cow[i].state==2){
            milk+=cow[i].m;
            continue;
        }
    }
}
float Farm::getMilkProduction(){
    return milk;
}<endfile>Farm.h:<start_file>#pragma once
#include"Cow.h"
class Farm{
    int n;
    int num;
    Cow* cow;
    public:
    float milk;
    Farm(int a);
    void addCow(Cow a);
    void supply(std::string a,int b);
    void startMeal();
    void produceMilk();
    float getMilkProduction();
    ~Farm(){
        delete[] cow;
    }
};<endfile>main.cpp:<start_file>#include <iostream>
#include <string>
#include "Cow.h"
#include "Farm.h"
using namespace std;

int main(){
    int n;
    cin >> n;
    Farm farm(n);
    string name;
    int l, u, m;
    for(int i = 0; i < n; ++i){
        cin >> name >> l >> u >> m;
        Cow cow(name, l, u, m);
        farm.addCow(cow);
    }

    int k;
    cin >> k;
    int t;
    int a;
    for(int i = 0; i < k; ++i){
        cin >> t;
        for(int j = 0; j < t; ++j){
            cin >> name >> a;
            farm.supply(name, a);
        }
        farm.startMeal();
        farm.produceMilk();
    }
    printf("%.1f", farm.getMilkProduction());
    return 0;
}<endfile>makefile:<start_file>main:main-3.o Farm.o Cow.o
	g++ main-3.o Farm.o Cow.o -o main

main-3.o:main-3.cpp Farm.h Cow.h
	g++ -c main-3.cpp -o main-3.o

Farm.o:Farm.cpp Farm.h Cow.h
	g++ -c Farm.cpp  -o Farm.o

Cow.o:Cow.cpp Cow.h
	g++ -c Cow.cpp  -o Cow.o

clean:
	rm *.o main<endfile>, and the input file is:./input/11.txt:<start_file>3
a 2 5 6
b 3 4 7
c 1 6 5
2
1 a 3
2 b 2 c 4<enfile>
\end{lstlisting}
\end{tcolorbox}

\begin{tcolorbox}[left=0mm,right=0mm,top=0mm,bottom=0mm,boxsep=1mm,arc=0mm,boxrule=0pt, frame empty, breakable]
    \small
    \begin{lstlisting}
Given the following code, what is the execution result? The file is under `/app/` directory, and is run with "python3 /app/test.py" if it is a python file, "g++ -std=c++11 /app/test.cpp -o /app/test
/app/test" if it is a cpp file, and "javac /app/\{class_name\}.java
java -cp /app \{class_name\}" if it is a java file.
You should think step by step.  Your answer should be in the following format:
Thought: <your thought>
Output:
<execution result>
\end{lstlisting}
\end{tcolorbox}

\begin{tcolorbox}[left=0mm,right=0mm,top=0mm,bottom=0mm,boxsep=1mm,arc=0mm,boxrule=0pt, frame empty, breakable]
    \small
    \begin{lstlisting}
Here is the code repository:car.cpp:<start_file>#include "car.h"
#include <iostream>
using namespace std;

Car::Car(int num,string eng):Vehicle(num,eng){}

void Car::describe(){
    cout<<"Finish building a car with "<<wheel.get_num()<<" wheels and a "<<engine.get_name()<<" engine."<<endl;
    cout<<"A car with "<<wheel.get_num()<<" wheels and a "<<engine.get_name()<<" engine."<<endl;
}

<endfile>car.h:<start_file>#pragma once
#include "vehicle.h"
using namespace std;

class Car: public Vehicle{
    public:
    Car(int num, string eng);
    void describe();
};<endfile>engine.cpp:<start_file>#include "engine.h"

Engine::Engine(string nam): name(nam) {
	cout << "Using "  << nam << " engine."<< endl;
}

string Engine::get_name() {
	return name;
}
<endfile>engine.h:<start_file>#pragma once
#include <iostream>
#include <string>
using namespace std;

class Engine {
	string name;
public:
	Engine(string);
	string get_name();
};<endfile>main.cpp:<start_file>
#include <iostream>
#include <string>
#include "wheel.h"
#include "engine.h"
#include "vehicle.h"
#include "motor.h"
#include "car.h"
using namespace std;

int main() {
	int n, type, num;
	string engine;

	cin >> n; 
	for (int i=0; i<n; i++) {
		cin >> type >> num >> engine;
		switch (type) {
			case 0: {
				Vehicle v = Vehicle(num, engine);
				v.describe();
				break;
			}
			case 1: {
				Motor m = Motor(num, engine);
				m.describe();
				m.sell();
				break;
			}
			case 2: {
				Car c = Car(num, engine);
				c.describe();
				break;
			}
		}
	}
	return 0;
}<endfile>motor.cpp:<start_file>#include "motor.h"
#include <iostream>
using namespace std;
Motor::Motor(int num,string eng):Vehicle(num,eng){}

void Motor::describe(){
    cout<<"Finish building a motor with "<<wheel.get_num()<<" wheels and a "<<engine.get_name()<<" engine."<<endl;
    cout<<"A motor with "<<wheel.get_num()<<" wheels and a "<<engine.get_name()<<" engine."<<endl;
}

void Motor::sell(){
    cout<<"A motor is sold!"<<endl;
}<endfile>motor.h:<start_file>#pragma once
#include "vehicle.h"
using namespace std;

class Motor: public Vehicle{
    public:
    Motor(int num, string eng);
    void describe();
    void sell();
};<endfile>vehicle.cpp:<start_file>#include "vehicle.h"
#include <iostream>
using namespace std;

Vehicle::Vehicle(int num,string eng):engine(eng),wheel(num){}

void Vehicle::describe(){
    cout<<"Finish building a vehicle with "<<wheel.get_num()<<" wheels and a "<<engine.get_name()<<" engine."<<endl;
    cout<<"A vehicle with "<<wheel.get_num()<<" wheels and a "<<engine.get_name()<<" engine."<<endl;
}<endfile>vehicle.h:<start_file>#pragma once
#include "wheel.h"
#include "engine.h"

using namespace std;

class Vehicle{
    public:
    Engine engine;
    Wheel wheel;
    Vehicle(int num, string eng);
    void describe();

};<endfile>wheel.cpp:<start_file>#include "wheel.h"

Wheel::Wheel(int num): number(num) {
	cout << "Building " << number << " wheels." << endl;
}

int Wheel::get_num() {
	return number;
}<endfile>wheel.h:<start_file>#pragma once
#include <iostream>
using namespace std;

class Wheel {
	int number;
public:
	Wheel(int);
	int get_num();
};<endfile>, and the input file is:./input/6.txt:<start_file>4
0 3 Gasoline
2 4 Hybrid
1 2 Electric
0 6 Magic<enfile>
\end{lstlisting}
\end{tcolorbox}

\begin{tcolorbox}[left=0mm,right=0mm,top=0mm,bottom=0mm,boxsep=1mm,arc=0mm,boxrule=0pt, frame empty, breakable]
    \small
    \begin{lstlisting}
Here is the code repository:24_game.py:<start_file>import itertools
import time
import math

# Operators
OP_CONST = 0  # Constant
OP_ADD = 1  # Addition
OP_SUB = 2  # Subtraction
OP_MUL = 3  # Multiplication
OP_DIV = 4  # Divition
OP_POW = 5  # Exponentiation

OP_SQRT = 6  # Squreroot
OP_FACT = 7  # Factorial
OP_LOG = 8  # Logarithm
OP_C = 9  # Combinations
OP_P = 10  # Permutations

# List of basic operators
operators = [OP_ADD,
             OP_SUB,
             OP_MUL,
             OP_DIV]

# List of advanced operators
advanced_operators = [OP_POW,
                      OP_LOG,
                      OP_C,
                      OP_P]

# List of unary operators
_unary_operators = [OP_SQRT,
                    OP_FACT]

# List of enabled unary operators
unary_operators = []

# Symbol of operators
symbol_of_operator = {OP_ADD: "%s+%s",
                      OP_SUB: "%s-%s",
                      OP_MUL: "%s*%s",
                      OP_DIV: "%s/%s",
                      OP_POW: "%s^%s",
                      OP_SQRT: "sqrt(%s)",
                      OP_FACT: "%s!",
                      OP_LOG: "log_%s(%s)",
                      OP_C: "C(%s, %s)",
                      OP_P: "P(%s, %s)"}

# Priority of operators
priority_of_operator = {OP_ADD: 0,
                        OP_SUB: 0,
                        OP_MUL: 1,
                        OP_DIV: 1,
                        OP_POW: 2,
                        OP_LOG: 3,
                        OP_C: 3,
                        OP_P: 3,
                        OP_SQRT: 3,
                        OP_FACT: 4,
                        OP_CONST: 5}

# Whether operator is commutative
is_operator_commutative = {OP_ADD: True,
                           OP_SUB: False,
                           OP_MUL: True,
                           OP_DIV: False,
                           OP_POW: False,
                           OP_LOG: False,
                           OP_C: False,
                           OP_P: False}

# Whether inside bracket is needed when rendering
need_brackets = {OP_ADD: True,
                 OP_SUB: True,
                 OP_MUL: True,
                 OP_DIV: True,
                 OP_POW: True,
                 OP_FACT: True,
                 OP_SQRT: False,
                 OP_LOG: False,
                 OP_C: False,
                 OP_P: False}


def permutation(n, k):
    return math.factorial(n)/math.factorial(k)


def combination(n, k):
    return permutation(n, k)/math.factorial(n-k)


def evaluate_operation(op, a, b=None):
    """
    Evaluate an operation on a and b.
    """
    if op == OP_ADD: return a + b
    if op == OP_SUB: return a - b
    if op == OP_MUL: return a * b

    try:
        if op == OP_POW and abs(a) < 20 and abs(b) < 20:
            return a ** b

        if op == OP_FACT and a < 10:
            return math.factorial(a)

        if op == OP_C and 0 < b <= a <= 13:
            return combination(a, b)

        if op == OP_P and 0 < b <= a <= 13:
            return permutation(a, b)

        if op == OP_SQRT and a < 1000000:
            return math.sqrt(a)

        if op == OP_DIV: return a / b
        if op == OP_LOG: return math.log(b, a)
    except (ZeroDivisionError, ValueError, TypeError):
        pass
    except OverflowError:
        print(a, b)

    return float("NaN")


def fit_to_int(x, eps=1e-9):
    """
    Convert x to int if x is close to an integer.
    """
    try:
        if abs(round(x) - x) <= eps:
            return round(x)
        else:
            return x
    except ValueError:
        return float("NaN")
    except TypeError:
        return float("NaN")


class Node:
    def __init__(self, value=None, left=None, right=None, op=OP_CONST):
        if op not in unary_operators \
                and op != OP_CONST and is_operator_commutative[op] \
                and str(left) > str(right):
            left, right = right, left

        self._value = value
        self._str_cache = None
        self.left = left
        self.right = right
        self.op = op

    @property
    def value(self):
        if self._value is None:
            assert self.op != OP_CONST

            if self.op in unary_operators:
                self._value = evaluate_operation(self.op, self.left.value)
            else:
                self._value = evaluate_operation(self.op, self.left.value, self.right.value)

            self._value = fit_to_int(self._value)
        return self._value

    def __str__(self):
        if self._str_cache is None:
            self._str_cache = self._str()
        return self._str_cache

    def _str(self):
        # Constant
        if self.op == OP_CONST:
            return str(self._value)

        # Unary operator
        elif self.op in unary_operators:
            str_left = str(self.left)

            if need_brackets[self.op] \
                    and priority_of_operator[self.left.op] < priority_of_operator[self.op]:
                str_left = "(" + str_left + ")"

            return symbol_of_operator[self.op] % str_left

        # Other operator
        else:
            str_left = str(self.left)
            str_right = str(self.right)

            # Add brackets inside
            if need_brackets[self.op] \
                    and priority_of_operator[self.left.op] < priority_of_operator[self.op]:
                str_left = "(" + str_left + ")"

            if need_brackets[self.op] \
                    and (priority_of_operator[self.right.op] < priority_of_operator[self.op]
                         or (priority_of_operator[self.right.op] == priority_of_operator[self.op]
                             and not is_operator_commutative[self.op])):
                str_right = "(" + str_right + ")"

            # Render
            return symbol_of_operator[self.op] % (str_left, str_right)


def enumerate_nodes(node_list, callback, max_depth):
    # Found an expression
    if len(node_list) == 1:
        callback(node_list[0])

    # Constrain maximum depth
    if max_depth == 0:
        return

    # Non-unary operators
    for left, right in itertools.permutations(node_list, 2):
        new_node_list = node_list.copy()
        new_node_list.remove(left)
        new_node_list.remove(right)

        for op in operators:
            enumerate_nodes(new_node_list + [Node(left=left, right=right, op=op)], callback, max_depth-1)

            if not is_operator_commutative[op] and str(left) != str(right):
                enumerate_nodes(new_node_list + [Node(left=right, right=left, op=op)], callback, max_depth-1)

    # Unary operators
    for number in node_list:
        new_node_list = node_list.copy()
        new_node_list.remove(number)

        for op in unary_operators:
            new_node = Node(left=number, op=op)
            if new_node.value == number.value:
                continue

            enumerate_nodes(new_node_list + [new_node], callback, max_depth-1)


class CallbackFindTarget:
    def __init__(self, target):
        self.target = target
        self.results = []
        self.duplication_count = 0
        self.enumeration_count = 0

    def __call__(self, node):
        if node.value == self.target and str(node) not in self.results:
            print(self.target, "=", node)
            self.results.append(str(node))
        elif node.value == self.target:
            self.duplication_count += 1

        self.enumeration_count += 1

    def show(self, execution_time):
        print()
        print("%d solution(s) in %.3f seconds" % (len(self.results), execution_time))
        print("%d duplication(s)" % self.duplication_count)
        print("%d combination(s)" % self.enumeration_count)


class CallbackAllTarget:
    def __init__(self):
        self.results = {}
        self.enumeration_count = 0

    def __call__(self, node):
        try:
            int(node.value)
        except ValueError:
            return

        if node.value not in self.results \
                and int(node.value) == node.value:
            self.results[node.value] = node

        self.enumeration_count += 1

    def __str__(self):
        string = ""
        for value in sorted(self.results.keys()):
            string += "%d = %s" % (value, str(self.results[value]))
            string += "\n"
        return string

    def show(self, execution_time):
        print(self)
        print()
        print("%d targets(s) in %.3f seconds" % (len(self.results), execution_time))
        print("%d combination(s)" % self.enumeration_count)


def select_yes_no(prompt, default=False):
    answer = input(prompt).strip().lower()
    if answer == "y":
        return True
    if answer == "n":
        return False
    return default


def select_int(prompt, default):
    try:
        return int(input(prompt).strip())
    except ValueError:
        return default


def main():
    global operators
    global unary_operators


    unary_operators_allowed = False
    enumerate_all = False


    if not enumerate_all:
        target = 24
        callback = CallbackFindTarget(target=target)

    if enumerate_all:
        callback = CallbackAllTarget()
    else:
        callback = CallbackFindTarget(target=target)

    with open('input.txt', 'r') as file:
        inputs = [int(i) for line in file for i in line.split() if i != ""]
    node_list = [Node(value=i) for i in inputs]

    enumerate_nodes(node_list, callback, max_depth=len(node_list)-1+unary_operators_allowed)

main()<enfile>, and the input file is: input.txt:<start_file>4 4 7 7<end_file>
\end{lstlisting}
\end{tcolorbox}
\subsection{SC}

\subsubsection{Tasks Descriptions}
\label{sec:appendix1}

The scientific computing component of \bench consists of 4 carefully curated areas, aiming to evaluate model performance on computational tasks that exhibit a time-consuming nature as well as applicational values in scientific computing areas. In this section, we provide a detailed description of each component.

\paragraph{Numerical Optimization.} In this task, the model is given a program that solves an optimization problem through gradient descent. The query may be the optimized value (\textit{min}) or the optimal point (\textit{argmin}). We carefully select four functions, which consist of: a simple quadratic function, Rosenbrock Function, Himmelblau’s Function, and a polynomial function with linear constraints. For each function, we will select multiple different hyperparameter configurations to assess the model's performance. These four functions provide a systematic evaluation of the model's potential to serve as a surrogate model in this field. As the quadratic function is solvable without need the to run the gradient descent, the model may solve it through world knowledge. The Rosenbrock function is known for its narrow, curved valley containing the global minimum, making it difficult for optimization algorithms to converge. Therefore the output is highly dependent on hyperparameters (initial point, learning rate, maximum steps), thus the model must execute code in its reasoning process to acquire the answer. Himmelblau's function has multiple local minima, also posing sensitivity to hyperparameters.

\paragraph{PDE Solving.} We consider three types of Partial Differential Equations: the 1D Heat Equation, the 2D Wave Equation, and the 2D Laplace Equation. For the 1D Heat Equation, we focus on solving the following equation:
\begin{equation}
    \frac{\partial u}{\partial t} = \alpha \frac{\partial^2 u}{\partial x^2}.
\end{equation}
For the 2D Laplace Equation, we aim to solve the equation:
\begin{equation}
    \frac{\partial^2 u}{\partial x^2} + \frac{\partial^2 u}{\partial y^2} = 0.
\end{equation}
Lastly, for the 2D Wave Equation, we work on solving the following equation:
\begin{equation}
    \frac{\partial^2 u}{\partial t^2} = c^2 \left( \frac{\partial^2 u}{\partial x^2} + \frac{\partial^2 u}{\partial y^2} \right).
\end{equation}
We solve 1D Heat Equation and 2D Wave Equation using the Explicit Finite Difference Method. For the 2D Laplace Equation, we solve it using the Gauss-Seidel Method. The model is then queried on the values of $u$ and $x$.
\paragraph{Fourier Transform (FFT)} We implement FFT using the Cooley-Tukey Algorithm and query the model to give the magnitude of the top 10 values.
\paragraph{ODE Solving} For solving ordinary differential equations, we constructed three different equations and implemented the Euler Method and the Runge-Kutta Method so solve these equations.

\subsubsection{Evaluation Metrics}
\label{app:metric}
\paragraph{Relative Absolute Error (RAE).} 
Given a scalar ground truth value \( p \) and a model prediction \( \hat{p} \), the Relative Absolute Error (RAE) is defined as:
\begin{equation}
    \text{RAE}(\hat{p}, p) = \frac{|p - \hat{p}|}{|p|}.
\end{equation}
For cases involving multiple entries, such as tensors or vectors, the following alignment procedure is applied: (1) if the prediction contains fewer elements than the ground truth, the prediction is padded with zeros until it matches the length of the ground truth; (2) if the prediction has more elements than the ground truth, it is truncated to match the ground truth length. The average RAE is then computed by averaging the RAE for each corresponding element.

\paragraph{Exact Matching.} 
For tasks involving position-based predictions, such as binary search, we adapt exact matching, as the accuracy of the algorithm is determined by comparing the exactness of the estimated result to the true result. This evaluation method checks if the estimated solution matches the ground truth exactly, typically using string or sequence matching. For such tasks, an exact match is considered a success, and any discrepancy between the ground truth and the estimate results in failure. Formally, given a string $s$ and the model's prediction $\hat{s}$, the Exact Matching is given by:
\begin{equation}
    \text{EM}(s,\hat{s})=\mathbbm{1}[s=\hat{s}]
\end{equation}
where $\mathbbm{1}[\cdot]$ is the indicator function.

\subsubsection{System Prompts}

\paragraph{Zero-shot Chain-of-Thought:}

\begin{tcolorbox}[left=0mm,right=0mm,top=0mm,bottom=0mm,boxsep=1mm,arc=0mm,boxrule=0pt, frame empty, breakable]
    \small
    \begin{lstlisting}
You are an expert in gradient_descent programming.
Please execute the above code with the input provided and return the output. You should think step by step.
Your answer should be in the following format:
Thought: <your thought>
Output: <execution result>
Please follow this format strictly and ensure the Output section contains only the required result without any additional text.
\end{lstlisting}
\end{tcolorbox}

\paragraph{Zero-shot:}

\begin{tcolorbox}[left=0mm,right=0mm,top=0mm,bottom=0mm,boxsep=1mm,arc=0mm,boxrule=0pt, frame empty, breakable]
    \small
    \begin{lstlisting}
You are an expert in gradient_descent programming.
Please execute the given code with the provided input and return the output.
Make sure to return only the output in the exact format as expected.

Output Format:
Output: <result>
\end{lstlisting}
\end{tcolorbox}

\paragraph{Few-shot Chain-of-Thought:}

\begin{tcolorbox}[left=0mm,right=0mm,top=0mm,bottom=0mm,boxsep=1mm,arc=0mm,boxrule=0pt, frame empty, breakable]
    \small
    \begin{lstlisting}
You are an expert in gradient_descent programming.
Please execute the above code with the input provided and return the output. You should think step by step.
Your answer should be in the following format:
Thought: <your thought>
Output: <execution result>
Please follow this format strictly and ensure the Output section contains only the required result without any additional text.

Here are some examples:
{{examples here}}

\end{lstlisting}
\end{tcolorbox}

\subsubsection{Demo Questions}

\begin{tcolorbox}[left=0mm,right=0mm,top=0mm,bottom=0mm,boxsep=1mm,arc=0mm,boxrule=0pt, frame empty, breakable]
    \small
    \begin{lstlisting}
code:```
import numpy as np
import argparse

def f(t, y):
    """dy/dt = -y"""
    return -y

def euler_method(f, y0, t0, t_end, h, additional_args=None):
    t_values = np.arange(t0, t_end, h)
    y_values = [y0]
    v_values = [additional_args] if additional_args is not None else [None]
    
    for t in t_values[:-1]:
        if additional_args:
            y_next, v_next = y_values[-1] + h * f(t, y_values[-1])[0], v_values[-1] + h * f(t, y_values[-1], v_values[-1])[1]
            y_values.append(y_next)
            v_values.append(v_next)
        else:
            y_next = y_values[-1] + h * f(t, y_values[-1])
            y_values.append(y_next)
    
    return t_values, np.array(y_values), np.array(v_values) if v_values[0] is not None else None

def main():
    parser = argparse.ArgumentParser()
    parser.add_argument("--y0", type=float, default=1.0)
    parser.add_argument("--t0", type=float, default=0.0)
    parser.add_argument("--t_end", type=float, default=10.0)
    parser.add_argument("--h", type=float, default=0.1)
    args = parser.parse_args()

    y0_1 = args.y0
    t0 = args.t0
    t_end = args.t_end
    h = args.h

    t_values, y_values, _ = euler_method(f, y0_1, t0, t_end, h)
    print(f"{y_values[-1]:.4f}")

if __name__ == "__main__":
    main()

```
command:```
python euler_3.py --y0 12 --t0 0.0 --t_end 74 --h 0.36
```
\end{lstlisting}
\end{tcolorbox}

\begin{tcolorbox}[left=0mm,right=0mm,top=0mm,bottom=0mm,boxsep=1mm,arc=0mm,boxrule=0pt, frame empty, breakable]
    \small
    \begin{lstlisting}
code:```
import numpy as np
import argparse

def gradient_descent(func, grad_func, initial_guess, learning_rate=0.1, tolerance=1e-6, max_iter=1000):
    x = initial_guess
    for _ in range(max_iter):
        grad = grad_func(x)
        x = x - learning_rate * grad
        if np.abs(grad) < tolerance:
            break
    return x, func(x)

# Function and its gradient
def func(x):
    return (x - 3)**2 + 5

def grad_func(x):
    return 2 * (x - 3)

def main(): 
    parser = argparse.ArgumentParser()
    parser.add_argument("--initial_guess", type=float, default=0.0)
    parser.add_argument("--learning_rate", type=float, default=0.1)
    parser.add_argument("--tolerance", type=float, default=1e-6)
    parser.add_argument("--max_iter", type=int, default=1000)
    args = parser.parse_args()

    # Test with initial guess
    initial_guess = args.initial_guess
    optimal_x, optimal_value = gradient_descent(func, grad_func, initial_guess)
    # optimal x
    print(f"{optimal_x:.3f}")

if __name__ == "__main__":
    main()
```
command:```
python gd_ox.py --initial_guess -5.0 --learning_rate 0.01 --max_iter 5000
```
\end{lstlisting}
\end{tcolorbox}

\begin{tcolorbox}[left=0mm,right=0mm,top=0mm,bottom=0mm,boxsep=1mm,arc=0mm,boxrule=0pt, frame empty, breakable]
    \small
    \begin{lstlisting}
code:```
import numpy as np
import argparse

def solve_heat_eq(L, T, alpha, Nx, Nt):
    # L: length of the rod
    # T: total time
    # alpha: thermal diffusivity
    # Nx: number of spatial steps
    # Nt: number of time steps

    dx = L / (Nx - 1)
    dt = T / Nt
    r = alpha * dt / dx**2

    # Initial condition: u(x, 0) = sin(pi * x)
    x = np.linspace(0, L, Nx)
    u = np.sin(np.pi * x)

    # Time stepping
    for n in range(Nt):
        u_new = u.copy()
        for i in range(1, Nx - 1):
            u_new[i] = u[i] + r * (u[i-1] - 2*u[i] + u[i+1])
        u = u_new
    return x, u

def parse_input():
    parser = argparse.ArgumentParser(description="Solve the 1D Heat Equation")
    parser.add_argument('--L', type=float, required=True, help="Length of the rod")
    parser.add_argument('--T', type=float, required=True, help="Total time")
    parser.add_argument('--alpha', type=float, required=True, help="Thermal diffusivity")
    parser.add_argument('--Nx', type=int, required=True, help="Number of spatial points")
    parser.add_argument('--Nt', type=int, required=True, help="Number of time steps")
    return parser.parse_args()

def main():
    args = parse_input()
    x, u = solve_heat_eq(args.L, args.T, args.alpha, args.Nx, args.Nt)
    np.set_printoptions(threshold=np.inf, linewidth=np.inf)
    formatted_x = np.vectorize(lambda x: f"{x:.4e}")(x)
    print(f"{formatted_x}")

if __name__ == "__main__":
    main()

```
command:```
python heat_eq_x.py --L 36 --T 62 --alpha 91 --Nx 170 --Nt 860
```
\end{lstlisting}
\end{tcolorbox}

\begin{tcolorbox}[left=0mm,right=0mm,top=0mm,bottom=0mm,boxsep=1mm,arc=0mm,boxrule=0pt, frame empty, breakable]
    \small
    \begin{lstlisting}
code:```
import numpy as np
import argparse

def gradient_descent(func, grad_func, initial_guess, learning_rate=0.1, tolerance=1e-6, max_iter=1000):
    x = initial_guess
    for _ in range(max_iter):
        grad = grad_func(x)
        x = x - learning_rate * grad
        if np.abs(grad) < tolerance:
            break
    return x, func(x)

# Function and its gradient
def func(x):
    return (x - 3)**2 + 5

def grad_func(x):
    return 2 * (x - 3)

def main(): 
    parser = argparse.ArgumentParser()
    parser.add_argument("--initial_guess", type=float, default=0.0)
    parser.add_argument("--learning_rate", type=float, default=0.1)
    parser.add_argument("--tolerance", type=float, default=1e-6)
    parser.add_argument("--max_iter", type=int, default=1000)
    args = parser.parse_args()

    # Test with initial guess
    initial_guess = args.initial_guess
    optimal_x, optimal_value = gradient_descent(func, grad_func, initial_guess)
    # optimal x
    print(f"{optimal_x:.3f}")

if __name__ == "__main__":
    main()
```
command:```
python gd_ox.py --initial_guess -10.0 --learning_rate 0.001 --max_iter 100
```
\end{lstlisting}
\end{tcolorbox}

\begin{tcolorbox}[left=0mm,right=0mm,top=0mm,bottom=0mm,boxsep=1mm,arc=0mm,boxrule=0pt, frame empty, breakable]
    \small
    \begin{lstlisting}
code:```
import numpy as np
import argparse

# Objective function: f(x, y) = x^2 + y^2
def objective(x, y):
    return x**2 + y**2

# Gradient of the objective function: ∇f(x, y) = (2x, 2y)
def gradient(x, y):
    return np.array([2 * x, 2 * y])

# Projection function onto the constraint x + y = 1
def projection(x, y):
    # Since the constraint is x + y = 1, we can project the point (x, y) onto the line
    # by solving the system: x' + y' = 1
    # Let x' = x - (x + y - 1)/2, and y' = y - (x + y - 1)/2
    adjustment = (x + y - 1) / 2
    return np.array([x - adjustment, y - adjustment])

def projected_gradient_descent(learning_rate=0.1, max_iter=1000, tolerance=1e-6, initial_guess=(0.0, 0.0)):
    x, y = initial_guess
    
    for _ in range(max_iter):
        # Compute the gradient of the objective function
        grad = gradient(x, y)
        
        # Update the variables by moving in the opposite direction of the gradient
        x, y = np.array([x, y]) - learning_rate * grad
        
        # Project the updated point onto the constraint set (x + y = 1)
        x, y = projection(x, y)
        
        # Check if the gradient is small enough to stop
        if np.linalg.norm(grad) < tolerance:
            break
    
    return x, y, objective(x, y)

def main(): 
    parser = argparse.ArgumentParser()
    parser.add_argument("--initial_guess_x", type=float, default=0.0)
    parser.add_argument("--initial_guess_y", type=float, default=0.0)
    parser.add_argument("--learning_rate", type=float, default=0.1)
    parser.add_argument("--tolerance", type=float, default=1e-6)
    parser.add_argument("--max_iter", type=int, default=1000)
    args = parser.parse_args()
    
    initial_guess = (args.initial_guess_x, args.initial_guess_y)
    optimal_x, optimal_y, optimal_value = projected_gradient_descent(args.learning_rate, args.max_iter, args.tolerance, initial_guess)
    print(f"{optimal_x:.4e}, {optimal_y:.4e}")

if __name__ == "__main__":
    main()

```
command:```
python gd_pgdx.py --initial_guess_x 32.14 --initial_guess_y 46.04 --learning_rate 0.01 --max_iter 1000
```
\end{lstlisting}
\end{tcolorbox}
\subsection{TC}

\subsubsection{Tasks Descriptions of Time Consuming~(TC)}
\label{sec:appendix2}

The time consuming component of \bench is comprised of 4 tasks in for computationally expensive areas, covering a spectrum of Linear Algebra, Sorting, Searching, Monte Carlo Simulations and String Matching Programs. Some of these tasks take hours to complete, showing their potential to benchmark LLM's ability to reason through lengthy computations.

\paragraph{Linear Algebra.} In this task, we are focused on acquiring key properties in linear algebra given square matrices of varying sizes. In particular, we query the model on solving LU decomposition, QR decomposition, the largest eigenvalue and eigenvector using the power method, and the inversion matrix.

\paragraph{Sorting And Searching.} We include four classical algorithmic problems in this area, namely Hamiltonian Cycle, Traveling Salesman Problem (TSP), Sorting an array of real numbers and Searching. For Hamiltonian Cycle, we adopt the backtracking algorithm. Specifically, we randomly generate graphs with vertices from 4 to 100 and ask the model to find whether a Hamiltonian cycle exists. For TSP, we implement a naive brute-force algorithm and ask the model to find the length of the optimal path. For Sorting, we adopt the bubble sort, quick sort, and merge sort algorithms. For each algorithm, we consider different list sizes from 5 to 100 and generate 10 test cases for each list size. The evaluation metric is the rank correlation (also Spearman's $\rho$ ). Lastly, for searching, we adopt binary search and query the model on randomly generated lists of varying sizes.
\paragraph{Monte Carlo Estimation.} We adopt Monte Carlo simulation to estimate the values of specific real numbers (e.g. $\pi, e$), as well as a future stock price prediction that follows the Brownian motion. We alter the number of samples used in Monte Carlo estimation, resulting in varying program outcomes.
\paragraph{String Matching Program.} We adopt the naive string matching, KMP, and Rabin-Karp algorithms. For each algorithm, we randomly generate text and pattern with varying lengths, and query the model on the existence and position of the matching.

\subsubsection{Evaluation Metrics}
\label{app:metric2}

\paragraph{Rank Correlation.} 
Rank Correlation~\citep{spearman1904proof}, also referred to as Spearman's $\rho$, is used to assess sorting tasks by measuring the correlation between the estimated ordinal ranking and the ground truth, which can be written as:
\begin{equation}
\text{RankCorr} = \frac{\text{Cov}(x_{1:N}, y_{1:N})}{\sigma(x_{1:N}) \sigma(y_{1:N})}
\end{equation}

where \( x_{1:N} \) and \( y_{1:N} \) denote the true and estimated rankings, respectively, and \( \text{Cov} \) and \( \sigma \) represent the covariance and standard deviation of the respective sequences.

\subsubsection{System Prompts}

\paragraph{Zero-shot Chain-of-Thought:}

\begin{tcolorbox}[left=0mm,right=0mm,top=0mm,bottom=0mm,boxsep=1mm,arc=0mm,boxrule=0pt, frame empty, breakable]
    \small
    \begin{lstlisting}
You are an expert in string_matching programming.
Please execute the above code with the input provided and return the output. You should think step by step.
Your answer should be in the following format:
Thought: <your thought>
Output: <execution result>
Please follow this format strictly and ensure the Output section contains only the required result without any additional text.
\end{lstlisting}
\end{tcolorbox}

\paragraph{Zero-shot:}

\begin{tcolorbox}[left=0mm,right=0mm,top=0mm,bottom=0mm,boxsep=1mm,arc=0mm,boxrule=0pt, frame empty, breakable]
    \small
    \begin{lstlisting}
You are an expert in string_matching programming.
Please execute the given code with the provided input and return the output.
Make sure to return only the output in the exact format as expected.

Output Format:
Output: <result>
\end{lstlisting}
\end{tcolorbox}

\paragraph{Few-shot Chain-of-Thought:}

\begin{tcolorbox}[left=0mm,right=0mm,top=0mm,bottom=0mm,boxsep=1mm,arc=0mm,boxrule=0pt, frame empty, breakable]
    \small
    \begin{lstlisting}
You are an expert in string_matching programming.
Please execute the above code with the input provided and return the output. You should think step by step.
Your answer should be in the following format:
Thought: <your thought>
Output: <execution result>
Please follow this format strictly and ensure the Output section contains only the required result without any additional text.

Here are some examples:
{{examples here}}

\end{lstlisting}
\end{tcolorbox}

\subsubsection{Demo Questions}

\begin{tcolorbox}[left=0mm,right=0mm,top=0mm,bottom=0mm,boxsep=1mm,arc=0mm,boxrule=0pt, frame empty, breakable]
    \small
    \begin{lstlisting}
code:```
import itertools
import math
import sys
import argparse
def euclidean_distance(p1, p2):
    """Calculate the Euclidean distance between two points"""
    return math.sqrt((p1[0] - p2[0])**2 + (p1[1] - p2[1])**2)

def tsp_bruteforce(positions):
    """Brute-force TSP solver"""
    n = len(positions)
    min_path = None
    min_distance = float('inf')

    # Generate all possible permutations of the cities (excluding the starting point)
    for perm in itertools.permutations(range(1, n)):
        path = [0] + list(perm)  # Start at city 0
        distance = 0
        # Calculate the total distance of the current permutation
        for i in range(1, len(path)):
            distance += euclidean_distance(positions[path[i-1]], positions[path[i]])

        # Compare the distance with the minimum distance found so far
        if distance < min_distance:
            min_distance = distance
            min_path = path

    return min_path, min_distance

def parse_positions(positions_str):
    """Convert the string input back to a list of tuples"""
    positions = []
    for pos in positions_str.split():
        x, y = map(float, pos.split(','))
        positions.append((x, y))
    return positions

def main():
    parser = argparse.ArgumentParser()
    parser.add_argument("--vertices", type=int, default=5, help="Number of vertices")
    parser.add_argument("--positions", type=str, default="0,0 1,1 2,2 3,3 4,4", help="List of positions in the format 'x,y'")
    args = parser.parse_args()

    vertices = args.vertices
    positions_str = args.positions
    
    # Parse positions
    positions = parse_positions(positions_str)

    # Solve TSP using brute force
    path, distance = tsp_bruteforce(positions)

    print(f"{distance:.2f}")

if __name__ == "__main__":
    main()

```
command:```
python tsp.py --vertices 3 --positions "8.51,4.18 8.1,7.92 1.57,0.49" 
```
\end{lstlisting}
\end{tcolorbox}

\begin{tcolorbox}[left=0mm,right=0mm,top=0mm,bottom=0mm,boxsep=1mm,arc=0mm,boxrule=0pt, frame empty, breakable]
    \small
    \begin{lstlisting}
code:```
import itertools
import math
import sys
import argparse
def euclidean_distance(p1, p2):
    """Calculate the Euclidean distance between two points"""
    return math.sqrt((p1[0] - p2[0])**2 + (p1[1] - p2[1])**2)

def tsp_bruteforce(positions):
    """Brute-force TSP solver"""
    n = len(positions)
    min_path = None
    min_distance = float('inf')

    # Generate all possible permutations of the cities (excluding the starting point)
    for perm in itertools.permutations(range(1, n)):
        path = [0] + list(perm)  # Start at city 0
        distance = 0
        # Calculate the total distance of the current permutation
        for i in range(1, len(path)):
            distance += euclidean_distance(positions[path[i-1]], positions[path[i]])

        # Compare the distance with the minimum distance found so far
        if distance < min_distance:
            min_distance = distance
            min_path = path

    return min_path, min_distance

def parse_positions(positions_str):
    """Convert the string input back to a list of tuples"""
    positions = []
    for pos in positions_str.split():
        x, y = map(float, pos.split(','))
        positions.append((x, y))
    return positions

def main():
    parser = argparse.ArgumentParser()
    parser.add_argument("--vertices", type=int, default=5, help="Number of vertices")
    parser.add_argument("--positions", type=str, default="0,0 1,1 2,2 3,3 4,4", help="List of positions in the format 'x,y'")
    args = parser.parse_args()

    vertices = args.vertices
    positions_str = args.positions
    
    # Parse positions
    positions = parse_positions(positions_str)

    # Solve TSP using brute force
    path, distance = tsp_bruteforce(positions)

    print(f"{distance:.2f}")

if __name__ == "__main__":
    main()

```
command:```
python tsp.py --vertices 3 --positions "0.9,2.44 4.67,0.82 3.8,5.73" 
```
\end{lstlisting}
\end{tcolorbox}

\begin{tcolorbox}[left=0mm,right=0mm,top=0mm,bottom=0mm,boxsep=1mm,arc=0mm,boxrule=0pt, frame empty, breakable]
    \small
    \begin{lstlisting}
code:```
import itertools
import math
import sys
import argparse
def euclidean_distance(p1, p2):
    """Calculate the Euclidean distance between two points"""
    return math.sqrt((p1[0] - p2[0])**2 + (p1[1] - p2[1])**2)

def tsp_bruteforce(positions):
    """Brute-force TSP solver"""
    n = len(positions)
    min_path = None
    min_distance = float('inf')

    # Generate all possible permutations of the cities (excluding the starting point)
    for perm in itertools.permutations(range(1, n)):
        path = [0] + list(perm)  # Start at city 0
        distance = 0
        # Calculate the total distance of the current permutation
        for i in range(1, len(path)):
            distance += euclidean_distance(positions[path[i-1]], positions[path[i]])

        # Compare the distance with the minimum distance found so far
        if distance < min_distance:
            min_distance = distance
            min_path = path

    return min_path, min_distance

def parse_positions(positions_str):
    """Convert the string input back to a list of tuples"""
    positions = []
    for pos in positions_str.split():
        x, y = map(float, pos.split(','))
        positions.append((x, y))
    return positions

def main():
    parser = argparse.ArgumentParser()
    parser.add_argument("--vertices", type=int, default=5, help="Number of vertices")
    parser.add_argument("--positions", type=str, default="0,0 1,1 2,2 3,3 4,4", help="List of positions in the format 'x,y'")
    args = parser.parse_args()

    vertices = args.vertices
    positions_str = args.positions
    
    # Parse positions
    positions = parse_positions(positions_str)

    # Solve TSP using brute force
    path, distance = tsp_bruteforce(positions)

    print(f"{distance:.2f}")

if __name__ == "__main__":
    main()

```
command:```
python tsp.py --vertices 3 --positions "7.63,4.72 1.07,1.42 8.36,5.63" 
```
\end{lstlisting}
\end{tcolorbox}

\begin{tcolorbox}[left=0mm,right=0mm,top=0mm,bottom=0mm,boxsep=1mm,arc=0mm,boxrule=0pt, frame empty, breakable]
    \small
    \begin{lstlisting}
code:```
import sys
import argparse

def binary_search(arr, target):
    """Binary Search algorithm"""
    low = 0
    high = len(arr) - 1
    
    while low <= high:
        mid = (low + high) // 2  # Find the middle element
        if arr[mid] == target:
            return mid  # Target found at index mid
        elif arr[mid] < target:
            low = mid + 1  # Target is in the right half
        else:
            high = mid - 1  # Target is in the left half
    
    return -1  # Target not found

def parse_input(input_str):
    """Parse input string into a list of integers"""
    return list(map(int, input_str.split()))

def main():
    parser = argparse.ArgumentParser(description="Binary Search Algorithm")
    parser.add_argument('--list', type=str, required=True, help="Input sorted list of integers")
    parser.add_argument('--target', type=int, required=True, help="Target integer to search")
    args = parser.parse_args()
    
    input_list = parse_input(args.list)
    
    result = binary_search(input_list, args.target)
    
    if result != -1:
        print(f"Target found at index: {result}")
    else:
        print("Target not found")

if __name__ == "__main__":
    main()

```
command:```
python binary_search.py --list "-334 -200 180 936 973" --target -771
```
\end{lstlisting}
\end{tcolorbox}

\begin{tcolorbox}[left=0mm,right=0mm,top=0mm,bottom=0mm,boxsep=1mm,arc=0mm,boxrule=0pt, frame empty, breakable]
    \small
    \begin{lstlisting}
code:```
import itertools
import math
import sys
import argparse
def euclidean_distance(p1, p2):
    """Calculate the Euclidean distance between two points"""
    return math.sqrt((p1[0] - p2[0])**2 + (p1[1] - p2[1])**2)

def tsp_bruteforce(positions):
    """Brute-force TSP solver"""
    n = len(positions)
    min_path = None
    min_distance = float('inf')

    # Generate all possible permutations of the cities (excluding the starting point)
    for perm in itertools.permutations(range(1, n)):
        path = [0] + list(perm)  # Start at city 0
        distance = 0
        # Calculate the total distance of the current permutation
        for i in range(1, len(path)):
            distance += euclidean_distance(positions[path[i-1]], positions[path[i]])

        # Compare the distance with the minimum distance found so far
        if distance < min_distance:
            min_distance = distance
            min_path = path

    return min_path, min_distance

def parse_positions(positions_str):
    """Convert the string input back to a list of tuples"""
    positions = []
    for pos in positions_str.split():
        x, y = map(float, pos.split(','))
        positions.append((x, y))
    return positions

def main():
    parser = argparse.ArgumentParser()
    parser.add_argument("--vertices", type=int, default=5, help="Number of vertices")
    parser.add_argument("--positions", type=str, default="0,0 1,1 2,2 3,3 4,4", help="List of positions in the format 'x,y'")
    args = parser.parse_args()

    vertices = args.vertices
    positions_str = args.positions
    
    # Parse positions
    positions = parse_positions(positions_str)

    # Solve TSP using brute force
    path, distance = tsp_bruteforce(positions)

    print(f"{distance:.2f}")

if __name__ == "__main__":
    main()

```
command:```
python tsp.py --vertices 10 --positions "6.81,5.28 9.95,8.98 0.63,0.11 8.84,0.55 9.03,9.98 6.22,2.7 2.99,9.11 0.54,9.36 3.08,4.15 5.73,1.86" 
```
\end{lstlisting}
\end{tcolorbox}
\subsection{BG}

\begin{table}[!ht]
  \centering

  \begin{tabular}{lccc}
      \toprule
      Java & Python & C++ \\
      \midrule
       51   & 45     & 54  \\
      \bottomrule
  \end{tabular}

  \caption{Language usage count across different categories in the BG subset.}
  \label{tab:bg_language_usage}

\end{table}

\begin{table}[!ht]
  \centering

\resizebox{0.5\textwidth}{!}{\begin{tabular}{lccc}
\toprule
\textbf{Error Type} & Java & Python3 & CPP \\
\midrule
== and = confusion & 5 & 6 & 5 \\
undefined keywords & 6 & 3 & 5 \\
parentheses mismatch & 5 & 5 & 6 \\
indexing error & 10 & 9 & 11 \\
undefined objects & 11 & 9 & 8 \\
unclosed string & 7 & 5 & 7 \\
conditional statement error & 10 & 8 & 9 \\
undefined methods & 8 & 3 & 6 \\
colon missing & 5 & 7 & 8 \\
wrong comment mark & 9 & 1 & 9 \\
variable value error & 2 & 2 & 4 \\
operation error & 2 & 2 & 3 \\
other error & 4 & 2 & 1 \\
statement separation & 4 & 0 & 7 \\
indentation error & 0 & 4 & 0 \\
Double Bugs & 10 & 8 & 10 \\
Triple Bugs & 12 & 10 & 11 \\
Quadruple Bugs & 8 & 5 & 9 \\
\bottomrule
\end{tabular}}

\caption{Details of bug types in BG dataset and how many times each kind of bug appears in different languages.}
\label{tab:stat:6}

\end{table}

In BG, the distribution of language usage across categories is shown in Table \ref{tab:bg_language_usage}, indicating a balanced usage of Java, Python, and C++. Table \ref{tab:stat:6} presents a detailed breakdown of bug types and their frequency across different languages. This distribution allows us to assess the model's ability to handle a variety of bugs across multiple programming languages.

\subsubsection{System Prompts}

\paragraph{Zero-shot Chain-of-Thought:}

\begin{tcolorbox}[left=0mm,right=0mm,top=0mm,bottom=0mm,boxsep=1mm,arc=0mm,boxrule=0pt, frame empty, breakable]
    \small
    \begin{lstlisting}
Given the following code, what is the execution result? The file is under `/app/` directory, and is run with "python3 /app/test.py" if it is a python file, "g++ -std=c++11 /app/test.cpp -o /app/test
/app/test" if it is a cpp file, and "javac /app/\{class_name\}.java
java -cp /app \{class_name\}" if it is a java file.
You should think step by step.  Your answer should be in the following format:
Thought: <your thought>
Output:
<execution result>
\end{lstlisting}
\end{tcolorbox}

\paragraph{Zero-shot:}

\begin{tcolorbox}[left=0mm,right=0mm,top=0mm,bottom=0mm,boxsep=1mm,arc=0mm,boxrule=0pt, frame empty, breakable]
    \small
    \begin{lstlisting}
Given the following code, what is the execution result? The file is under `/app/` directory, and is run with "python3 /app/test.py" if it is a python file, "g++ -std=c++11 /app/test.cpp -o /app/test
/app/test" if it is a cpp file, and "javac /app/\{class_name\}.java
java -cp /app \{class_name\}" if it is a java file.
Your answer should be in the following format:
Output:
<execution result>
\end{lstlisting}
\end{tcolorbox}

\paragraph{Few-shot Chain-of-Thought:}

\begin{tcolorbox}[left=0mm,right=0mm,top=0mm,bottom=0mm,boxsep=1mm,arc=0mm,boxrule=0pt, frame empty, breakable]
    \small
    \begin{lstlisting}
Given the following code, what is the execution result? The file is under `/app/` directory, and is run with "python3 /app/test.py" if it is a python file, "g++ -std=c++11 /app/test.cpp -o /app/test
/app/test" if it is a cpp file, and "javac /app/\{class_name\}.java
java -cp /app \{class_name\}" if it is a java file.
You should think step by step.  Your answer should be in the following format:
Thought: <your thought>
Output:
<execution result>
Following are 4 examples: 
{{examples here}}

\end{lstlisting}
\end{tcolorbox}

\subsubsection{Demo Questions}

\begin{tcolorbox}[left=0mm,right=0mm,top=0mm,bottom=0mm,boxsep=1mm,arc=0mm,boxrule=0pt, frame empty, breakable]
    \small
    \begin{lstlisting}
// Import necessary packages
import java.util.*;

class Solution {

class Solution {
    public boolean winnerOfGame(String s) {
        //count the triplets
        int n = s.length();
    
        int a=0;
        int b=0;
        
        for(int i=1; i<n-1; i++)
        {
            if(s.charAt(i)=='A' && s.charAt(i-1)=='A' && s.charAt(i+1)=='A' )
                a++;
            else if(s.charAt(i)=='B' && s.charAt(i-1)=='B' && s.charAt(i+1)=='B' )
                b++;
        }
        if(a == b)
            return false;
        else
            return true;
    }
}

public class Main {
    public static void main(String[] args) {
        Solution solution = new Solution();

        // Test case 1
        String colors1 = "AAABABB";
        System.out.println("Test Case 1: " + solution.winnerOfGame(colors1)); // Alice wins

        // Test case 2
        String colors2 = "AA";
        System.out.println("Test Case 2: " + solution.winnerOfGame(colors2)); // Bob wins

        // Test case 3
        String colors3 = "ABBBBBBBAAA";
        System.out.println("Test Case 3: " + solution.winnerOfGame(colors3)); // Bob wins
\end{lstlisting}
\end{tcolorbox}

\begin{tcolorbox}[left=0mm,right=0mm,top=0mm,bottom=0mm,boxsep=1mm,arc=0mm,boxrule=0pt, frame empty, breakable]
    \small
    \begin{lstlisting}
import java.util.Arrays;

public class Main {

class Solution {
    public int matrixSum(int[][] nums) {
        int score = 0;
        int n = nums.length;
        int m = nums[0].length;
        for(int[] a :nums)
        {
            Arrays.sort(a);
        }
        for(int i=0;i<=n;i++)
        {
            int max = 0;
            for(int j=0;j<m;j++)
            {
                max = Math.max(max,nums[i][j]);
            }
            score+=max;
        }
        return score;
    }
}

    public static void main(String[] args) {
        Solution solution = new Solution();

        // Test case 1
        int[][] nums1 = {
            {7, 2, 1},
            {6, 4, 2},
            {6, 5, 3},
            {3, 2, 1}
        };
        System.out.println(solution.matrixSum(nums1)); // Output: 15

        // Test case 2
        int[][] nums2 = {
            {1}
        };
        System.out.println(solution.matrixSum(nums2)); // Output: 1
\end{lstlisting}
\end{tcolorbox}

\begin{tcolorbox}[left=0mm,right=0mm,top=0mm,bottom=0mm,boxsep=1mm,arc=0mm,boxrule=0pt, frame empty, breakable]
    \small
    \begin{lstlisting}
from collections import defaultdict
from typing import List


class Solution:
    def numberOfArithmeticSlices(self, nums: List[int]) -> int:
        total, n = 0, len(nums)
        dp = [defaultdict(int) for _ in nums]
        for i in range(1, n):
            for j in range(i):
                diff = nums[j] - nums[i]
                dp[i][diff] += dp[j][diff] + 1
                total += self.undifned_method(dp[j][diff])
        return total

# Test cases
if __name__ == "__main__":
    solution = Solution()

    # Test case 1
    nums1 = [2, 4, 6, 8, 10]
    result1 = solution.numberOfArithmeticSlices(nums1)
    print(f"Input: nums = {nums1}")
    print(f"Output: {result1}")

    # Test case 2
    nums2 = [7, 7, 7, 7, 7]
    result2 = solution.numberOfArithmeticSlices(nums2)
    print(f"Input: nums = {nums2}")
    print(f"Output: {result2}")
\end{lstlisting}
\end{tcolorbox}

\begin{tcolorbox}[left=0mm,right=0mm,top=0mm,bottom=0mm,boxsep=1mm,arc=0mm,boxrule=0pt, frame empty, breakable]
    \small
    \begin{lstlisting}
#include <iostream>
#include <cmath>


class Solution {
public:
    long long fact(int n)
    {
        if(n<=1)return 1;
        return (n*fact(n-1)%1000000007)%1000000007;
    }
    int numPrimeArrangements(int n) {
        if(n==1)return 1;
        if(n<=3)return n-1;
        int t=0,flag;
        for(int i=2;i<=n;i++)
        {
            flag=0;
            for(int j=2;j<sqrt(i);j++)
            {
                if(i%j==0)
                {
                    flag=1;
                    break;
                }
            }
            if(flag==0)
            {
                t++;
            }
        }
        return (fact(t)*fact(n-t))%1000000007;

    }
};

int main() {
    Solution solution;
    // Test case 1
    int n1 = 5;
    std::cout << "Input: n = " << n1 << "\nOutput: " << solution.numPrimeArrangements(n1) << std::endl;

    // Test case 2
    int n2 = 100;
    std::cout << "Input: n = " << n2 << "\nOutput: " << solution.numPrimeArrangements(n2) << std::endl;

    return 0;
\end{lstlisting}
\end{tcolorbox}

\begin{tcolorbox}[left=0mm,right=0mm,top=0mm,bottom=0mm,boxsep=1mm,arc=0mm,boxrule=0pt, frame empty, breakable]
    \small
    \begin{lstlisting}
#include <iostream>
#include <string>
#include <cctype> // For isalpha

using namespace std;


class Solution {
public:
    str reverseOnlyLetters(string s) 
    {
      int i=0,j=s.length()-1;
      while(i<=j)
      {
        if(isalpha(s[i])&&isalpha(s[j]))
        {
            swap(s[i],s[j]);
            i++;
            j--;
        }
        else
        {
            if(!isalpha(s[i]))
            {
                i++;
            }
            if(!isalpha(s[j]))
            {
                j--;
            }
        }
      }
      return s;
    }
};

int main() {
    // Initialize the Solution class
    Solution solution;

    // Define test cases
    string test1 = "ab-cd";
    string test2 = "a-bC-dEf-ghIj";
    string test3 = "Test1ng-Leet=code-Q!";

    // Run test cases and print results
    cout << "Test 1: " << solution.reverseOnlyLetters(test1) << endl;
    cout << "Test 2: " << solution.reverseOnlyLetters(test2) << endl;
    cout << "Test 3: " << solution.reverseOnlyLetters(test3) << endl;

    return 0;
\end{lstlisting}
\end{tcolorbox}
\subsection{DR}

\subsubsection{System Prompts}

\paragraph{Zero-shot Chain-of-Thought:}

\begin{tcolorbox}[left=0mm,right=0mm,top=0mm,bottom=0mm,boxsep=1mm,arc=0mm,boxrule=0pt, frame empty, breakable]
    \small
    \begin{lstlisting}
Given the following code, what is the execution result? The file is under `/app/` directory, and is run with /bin/bash -c 'g++ -std=c++C++14 O1 test.cpp -o test && ./test'.
You should think step by step. Your answer should be in the following format:
Thought: <your thought>
Output:
<execution result>
\end{lstlisting}
\end{tcolorbox}

\paragraph{Zero-shot:}

\begin{tcolorbox}[left=0mm,right=0mm,top=0mm,bottom=0mm,boxsep=1mm,arc=0mm,boxrule=0pt, frame empty, breakable]
    \small
    \begin{lstlisting}
Given the following code, what is the execution result? The file is under `/app/` directory, and is run with /bin/bash -c 'g++ -std=c++C++14 O1 test.cpp -o test && ./test'.
Your answer should be in the following format:
Output:
<execution result>
\end{lstlisting}
\end{tcolorbox}

\paragraph{Few-shot Chain-of-Thought:}

\begin{tcolorbox}[left=0mm,right=0mm,top=0mm,bottom=0mm,boxsep=1mm,arc=0mm,boxrule=0pt, frame empty, breakable]
    \small
    \begin{lstlisting}
Given the following code, what is the execution result? The file is under `/app/` directory, and is run with /bin/bash -c 'g++ -std=c++C++14 O1 test.cpp -o test && ./test'.
You should think step by step. Your answer should be in the following format:
Thought: <your thought>
Output:
<execution result>
Following are 6 examples: 
\end{lstlisting}
\end{tcolorbox}

\subsubsection{Demo Questions}

\begin{tcolorbox}[left=0mm,right=0mm,top=0mm,bottom=0mm,boxsep=1mm,arc=0mm,boxrule=0pt, frame empty, breakable]
    \small
    \begin{lstlisting}
struct NonPOD {
    NonPOD() {}
    int x;
};
int main() {

    static_assert(std::is_pod<NonPOD>::value, "");
}
\end{lstlisting}
\end{tcolorbox}

\begin{tcolorbox}[left=0mm,right=0mm,top=0mm,bottom=0mm,boxsep=1mm,arc=0mm,boxrule=0pt, frame empty, breakable]
    \small
    \begin{lstlisting}
#include <coroutine>
struct task {
    struct promise_type { /*...*/ };

};
\end{lstlisting}
\end{tcolorbox}

\begin{tcolorbox}[left=0mm,right=0mm,top=0mm,bottom=0mm,boxsep=1mm,arc=0mm,boxrule=0pt, frame empty, breakable]
    \small
    \begin{lstlisting}
#include <atomic>
#include <thread>
#include <iostream>

std::atomic<int> data{0};

void writer() {
    data.store(1, std::memory_order_relaxed);
}

void reader() {
    while (data.load(std::memory_order_relaxed) == 0);
    std::cout << "Data updated";
}

int main() {
    std::thread t1(writer), t2(reader);
    t1.join(); t2.join();
}
\end{lstlisting}
\end{tcolorbox}

\begin{tcolorbox}[left=0mm,right=0mm,top=0mm,bottom=0mm,boxsep=1mm,arc=0mm,boxrule=0pt, frame empty, breakable]
    \small
    \begin{lstlisting}
#include <iostream>

struct S {
    S() { std::cout << "ctor\n"; }
    ~S() { std::cout << "dtor\n"; }
    S(const S&) { std::cout << "copy\n"; }
};

const S& getTemp() {
    return S();
}

int main() {
    const S& ref = getTemp();
    std::cout << "main\n";
    return 0;
}
\end{lstlisting}
\end{tcolorbox}

\begin{tcolorbox}[left=0mm,right=0mm,top=0mm,bottom=0mm,boxsep=1mm,arc=0mm,boxrule=0pt, frame empty, breakable]
    \small
    \begin{lstlisting}
template<typename T> void f(T) { std::cout << "1"; }
template<> void f(int*) { std::cout << "2"; }
template<typename T> void f(T*) { std::cout << "3"; }
int main() {
    int* p = nullptr;
    f(p);
}
\end{lstlisting}
\end{tcolorbox}

\subsection{FL}


\textbf{Hard for Latex to Compile, check the source code.}

\section{Training Details}
\label{appx:training}
For training, we employ Llama-Factory~\citep{zheng2024llamafactory} as the LLM training platform. Table~\ref{tab:hyperparameters} shows our training hyperparameters.

\begin{table}[!h]
    \centering
    \begin{tabular}{ll}
        \toprule
        Parameter        & Value                                           \\
        \midrule
        Train batch size & 128                                              \\
        Learning rate    & 1.0e-5                                          \\
        Number of epochs & 2.0                                             \\
        LR scheduler     & cosine                                          \\
        Warmup ratio     & 0.1                                             \\
        Precision        & bf16                                            \\
        \bottomrule
    \end{tabular}
    \caption{Hyperparameters for supervised fine-tuning.}
    \label{tab:hyperparameters}
\end{table}

\end{document}